\documentclass{article}

\usepackage{microtype}
\usepackage{graphicx}
\usepackage{booktabs} %

\usepackage{hyperref}

\usepackage[accepted]{icml2023}

\usepackage{amsmath}
\usepackage{amssymb}
\usepackage{mathtools}
\usepackage{amsthm}

\usepackage{mathtools}
\usepackage{pifont}         %
\usepackage[utf8]{inputenc} %
\usepackage[T1]{fontenc}    %
\usepackage{url}            %
\usepackage{amsfonts}       %
\usepackage{nicefrac}       %
\usepackage{lipsum}
\usepackage{fancyhdr}       %
\usepackage{caption}
\usepackage{flushend}
\usepackage{multirow}
\usepackage{subcaption}
\usepackage{wrapfig}
\usepackage{xcolor}         %
\usepackage[frozencache]{minted}         %
\usepackage{tikz}           %
\usepackage{pifont}         %
\usepackage{makecell}       %
\usepackage[export]{adjustbox} %
\usepackage{dblfloatfix}    %
\usepackage{placeins}       %
\usepackage{siunitx}        %

\definecolor{skyblue}{rgb}{0.37,0.87,0.96}%

\newcommand{\cmark}{\ding{51}}%
\newcommand{\xmark}{\ding{55}}%
\DeclareMathSymbol{\shortminus}{\mathbin}{AMSa}{"39}

\graphicspath{assets}

\usepackage{amsmath,amsfonts,bm}

\def\eqref#1{equation~\ref{#1}}

\def\1{\bm{1}}

\def\vj{{\bm{j}}}

\def\vs{{\bm{s}}}

\def\vx{{\bm{x}}}

\def\evd{{d}}

\def\evs{{s}}

\def\evx{{x}}

\def\mA{{\bm{A}}}

\def\mD{{\bm{D}}}

\def\mH{{\bm{H}}}
\def\mI{{\bm{I}}}

\def\mK{{\bm{K}}}
\def\mL{{\bm{L}}}

\def\mQ{{\bm{Q}}}
\def\mR{{\bm{R}}}
\def\mS{{\bm{S}}}
\def\mT{{\bm{T}}}
\def\mU{{\bm{U}}}
\def\mV{{\bm{V}}}

\def\mX{{\bm{X}}}

\DeclareMathAlphabet{\mathsfit}{\encodingdefault}{\sfdefault}{m}{sl}
\SetMathAlphabet{\mathsfit}{bold}{\encodingdefault}{\sfdefault}{bx}{n}

\def\gG{{\mathcal{G}}}

\def\emA{{A}}

\def\emR{{R}}
\def\emS{{S}}
\def\emT{{T}}

\def\emX{{X}}

\newcommand{\E}{\mathbb{E}}

\newcommand{\R}{\mathbb{R}}

\newcommand{\normltwo}{L^2}

\DeclareMathOperator*{\argmax}{arg\,max}

\newcommand*{\rom}[1]{\uppercase\expandafter{\romannumeral #1\relax}}

\newcommand{\PE}{\operatorname{PE}}
\newcommand{\eEigval}{\Lambda}
\newcommand{\Eigval}{\boldsymbol{\Lambda}}
\newcommand{\eigval}{\lambda}
\newcommand{\eEigvec}{\Gamma}
\newcommand{\Eigvec}{\boldsymbol{\Gamma}}
\newcommand{\eigvec}{\boldsymbol{\gamma}}
\newcommand{\C}{\mathbb{C}}

\usepackage[capitalize,noabbrev]{cleveref}

\icmltitlerunning{Transformers Meet Directed Graphs}

\begin{document}

\twocolumn[
\icmltitle{Transformers Meet Directed Graphs}

\icmlsetsymbol{equal}{*}

\begin{icmlauthorlist}
\icmlauthor{Simon Geisler}{equal,tum}
\icmlauthor{Yujia Li}{dm}
\icmlauthor{Daniel Mankowitz}{dm}
\icmlauthor{Ali Taylan Cemgil}{dm}
\icmlauthor{Stephan G\"unnemann}{tum}
\icmlauthor{Cosmin Paduraru}{dm}
\end{icmlauthorlist}

\icmlaffiliation{tum}{Dept. of Computer Science \& Munich Data Science Institute, Technical University of Munich}
\icmlaffiliation{dm}{Google DeepMind}

\icmlcorrespondingauthor{Simon Geisler}{s.geisler@tum.de}

\icmlkeywords{Transformer, Graphs, Positional Encoding}

\vskip 0.3in
]

\printAffiliationsAndNotice{}  %

\begin{abstract}
Transformers were originally proposed as a sequence-to-sequence model for text but have become vital for a wide range of modalities, including images, audio, video, and \emph{undirected} graphs. 
However, transformers for \emph{directed} graphs are a surprisingly underexplored topic, despite their applicability to ubiquitous domains, including source code and logic circuits. 
In this work, we propose two direction- and structure-aware positional encodings for \emph{directed} graphs: (1) the eigenvectors of the Magnetic Laplacian -- a direction-aware generalization of the combinatorial Laplacian; (2) directional random walk encodings. Empirically, we show that the extra directionality information is useful in various downstream tasks, including correctness testing of sorting networks and source code understanding. 
Together with a data-flow-centric graph construction, our model outperforms the prior state of the art on the Open Graph Benchmark Code2 relatively by 14.7\%.\hyperlink{code}{\textsuperscript{3}}
\end{abstract}

\section{Introduction}\label{sec:intro}

Transformers have become a central component in many state-of-the-art machine learning models spanning a wide range of modalities. For example, transformers are used to generate solutions for competitive programming tasks from textual descriptions~\citep{li_competition-level_2022}, for conversational question answering with the popular ChatGPT~\citep{openai_chatgpt_2022}, or to find approximate solutions to combinatorial optimizations problems like the Traveling Salesman Problem~\citep{kool_attention_2019}. 
Transformers have also had success in graph learning tasks, e.g., for predicting the properties of molecules~\citep{min_transformer_2022}. While virtually all prior works focus on undirected graphs, we advocate the use of directed graphs as they are omnipresent, and directedness can rule semantics. Transformers that handle both undirected and directed graphs could become an important building block for many applications. For this, the attention mechanism needs to become aware of the graph structure. For example, prior work modified the attention mechanism to incorporate structural information \citep{ying_transformers_2021} or proposed hybrid architectures that also contain Graph Neural Networks (GNNs) \citep{mialon_graphit_2021, chen_structure-aware_2022}. Another (complementary) option are positional encodings that are used by many, if not most, structure-aware transformers~\citep{min_transformer_2022, muller_attending_2023}.

\begin{figure}[b]
  \centering
  \makebox[\linewidth][c]{
    \(\begin{array}{ccccc}
      \subfloat[Sequence\label{fig:example_graphs:a}]{
      \includegraphics[width=0.22\linewidth]{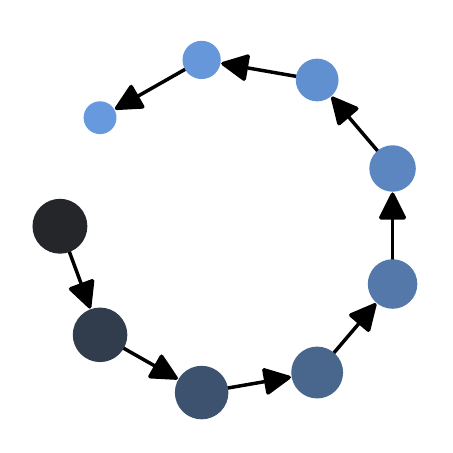}} &
      \subfloat[Undir. seq.\label{fig:example_graphs:b}]{
      \includegraphics[width=0.22\linewidth]{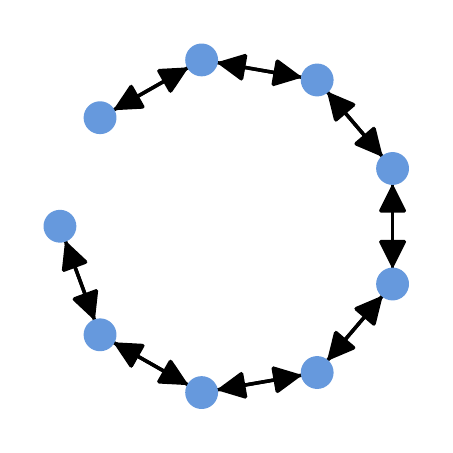}} &
      \subfloat[Binary tree\label{fig:example_graphs:c}]{
      \includegraphics[width=0.22\linewidth]{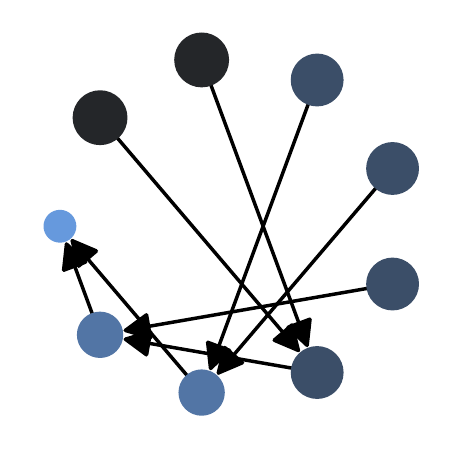}} &
      \subfloat[Trumpet\label{fig:example_graphs:d}]{
      \includegraphics[width=0.22\linewidth]{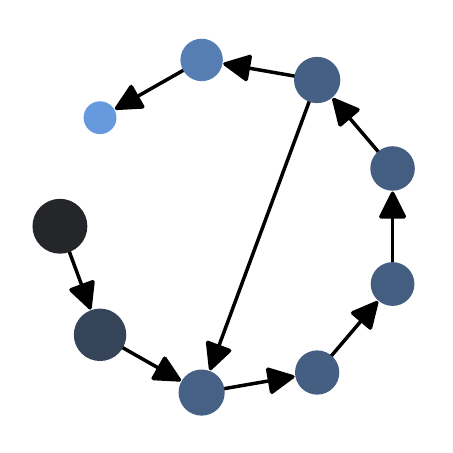}} \\
    \end{array}\)
  }
  \caption{First eigenvector of Magnetic Laplacian. Node size encodes the real value and colors the imaginary value.}
  \label{fig:example_graphs}
\end{figure}

\textbf{Directional positional encodings.}
Most of the literature for structure-aware positional encodings either uses basic measures like pair-wise shortest path distances~\citep{guo_graphcodebert_2021, ying_transformers_2021} or symmetrizes the graph for principled positional encodings, e.g., based on the combinatorial Laplacian~\citep{dwivedi_generalization_2021,kreuzer_rethinking_2021}. Importantly, symmetrization might discard essential information that determines the semantics of the input. 
For these reasons, we propose two direction-aware positional encodings: (1) the eigenvectors of the Magnetic Laplacian (\autoref{sec:spectral}), which naturally generalizes the well-known combinatorial Laplacian to directed graphs (see \autoref{fig:example_graphs}); and (2) directional random walk encodings  (\autoref{sec:random_walks}) that generalize basic measures like the shortest path distances. We show that our positional encodings are predictive for various distances on graphs (\autoref{sec:playground}) and are useful in downstream tasks. Moreover, our positional encodings can also improve GNNs (see \autoref{fig:sn_gnn}).

\textbf{Motivation for directed graphs.} We make the impact of appropriately modeling inputs via directed graphs explicit for our applications. One example is the correctness prediction of sorting networks (\autoref{sec:sorting_networks}). Sorting networks~\citep{knuth_art_1973} are a certain sorting procedures that can be represented by a fixed sequence of operations. The goal is then to predict whether the sequence is a correct sorting network. Based on the sequence of operations, we can construct a (directed acyclic) data-flow graph that models the dependencies between the operations. Conversely, the topological sorts of this graph correspond to different but semantically equivalent sequences of operations. Considering the potentially large number of topological sorts, directed graphs can drastically reduce the effective input dimensionality (e.g., see \autoref{fig:sorting_batcher_n_topsorts}). Moreover, we show that ignoring the edge directions maps both correct and incorrect sorting networks to the same \emph{undirected} graph, losing critical information.

Interestingly, representing source code as a sequence is the de facto standard~\citep{li_competition-level_2022, feng_codebert_2020, chen_evaluating_2021, openai_chatgpt_2022}. Even graph-based representations of code~\citep{allamanis_learning_2018, hu_open_2020,  cummins_programl_2020, guo_graphcodebert_2021, bieber_library_2022} only enrich sequential source code, e.g., with an Abstract Syntax Tree (AST). However, similar to sorting networks, we can often reorder certain statements without affecting the code's functionality. This motivates us to rethink the graph construction for source code, which not only boosts performance but makes the model invariant w.r.t.\ certain meaningless reorderings of statements (see \autoref{sec:source_code} for details).

\textbf{Contributions:} 
\textbf{[\rom{1}]} We make the connection between sinusoidal positional encodings and the eigenvectors of the Laplacian explicit (\autoref{sec:sinlap}).\;
\textbf{[\rom{2}]} We propose \emph{spectral positional encodings} that also generalize to directed graphs  (\autoref{sec:spectral}).\;
\textbf{[\rom{3}]} We extend random walk positional encodings to directed graphs (\autoref{sec:random_walks}).\;
\textbf{[\rom{4}]} As a plausibility check, we assess the predictiveness of structure-aware positional encodings for a set of graph distances (\autoref{sec:playground}).\;
\textbf{[\rom{5}]} We introduce the task of predicting the correctness of \emph{sorting networks}, a canonical ambiguity-free application where directionality is essential (\autoref{sec:sorting_networks}).\;
\textbf{[\rom{6}]} We quantify the benefits of modeling a sequence of program statements as a directed graph and rethink the graph construction for source code to boost predictive performance and robustness (\autoref{sec:sorting_networks} \& \ref{sec:source_code}).\;
\textbf{[\rom{7}]} We set a new \emph{state of the art} on the OGB Code2 dataset (2.85\% higher F1 score, 14.7\% relatively) for function name prediction (\autoref{sec:source_code}).

\section{Sinusoidal and Laplacian Encodings}
\label{sec:sinlap}

\begin{figure}[t]
  \centering
  \subfloat[Sinusoidal]{
  \includegraphics[width=0.44\linewidth]{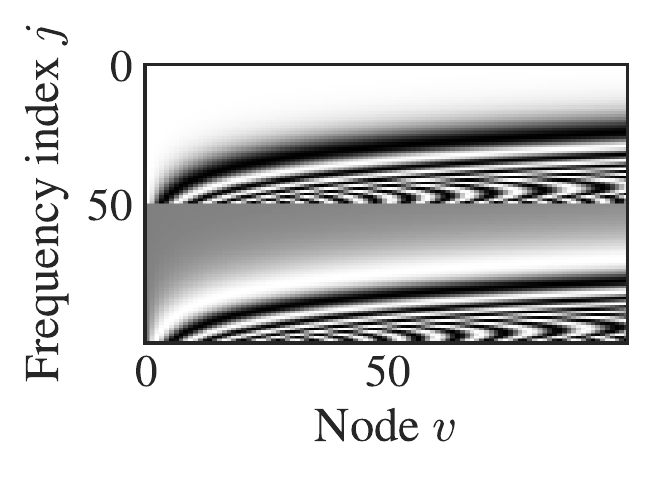}}
  \hfill{}
  \subfloat[Eigenvec.\ \(\Eigvec\) of Laplacian]{
  \includegraphics[width=0.44\linewidth]{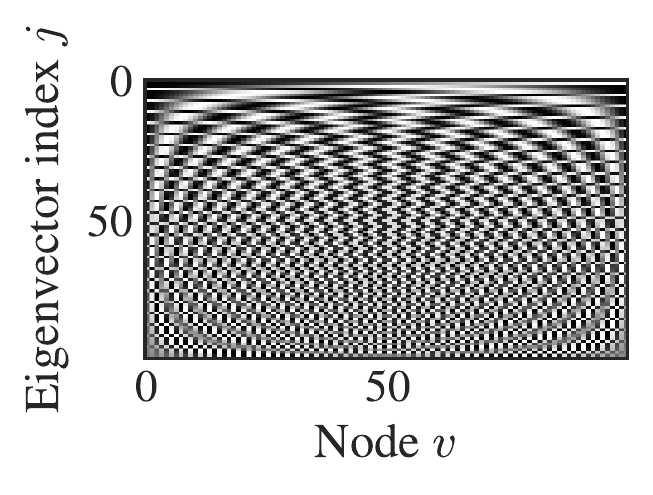}}
  \caption{(a) Sinusoidal encodings (\(sin\) components top and \(cos\) below) with denominator \(1,000^{2j / d_{\text{model}}}\) and \(d_{\text{model}}=100\). (b) Lap.\ eigenvec.\ of sequence \autoref{fig:example_graphs:b} of len.\ \(n=100\).
  \label{fig:sin_lap_bitmap}}
  \vspace{-0.1in}
\end{figure}

Due to the permutation equivariant attention, one typically introduces domain-specific inductive biases with Positional Encodings (PEs). For example, \citet{vaswani_attention_2017} proposed sinusoidal positional encodings for sequences along with the transformer architecture. It is commonly argued~\citep{bronstein_geometric_2021, dwivedi_generalization_2021} that the eigenvectors of the (combinatorial) Laplacian can be understood as a generalization of the sinusoidal positional encodings (see \autoref{fig:sin_lap_bitmap}) to graphs, due to their relationship via the Graph Fourier Transformation (GFT) and Discrete Fourier Transformation (DFT)~\citep{smith_scientist_1999}. Even though sinusoidal positional encodings capture the direction, eigenvectors of the Laplacian do not. But why is this the case? To understand differences and commonalities, we next contrast sinusoidal encodings, DFT, and Laplacian eigenvectors for a sequence (\autoref{fig:example_graphs:a},\ref{fig:example_graphs:b}).

\textbf{Sequence encodings.} Sinusoidal encodings~\citep{vaswani_attention_2017} form a \(d_{\text{model}}\)-dimensional embedding of token \(v\)'s integer position in the sequence using cosine \(\PE^{(\sin)}_{v, 2j} \coloneqq \cos(\nicefrac{v}{10,000^{2 j / d_{\text{model}}}})\) and sinus \(\PE^{(\sin)}_{v, 2j + 1} \coloneqq \sin(\nicefrac{v}{10,000^{2 j / d_{\text{model}}}})\) waves of varying frequencies with \(j \in \{0, 1, ..., \nicefrac{d_{\text{model}}}{2} - 1\}\). Analogously, the \textbf{DFT} could be used to define positional encodings:
\begin{equation}\label{eq:dft}
    \resizebox{!}{0.99\height}{\(
      X_j \coloneqq
      \sum\limits_{v = 0}^{n - 1} x_v \Big[ \underbrace{\cos\Big(v \cdot \frac{2 \pi}{n} j \Big)}_{\PE^{(\text{DFT})}_{v, 2j}} - i \cdot \underbrace{\sin\Big(v \cdot \frac{2 \pi}{n} j \Big)}_{\PE^{(\text{DFT})}_{v, 2j + 1}} \Big]
      \)}
\end{equation}
\\
Here \(X\) corresponds to signal \(x\) in the frequency domain. In contrast to the DFT, sinusoidal encodings (a) sweep the frequencies using a geometric series instead of linear; (b) also contain frequencies below \(\nicefrac{1}{n}\) (longest wavelength is \(2 \pi 10,000\)); and (c) have \(d_{\text{model}}\) components instead of \(2n\) (i.e., \(0 \le j < n\) in \autoref{eq:dft}).

\textbf{Graphs} generalize sequences to sets of tokens/nodes with arbitrary connections. In a graph \(\gG=(V, E)\), the \(m\) edges \(E\) represent connections between the \(n\) nodes \(V\). We use \(\mX^{(n)}\) for node features and \(\mX^{(m)}\) for edge features. We denote the in-degree of node \(u\) with \(|\{v | (v, u) \in E\}|\) and out-degree with \(|\{v | (u, v) \in E\}|\). Alternatively to \(E\), we denote the adjacency matrix \(\mA \in \{0, 1\}^{n \times n}\) and refer with \(\mD \in \R^{n \times n}\) to the diagonalized degree matrix. Analogously, we describe the symmetrized adjacency matrix \(\mA_s = \mA \lor \mA^\top\) with set of edges \(E_s\) and degree matrix  \(\mD_s\). In the main part of the paper, we only discuss unweighted graphs; however, our methods naturally generalize to weighted graphs (see \autoref{sec:appendix_weighted}).

\textbf{Eigenvectors of Laplacian.} The Graph Fourier Transformation (GFT) for undirected graphs \(\mX = \Eigvec^\top \vx\)) can be defined based on the eigendecomposition of the combinatorial Laplacian \(\mL = \Eigvec \Eigval \Eigvec^{-1}\), with diagonal matrix \(\Eigval\) of eigenvalues and orthogonal matrix \(\Eigvec \in \R^{n \times n}\)
of eigenvectors (see \autoref{sec:appendix_graph_fourier_transformation} for details). Similarly to the DFT, \(\Eigvec\) are suitable positional encodings. The unnormalized Laplacian \(\mL_{U}\) as well as degree-normalized Laplacian \(\mL_{N}\) are defined as:\\
\begin{minipage}{0.37\linewidth}
\begin{equation}\label{eq:laplacian_u}
    \mL_{U} \coloneqq \mD_s - \mA_s
\end{equation}
\end{minipage}
\begin{minipage}{0.62\linewidth}
\begin{equation}\label{eq:laplacian_s}
\mL_{N} \coloneqq \mI - ( \mD_{s}^{- \nicefrac{1}{2}} \mA_{s} \mD_{s}^{- \nicefrac{1}{2}}  )
\end{equation}
\end{minipage}
\vspace{3pt}\\
The symmetrization \(\mA_s\) discards directionality but is required s.t.\ \(\mL\) is guaranteed to be symmetric and positive semi-definite. This ensures that \(\Eigvec\) form an orthogonal basis, which entails important properties of the GFT and for PEs (see \autoref{sec:appendix_laplacians_directed_graphs} for a discussion). In the following, we index eigenvalues and eigenvectors by order: \(0 \le \eigval_0 \le \eigval_1 \le \dots \le \eigval_{n-1}\). We call \(\eigval_0\) or \(\eEigval_{0, 0}\) the first eigenvalue and \(\eigvec_0\) or \(\Eigvec_{:, 0}\) the first eigenvector that reflect the lowest frequency.

\textbf{Laplacian vs. DFT.} Two notable differences to the DFT in \autoref{eq:dft} are (1) that the eigenvectors of the Laplacian are real-valued; (2) the eigenvectors are not unique, e.g., due to the \emph{sign ambiguity}. That is, if \(\eigvec\) is an eigenvector, so is \(-\eigvec\).

\textbf{Cosine Transformation.} A possible set of eigenvectors of the combinatorial Laplacian for a sequence (\autoref{fig:example_graphs:b}) is given by the Cosine Transformation Type \rom{2} \citep{strang_discrete_1999}: 
\(\eEigvec_{v, j} = \pm \cos((v + \nicefrac{1}{2}) j \nicefrac{\pi}{n})\), where we must choose the same sign per \(j\). The \(\pm\) is due to the \emph{sign ambiguity} (2) of the eigenvector and, thus, we cannot distinguish the embedding of, e.g., the first and last node. Note that in traditional applications of the Cosine Transformation, it is possible to identify the first token and fix the sign. 
However, for general graphs, it is not that easy to resolve the sign ambiguity (e.g., multiple sink and source nodes). Thus, in positional encodings, we typically use an arbitrary sign for each \(\eigvec\)~\citep{dwivedi_generalization_2021}, and are not able to distinguish direction.

\section{Directional Spectral Encodings}
\label{sec:spectral}

The Magnetic Laplacian is a generalization of the combinatorial Laplacian that encodes the direction with complex numbers. We then use its eigenvectors for a structure-aware positional encoding that acknowledges the directedness.

We define the \textbf{Magnetic Laplacian} ~\citep{forman_determinants_1993, shubin_discrete_1994, de_verdiere_magnetic_2013, furutani_graph_2020}
\begin{equation}\label{eq:unsymmetric_magnetic_laplacian}
\begin{aligned}
    \mL^{(q)}_{U} 
    &\coloneqq \mD_{s} - \mA_{s} \odot \exp \big( i \boldsymbol{\Theta}^{(q)} \big)
\end{aligned}
\end{equation}
with Hadamard product \(\odot\), element-wise \(\exp\), \(\smash{i = \sqrt{\shortminus 1}}\),
\(
    \Theta^{(q)}_{u,v} \coloneqq 2\pi q (\emA_{u,v} - \emA_{v,u})
\), and potential \(q \ge 0\). 
Recall, \(\mD_{s}\) is the symmetrized degree matrix and \(\mA_{s}\) the symmetrized adjacency matrix. The Magnetic Laplacian is a Hermitian matrix since it is equal to its conjugate transpose \(\mL^{(q)} = (\bar{\mL}^{(q)})^\top\) and, thus, comes with complex eigenvectors \(\Eigvec \in \C^{n \times n}\).  \autoref{eq:unsymmetric_magnetic_laplacian} is equivalent to the combinatorial Laplacian for \(q=0\). 
Moreover, if the graph is undirected, we recover the combinatorial Laplacian for any finite \(q \in \R\).

The \(\exp \left( i \boldsymbol{\Theta}^{(q)}\right)\) term in \autoref{eq:unsymmetric_magnetic_laplacian} encodes the edge direction. It resolves to \(1\) if \(\emA_{u,v} = \emA_{v,u}\) and, otherwise, to \(\exp(\pm i 2\pi q)\), with the sign encoding the edge direction. The \emph{potential} \(q\) determines the ratio of real and imaginary parts. Recall that 
\(
    \exp(i \alpha) = \cos(\alpha) + i \sin(\alpha)
\). Conversely, \(\angle(\eEigvec_{u, 0}) = \operatorname{arctan2}(\Im(\eEigvec_{u, 0}), \Re(\eEigvec_{u, 0}))\) with real / imag.\ value \(\Re\) / \(\Im\). For illustration, we next give the Magnetic Laplacian for a sequence with \(q=0\) and \(q=\nicefrac{1}{4}\) (\autoref{eq:lap_chain_graph} \& \ref{eq:maglap_chain_graph}), as well as their first eigenvector in \autoref{fig:example_graphs:b} and \ref{fig:example_graphs:a}.\newline
\begin{minipage}{0.495\linewidth}
\begin{equation}\label{eq:lap_chain_graph}
\resizebox{0.81\linewidth}{!}{\(
    \mL_{U}^{(0)} = \begin{bsmallmatrix}
        1 & \shortminus 1 & \cdots & 0 & 0 \\
        \shortminus 1 & 2 & \cdots & 0 & 0 \\
        \vdots & \vdots & \ddots & \vdots & \vdots \\
        0 & 0 & \cdots & 2 & \shortminus 1 \\
        0 & 0 & \cdots & \shortminus 1 & 1 \\
        \end{bsmallmatrix}
\)}
\end{equation}
\end{minipage}
\begin{minipage}{0.495\linewidth}
\begin{equation}\label{eq:maglap_chain_graph}
\resizebox{0.81\linewidth}{!}{\(
\mL_U^{(\nicefrac{1}{4})} =
    \begin{bsmallmatrix}
        1 & \shortminus i & \cdots & 0 & 0 \\
        i & 2 & \cdots & 0 & 0 \\
        \vdots & \vdots & \ddots & \vdots & \vdots \\
        0 & 0 & \cdots & 2 & \shortminus i \\
        0 & 0 & \cdots & i & 1 \\
        \end{bsmallmatrix}
\)}
\end{equation}
\end{minipage}

In our experiments, we use the degree-normalized counterpart \(
    \mL^{(q)}_{N} 
    \coloneqq \mI - \left( \mD_{s}^{- \nicefrac{1}{2}} \mA_{s} \mD_{s}^{- \nicefrac{1}{2}}  \right) \odot \exp \left( i \boldsymbol{\Theta}^{(q)} \right)
\) that we find to result in slightly superior performance.

\begin{figure}[t]
  \centering
  \includegraphics[width=0.925\linewidth]{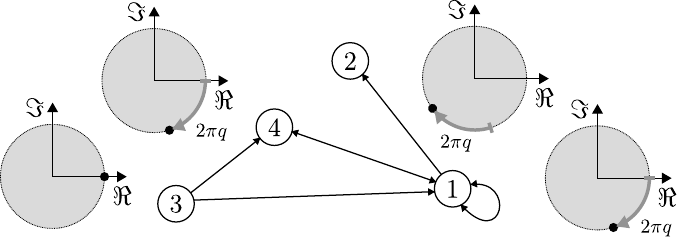}
  \caption{First eigenvec.\ \(\eigvec_0\) of Magnetic Laplacian (\autoref{eq:unsymmetric_magnetic_laplacian}).}
  \label{fig:exemplary_eigenvecotr_maglap}
  \vspace{-0.1in}
\end{figure}

\textbf{Directedness.} We next illustrate how the \emph{eigenvectors} of the Magnetic Laplacian \(\mL_U^{(q)}\) encode direction. For the special case of a sequence (\autoref{fig:example_graphs:a}), the eigenvectors are given by \(\eEigvec^{(q)}_{v, j} = c \exp(\shortminus i 2 \pi q v) \eEigvec^{(0)}_{v, j} = c \exp(\shortminus i 2 \pi q v) \cos((v + \nicefrac{1}{2}) j \nicefrac{\pi}{n})\) with \(c \in \C \setminus \{0\}\). This corresponds to the Cosine Transformation Type II (see \autoref{sec:sinlap}) with additional factor \(\exp(\shortminus i 2 \pi q v)\) that encodes the node position \(v\). Importantly, the eigenvectors of the Magnetic Laplacian also encode the directionality in arbitrary (directed) graph topologies, where each \emph{directed} edge \((u, v)\) encourages a phase difference in the (otherwise constant) first eigenvector \(\eigvec_0\), i.e., between \(\eEigvec_{u, 0}\) and \(\eEigvec_{v, 0}\). 
For simple cases with \(\mL_U^{(q)}\), as in \autoref{fig:exemplary_eigenvecotr_maglap}, each directed edge induces a rotation of \(2 \pi q\) while each undirected edge synchronizes the rotation of the adjacent nodes. Note that self-loops are assumed to be undirected and only affect the symmetrically degree-normalized \(\mL_N^{(q)}\). In this example, the one-hop neighbors of node 3, namely nodes 1 and 4, have a relative rotation of \(2 \pi q\), while the two-hop neighbor 2 has a relative rotation of \(4 \pi q\). In general, the first normalized eigenvector \(\eigvec\) minimizes the Rayleigh quotient
\begin{equation}\label{eq:maglap_raleigh}
    \min_{\vx \in \C} \frac{\bar{\vx}^\top \mL_U^{(q)} \vx}{\bar{\vx}^\top \vx}
    = \frac{1}{2} \sum_{(u,v) \in E_s} | \eEigvec_{u, 0} - \eEigvec_{v, 0} \exp(i \Theta^{(q)}_{u, v}) | ^2,
\end{equation}
Thus, the eigenvectors trade off conflicting edges, e.g., if multiple (directed) routes of different lengths exist between nodes \(u\) and \(v\). For more details, see \autoref{sec:appendix_magnetic_laplacian}.

\textbf{The potential \(q\)} determines the strength of the induced phase shift by each edge. Thus, \(q\) plays a similar role as the lowest frequency in sinusoidal positional encodings (typically \(\nicefrac{1}{2 \pi 10,000}\)). Following the convention of sinusoidal encodings, one could fix \(q\) to an appropriate value for the largest expected graphs. However, scaling potential \(q\) with the number of nodes \(n\) and the amount of directed edges leads to slightly superior performance in our experiments. 
Specifically, we choose \(q = \nicefrac{q'}{d_\gG}\) with \emph{relative potential} \(q'\) 
and graph-specific normalizer \(d_\gG\). This normalizer is an upper bound on the number of directed edges in a simple path \(d_\gG = \max(\min(\vec{m}, n), 1)\) with the number of \emph{purely directed edges} \(\vec{m} = |\{(u, v) \in E \,|\, (v, u) \notin E\}|\) 
and is motivated by \autoref{eq:maglap_raleigh} (see \autoref{sec:appendix_maglap_influence_q}). We typically choose \(q' \in \{0.1, 0.25\}\) and empirically verify this in \autoref{fig:sn_qsweep} where it is among the best. For high values of \(q'\) the performance drops severely (corresponds to absolute \(q > 0.05\)).

\textbf{Scale and rotation.} Eigenvectors are not unique and we normalize them by a convention. We visualize the first eigenvector \emph{after applying our normalization} for different graphs in \autoref{fig:example_graphs} \& \ref{fig:appendix_example_graphs}. One source of ambiguity is its scale and rotation. If \(\eigvec\) is an eigenvector of \(\mL\) then so is \(c \eigvec\), even if \(c \in \C \setminus \{0\}\) (proof: \(c \mL \eigvec = c \eigval \eigvec \implies \mL (c \eigvec) = \eigval (c \eigvec)\)). For real symmetric matrices, there is the convention to choose \(c \in \R\) s.t.\ \(\Eigvec\) is orthonormal (\(\Eigvec^\top\Eigvec = \mI\)). 
Similarly, (1) we choose \(|c|\) s.t.\ \(\Eigvec\) is unitary (\(\bar{\Eigvec}^\top\Eigvec = \mI\)). Moreover, if not using a sign-invariant architecture, as described below, (2) we determine the sign of each eigenvector such that the maximum real magnitude is positive. This resolves the sign-ambiguity up to ties in the maximum real magnitude and numerical errors. (3) we fix the rotation. If possible for the task at hand, we use the graph's distinct ``root'' node \(u\). For example, in function name prediction, the root node marks the start of the function definition. Alternatively, we use the foremost (source) node \(u\) as root, i.e., the node with maximum phase shift in the first eigenvector \(u = \argmax_v\angle(\eEigvec_{v, 0})\). In both cases, we then rotate all eigenvectors, such that the phase shift \(\angle(\eEigvec_{u,j})\) is 0 for all \(j\in\{0,1,\dots,n-1\}\). Due to the rotation in (3), our normalization is best suited for graphs with root/source node(s). For details, see \autoref{sec:appendix_maglap_scale_rotate}.

\begin{figure}[t]
  \centering
  \hfill{}
  \subfloat[Transformer Encoder\label{fig:maglapnet_with_trans:a}]{\includegraphics[height=5.6cm]{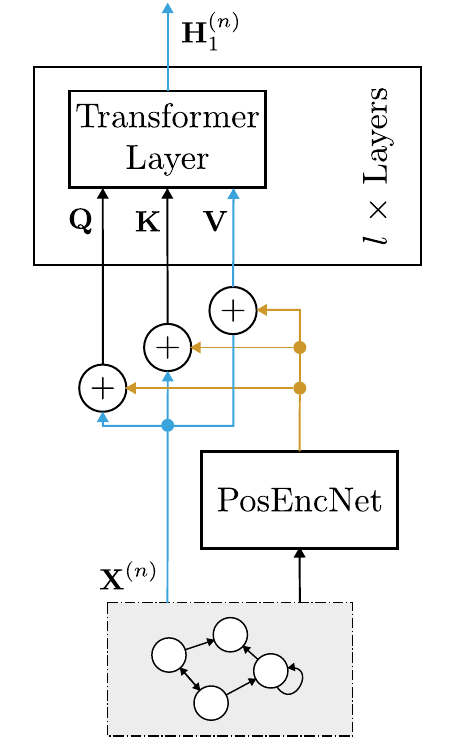}}
  \hfill{}
  \subfloat[MagLapNet\label{fig:maglapnet_with_trans:b}]{\includegraphics[height=5.6cm]{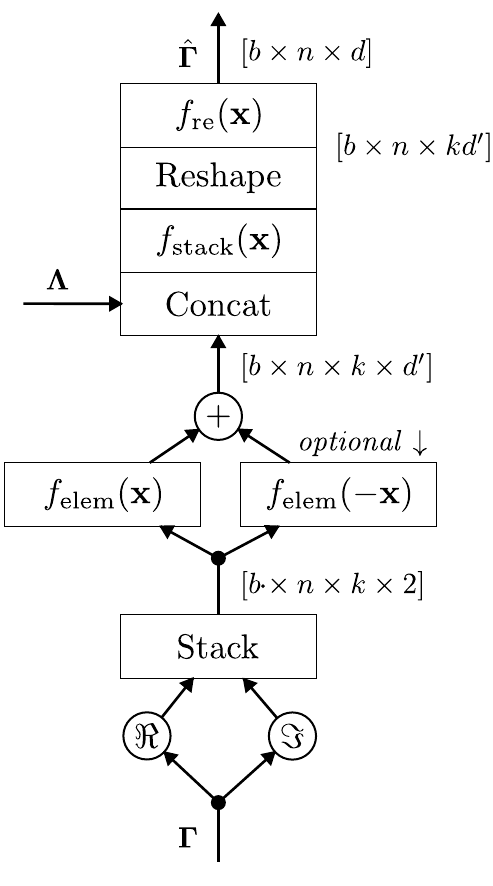}}
  \hfill{}
  \caption{(a) shows a transformer encoder operating on a graph with omitted residual connection. (b) is one specific instantiation of the (optional) ``PosEncNet'' using the eigenvectors of the Magnetic Lap (see ``MagLapNet'' paragraph) with batch size \(b\). See \autoref{sec:random_walks} for random walk encodings.}
  \label{fig:maglapnet_with_trans}
  \vspace{-0.12in}
\end{figure}

\textbf{MagLapNet.} Inspired by prior approaches~\citep{lim_sign_2022, kreuzer_rethinking_2021}, we also preprocess eigenvectors before using them as positional encodings (\autoref{fig:maglapnet_with_trans:b}) to obtain a structure-aware transformer (\autoref{fig:maglapnet_with_trans:a}). We consider the eigenvectors associated with the \(k\) lowest eigenvalues \(\Eigvec_{:, :k \shortminus 1}\) and treat \(k\) as hyperparameter. 
We study two architecture variants for processing the eigenvectors after stacking real and imaginary components: (a) a model that ignores the sign-invariance \(f_{\text{elem}}(\eigvec)\) and (b) the sign-invariant SignNet~\citep{lim_sign_2022} that processes each eigenvector as \(f_{\text{elem}}(-\eigvec_j) + f_{\text{elem}}(\eigvec_j)\), where \(f_{\text{elem}}\) is permutation equivariant over the nodes, like a point-wise Multi-Layer Perceptron (MLP) or GNN. 
However, when utilizing an MLP, we observe that choice (a) yields superior performance. This outcome can be attributed to several factors, including our aforementioned selection of the sign (see above) and the characteristic of a point-wise MLP to disregard relative differences in \(\eigvec\). Note that we always process the first eigenvector as \(f_{\text{elem}}(\eigvec_j)\) since we fully resolve its sign ambiguity.

Thereafter, we apply LayerNorm, Self-Attention, and Dropout. Similar to \citet{kreuzer_rethinking_2021}, we apply self-attention independently for each node \(u\) over its \(k\) eigenvector embeddings. In other words, for each node, we apply self-attention over a set of \(k\) tokens. This models the node-wise interactions between the eigenvectors, i.e., (\(\Eigvec_{u, :k \shortminus 1}\)) for node \(u\). The last reshape stacks each node's encoding, and the MLP \(f_{\text{re}}\) matches the transformer dimensions.

\section{Directional Random Walks}\label{sec:random_walks}

An alternative principled approach for encoding node positions in a graph is through random walks. \citet{li_distance_2020} have shown that such positional encodings can provably improve the expressiveness of GNNs, and such random walk encodings have been applied to transformers as well~\citep{mialon_graphit_2021}. Interestingly, random walks generalize, e.g., shortest path distances via the number of steps required for a non-zero landing probability. However, na\"ively applying random walks to directed graphs comes with caveats. 

\textbf{Random walks on graphs.} A \(k\)-step random walk on a graph is naturally expressed via the powers \(\mT^k\) of the transition matrix \(\mT = \mA \mD_{\text{out}}^{-1}\). In each step, the random walk proceeds along one of the outgoing edges with equal probability or probability proportional to the edge weight. We then obtain the landing probability \((\mT^k)_{u,v}\) at node \(u\) if starting from node \(v\). Note that even in connected graphs, we might have node pairs \(v, u\) that have zero transition probability regardless of \(k\). Thus, the na\"ive application of random walks for positional encodings on directed graphs is not ideal.

\textbf{Directedness.} To overcome the issue of only walking in \emph{forward} direction and in contrast to \citet{li_distance_2020}, we additionally consider the \emph{reverse} direction \(\mR = \mA^\top \mD_{\text{in}}^{-1}\). Additionally, we add self-loops to sink nodes (nodes with zero out or in degree for \(\mT\) or \(\mR\), respectively). This avoids that \(\mA\) might be nilpotent and ensures that the landing probabilities sum up to one. We then define the positional encoding for node \(v\) as \(\zeta(v | \gG) = f_{\text{rw}}^{(1)}(\operatorname{AGG}(\{\zeta(v | u)\,|\,u \in V\}))\), where \(\zeta(v | u) = \resizebox{\linewidth}{\height}{\( f^{(2)}_{\text{rw}}[(\mR^k)_{v,u}, \dots, (\mR^2)_{v,u}, \emR_{v,u}, \emT_{v,u}, (\mT^2)_{v,u}, \dots, (\mT^k)_{v,u}]\)}\) and \(\operatorname{AGG}\) performs summation. \(f^{(1)}_{\text{rw}}\) and \(f^{(2)}_{\text{rw}}\) is an MLP.

\textbf{Large distances.} A large amount of random walk steps \(k\) is expensive and for a sufficiently large \(k\) the probability mass concentrates in sink nodes. Thus, the random walk positional encodings are best suited for capturing short distances. For the global relations, we extend \(\zeta(v | u)\) with a forward and reverse infinite step random walk, namely Personalized Page Rank (PPR)~\citep{page_pagerank_1999}. Importantly, PPR includes the restart probability \(p_r\) to jump back to the starting node \(u\) and has closed form solution \(p_r (\mI - (1 - p_r) \mT)^{-1}\).

We provide an overview of our positional encodings in \autoref{sec:appendix_overview} and discuss computational cost/complexity in \autoref{sec:appendix_scalablity}.

\section{Positional Encodings Playground}\label{sec:playground}

We next assess the efficacy of our two directional structure-aware positional encodings. As there is no (established) way of assessing positional encodings standalone, we rely on downstream tasks. In our first task, we verify if the encodings are predictive for distances on graphs.

\textbf{Tasks.} We hypothesize that a good positional encoding should be able to distinguish between ancestors/successors and should have a notion of distance on the graph. To cope with general graphs, instead of ancestor/successor nodes, we predict if a node is reachable, acknowledging the edge directions. As distance measures, we study the prediction of adjacent nodes as well as the \emph{directed} and \emph{undirected} shortest path distance. With \emph{undirected} shortest path distance, we refer to the path length on the symmetrized graph, and in both cases we ignore node pairs for which no path exists. In summary, we study pair-wise binary classification of \emph{(1) reachability} and \emph{(2) adjacency} as well as pair-wise regression of \emph{(3) undirected distance} and \emph{(4) directed distance}.

\textbf{Models.} We use a vanilla transformer encoder~\citep{vaswani_attention_2017} with positional encodings (see \autoref{fig:maglapnet_with_trans:a}). We compare our Magnetic Laplacian (ML) positional encodings w/o SignNet (\autoref{sec:spectral}) with our direction-aware random walk (RW) of \autoref{sec:random_walks} and eigenvectors of the combinatorial Laplacian (Lap.) from \autoref{sec:sinlap}. The eigenvectors of the combinatorial Laplacian are preprocessed like the ones of the Magnetic Laplacian (\autoref{fig:maglapnet_with_trans:b}), except that the ``Stack'' step is superfluous due to the real eigenvectors. Additionally, we compare to the SVD encodings of \citet{hussain_global_2022} that perform a low-rank decomposition of the (directed) adjacency matrix. Moreover, with the goal of obtaining general positional encodings, we do not study any heuristics that can be considered ``directional''. For example, if solely considering trees, it might be sufficient to add features for source and sink nodes next to undirected positional encodings.

All studied tasks are instances of link prediction or link regression where the predictions are of shape \(n \times n\) (ignoring the disconnected pairs of nodes in distance regression), modeling the relative interactions between nodes. For this, we broadcast the resulting node encodings \(\mH_l^{(n)}\) (see \autoref{fig:maglapnet_with_trans:a}) of the sender and receiver nodes and stack a global readout. Thereafter, we use a shallow MLP with 3-layers in total and task-dependent output activation (softmax or softplus).

\begin{figure}[t]
  \centering
  \includegraphics[width=\linewidth]{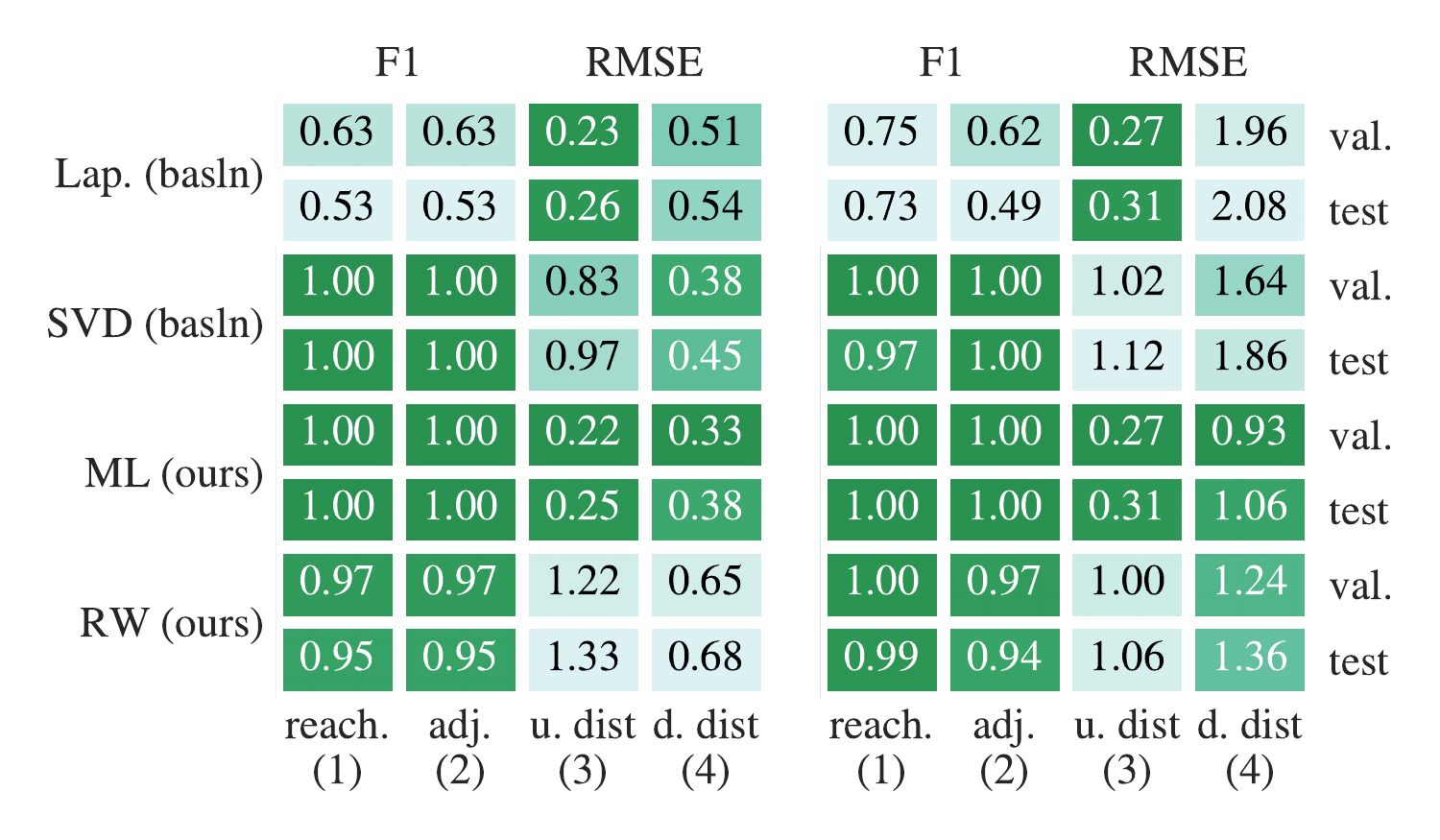}\\[-4ex]
  \subfloat[Directed Acyclic Graph\label{fig:playground:a}]{\hspace{0.5\linewidth}}
  \hfill
  \subfloat[Regular Directed Graph\label{fig:playground:b}]{\hspace{0.48\linewidth}}
  \caption{Positional encodings playground results. Dark green encodes the best scores and bright green bad ones. For F1 score high values are better and for RMSE low values.}
  \label{fig:playground}
  \vspace{-0.175in}
\end{figure}

\textbf{Setup.} We use cross-entropy for classification and \(\normltwo\) loss for regression. We assess classification with the F1 score and regression with the Root Mean Squared Error (RMSE). We sample Erd\H{o}s-R\`enyi graphs with equally probable average degree \(\{1, 1.5, 2\}\) and, additionally, Directed Acyclic Graphs (DAGs), where we draw the average degree from \(\{1, 1.5, 2, 2.5, 3\}\) to account for the greater sparsity. Then, we extract the largest (weakly) connected component. For the regression tasks, we sample graphs with 16 to 63, 64 to 71, and 72 to 83 nodes for train, validation, and test, respectively. To counteract a too-severe class imbalance, we choose 16 to 17, 18 to 19, and 20 to 27 nodes for the classification tasks, respectively. We report the average over three random reruns. We sample 400,000 training instances and for test/validation 2,500 for each number of nodes \(n\).

\textbf{Results.} In \autoref{fig:playground}, we show the performance of the positional encodings for the four curated tasks. We see that the eigenvectors of the Magnetic Laplacian outperform the eigenvectors of the combinatorial Laplacian for all measures that rely on directedness. For \emph{(3) undirected distance}, they perform similarly well. On the classification tasks \emph{(1)} \& \emph{(2)}, the random walk encodings perform comparably to the Magnetic Laplacian. However, random walk encodings are less predictive for regressing distances.  %
In general, random walks seem to show their strength for tasks that are well-aligned with their design. For example, a random walk with \(k=1\) resembles the adjacency matrix that we predict in task \emph{(2)}. However, the random walk encodings only achieve mediocre scores for \emph{(3) undirected distance} prediction.

The Magnetic Laplacian encodings consistently outperform the SVD encodings and achieve an up to 4 times lower RMSE. Nevertheless, the SVD encodings are a surprisingly strong baseline on the DAG (\autoref{fig:playground:a}), where they even outperform the random walk encodings. However, on general directed graphs (\autoref{fig:playground:b}), the random walk encodings outperform the SVD encodings on \emph{(4) directed distance} with a roughly 30\% lower RMSE. Moreover, we did not achieve similarly strong performance with SVD in the other studied tasks. For example, in the sorting network task (see \autoref{sec:sorting_networks}), we were not able to achieve meaningful performance after a basic hyperparameter search. This might be due to the undesirable properties low-rank SVD positional encodings have for certain graph structures (see \autoref{sec:appendix_maglap_svd}).

In \autoref{sec:appendix_playground}, we provide additional comparisons. These include (a) a comparison to GNNs and (b) the study of relative random walk encodings. For (b), we use the pair-wise encodings \(\zeta(v | u)\) before the node-level aggregation and add the \(n \times n \times d\) encodings to the attention matrix before applying the softmax. The relative random walk encodings can be understood as a generalization of the pair-wise shortest distances used by \citet{ying_transformers_2021}. Moreover, in \autoref{sec:appendix_rw_ablation}, we give a hyperparameter study of the random walk encodings and compare them to the (forward) random walk encodings of \citet{li_distance_2020} that are designed for undirected graphs.

\section{Application: Sorting Networks}\label{sec:sorting_networks}

\begin{figure}[b]
    \centering
    \hspace{5pt}
    \subfloat[Sorting Network\label{fig:sorting_graphs:a}]{
        \resizebox{0.35\linewidth}{!}{
        \centering
        \begin{tikzpicture}
        \foreach \a in {0,...,4}
          \draw[thick] (0,\a+1) node[left]{\a} -- ++(11/3,0);
        \foreach \p/\x/\y in {{1/4/5}, {2/2/3}, {3/2/4}, {4/1/2}, {5/2/5}, {6/2/3}, {7/2/4}, {9/3/5}, {10/3/4}}
        {
          \filldraw (\p/3, \x) circle (1.5pt);
          \filldraw (\p/3, \y) circle (1.5pt);
          \draw[thick] (\p/3, \x) -- (\p/3, \y);
        }
        \end{tikzpicture}
        }}
    \hspace{15pt}
    \subfloat[Directed Graph\label{fig:sorting_graphs:b}]{
      \includegraphics[width=0.35\linewidth]{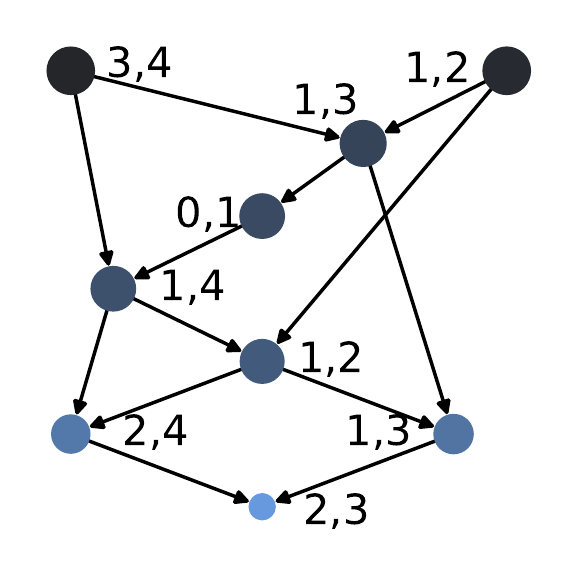}}
    \caption{(a) Common illustration for a sorting network with sequence length \(p=5\) and (b) as directed graph.}
    \label{fig:sorting_graphs}
\end{figure}

\begin{figure*}
    \centering
    \begin{minipage}{0.1725\linewidth}
        \centering
        \vspace{21pt}
        \includegraphics[width=\linewidth]{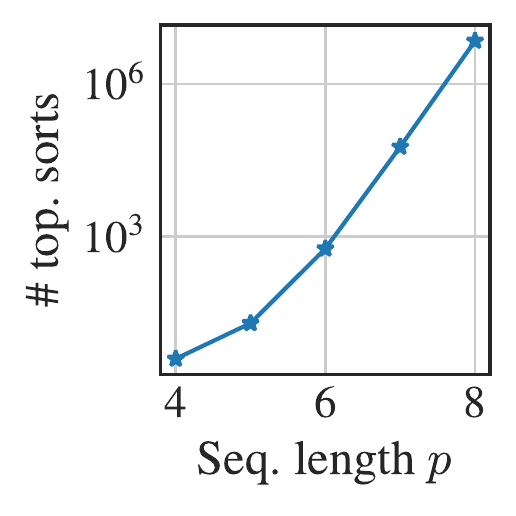}
        \caption{\# topological sorts for Batcher even odd mergesort.}
        \label{fig:sorting_batcher_n_topsorts}
    \end{minipage}
    \hfill
    \begin{minipage}{0.8125\linewidth}
        \includegraphics[width=1.025\linewidth]{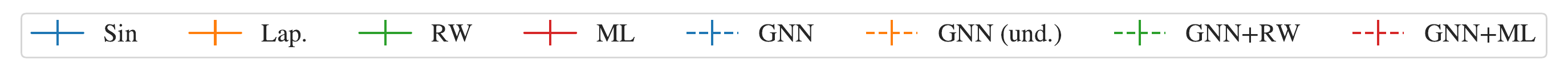}\\
        \begin{minipage}{0.235\linewidth}
          \centering
          \includegraphics[width=\linewidth]{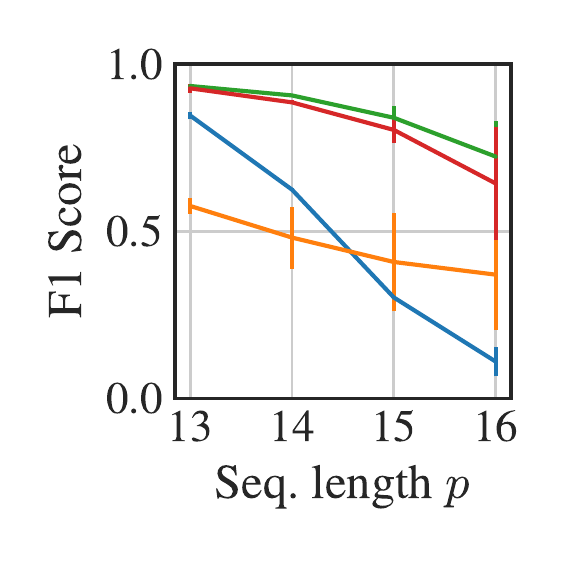}
          \caption{Comparing positional enc. over length \(p\) (sorting netw.).}
          \label{fig:sn_main}
        \end{minipage}
        \hfill
        \begin{minipage}{0.235\linewidth}
          \centering
          \includegraphics[width=\linewidth]{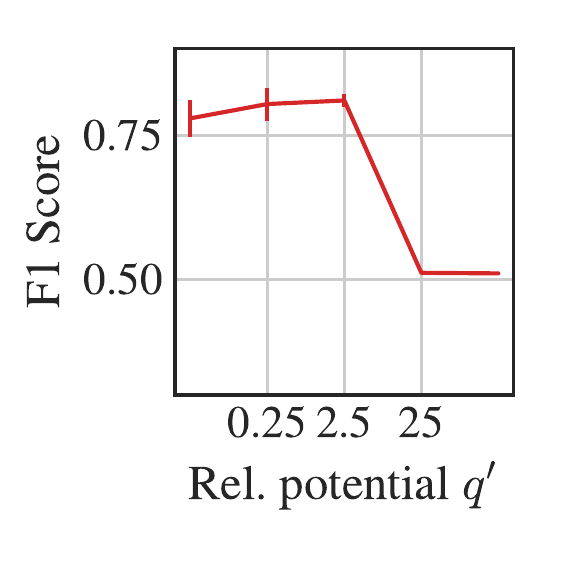}
          \caption{Relative pot. \(q = \nicefrac{q'}{\max(\min(\vec{m}, n), 1)}\) with \(k=25\) eigenvec.}
          \label{fig:sn_qsweep}
        \end{minipage}
        \hfill
        \begin{minipage}{0.49\linewidth}
            \subfloat[\# mess. pass. steps\label{fig:sn_gnn:a}]{\includegraphics[width=0.48\linewidth]{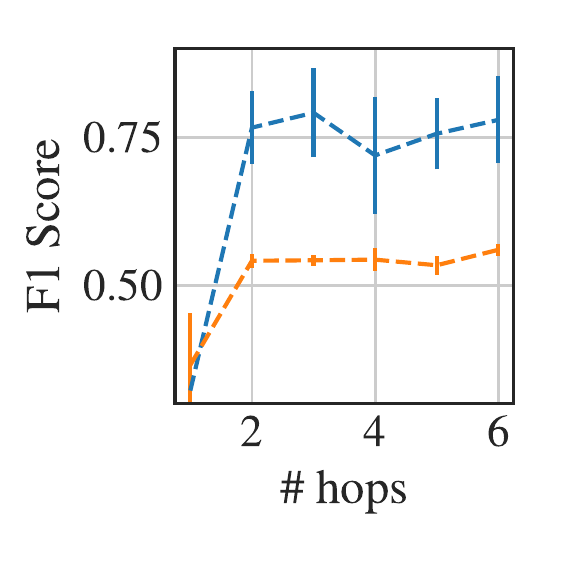}}
            \hfill
            \subfloat[GNN only\label{fig:sn_gnn:b}]{\includegraphics[width=0.48\linewidth]{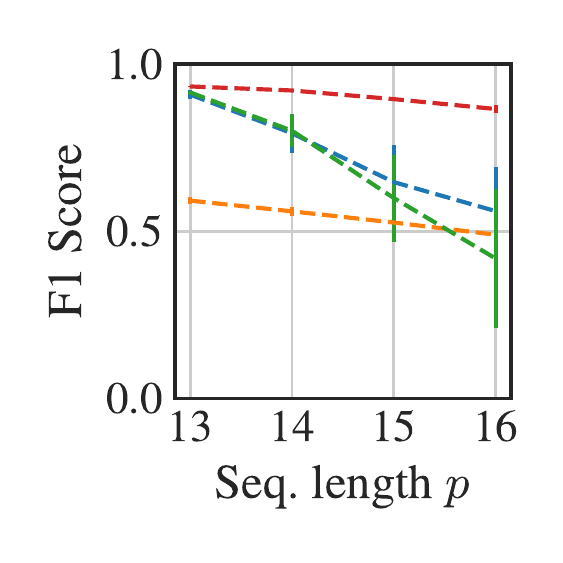}}
            \hfill
            \caption{(a) \# message passing steps of GNN and (b) for GNN w/ and w/o pos.\ enc.}
            \label{fig:sn_gnn}
        \end{minipage}
    \end{minipage}
\end{figure*}

Sorting networks~\citep{knuth_art_1973} are a certain class of comparison-based algorithms that have the goal of sorting any input sequence of fixed size with a static sequence of comparators. Sorting networks are a particularly interesting application since they mark the middle ground between logical statements and source code. Specifically, the sequence of comparators reflects a sequence of program instructions while asserting their correctness is related to satisfiability~\cite{knuth_art_1968}. We use this task to make the implications of symmetrization (Laplacian encodings) and sequentialization (sinusoidal encodings) explicit.

We consider sorting networks that consist of a sequence of conditional exchange instructions. 
In \autoref{fig:sorting_graphs:a}, the \(p\) horizontal lines represent the variables that are to be sorted, and the vertical lines are comparators that sort the connected variables. Thus, a sorting network can also be expressed by \(n\) \verb|v_i, v_j = sorted((v_i, v_j))| statements, where \verb|v_i| and \verb|v_j| are two of the \(p\) variables, i.e.\ \(i, j \in \{0, 1, \dots, p-1\}\). Our graph construction (\autoref{fig:sorting_graphs:b}) treats every instruction as a node with \(i\) and \(j\) as features (sinusoidal encoding). If a node operates on indices \(i\) and \(j\), we add an edge from the last occurrences of \(i\) and \(j\) (if there are any). Thus, in this data-flow graph, the in- and outdegree equal two for all nodes except source and sink nodes. 

\textbf{Directed graph vs. sequence.} An important implication is that each topological sort of the directed graph is an equivalent ``program'', i.e., a different ordering of statements that yields the same result.
In \autoref{fig:sorting_batcher_n_topsorts}, we show the number of topological sorts over the sequence lengths \(p\) for a type of compact and deterministically constructed sorting network. For such networks and a sequence length of just 8, the number of equivalent sequentializations already exceeds 1 million (see also \autoref{sec:appendix_sorting_networks_near_seq}). Note that in the worst case, a directed graph has \(n!\) topological sorts. Therefore, representing directed graphs as sequences can introduce a huge amount of arbitrary orderedness. In contrast to sequential modeling, a graph-based representation can significantly reduce the size of the effective input space.

\textbf{Symmetrization hurts.} There exist correct and incorrect sorting networks that map to the same undirected graph. Hence, a model that uses undirected graphs cannot distinguish these cases. For example, the \emph{correct sorting network} for length three with comparators \([(0, 2), (0, 1), (1, 2)]\) and its reversed version (\emph{incorrect}) map to the same undirected graph. In summary, symmetrizing may hurt expressiveness.

\textbf{Dataset.} We construct a dataset consisting of 800,000 training instances for equally probable sequence lengths \(7 \le p_{\text{train}} \le 11\), generate the validation data with \(p_{\text{val}} = 12\), and assess performance on sequence lengths \(13 \le p_{\text{test}} \le 16\). We construct the sorting networks greedily until we have a correct sorting network. For this, we draw a random pair of comparators, excluding immediate duplicates and comparators between inputs that are already sorted. We then generate an incorrect example by omitting the last comparator (i.e., train is balanced). This procedure is similar to datasets for the related task of satisfiability~\citep{selsam_learning_2019}. Moreover, we add additional incorrect sorting networks by reversing the directions of the correct networks to make the test sets more challenging. 
Thus, the test and validation data consist of \(\nicefrac{1}{3}\) correct sorting networks (20,000) and \(\nicefrac{2}{3}\) of incorrect ones (40,000). Therefore, the task is to generalize the correctness prediction to longer sequences and reversed (slightly out of distribution) sorting networks. See \autoref{sec:appendix_sorting_networks} for more details on the dataset construction.

\textbf{Empirical Evaluation.} We follow the setup of \autoref{sec:playground} and include sinusoidal positional encodings (Sin). As shown in \autoref{fig:sn_main}, the eigenvectors of the Magnetic Laplacian (ML) perform comparably to the random walk encodings. Both outperform the other positional encodings by a large margin. This shows that, without bells and whistles, the Magnetic Laplacian and random walk encodings provide a transformer a considerable structural awareness for directed graphs (see \autoref{fig:maglapnet_with_trans:a}). On the other hand, the eigenvectors of the combinatorial Laplacian barely outperform the na\"ive baseline that randomly chooses a class based on the prior probabilities. 

\textbf{Sinusoidal positional encodings} perform well for sequences close to the training data but do not generalize well to longer sequences. The lacking generalization might be due to the much larger input space if measuring input space size in terms of unique inputs for the transformer.

\begin{figure*}
\vskip 0.025in
    \setlength{\fboxsep}{0pt}
    \hspace{-25pt}
    \begin{minipage}{0.3\textwidth}
        \centering
        \hspace{12pt}\textbf{Prediction:} \verb|unknown|
        \begin{minted}[fontsize=\scriptsize, escapeinside=@@]{python}
        def f1_score(pred, label):
          correct = pred == label
          tp = (correct & label).sum()
          fn = (~correct & pred).sum()
          fp = (~correct & ~pred).sum()
          precision = tp / (tp + fp)
          @\colorbox{yellow}{recall = tp / (tp + fn)}@
          return (
            2 * (@\colorbox{skyblue}{recall}@ * @\colorbox{green}{precision}@) / 
            (@\colorbox{skyblue}{recall}@ + @\colorbox{green}{precision}@)
          )
        \end{minted}
    \end{minipage}
    \hfill{}
    \begin{minipage}{0.3\textwidth}
        \centering
        \hspace{17pt}\textbf{Prediction:} \verb|accuracy|
        \begin{minted}[fontsize=\scriptsize, escapeinside=@@]{python}
        def f1_score(pred, label):
          correct = pred == label
          tp = (correct & label).sum()
          fn = (~correct & pred).sum()
          fp = (~correct & ~pred).sum()
          precision = tp / (tp + fp)
          @\colorbox{yellow}{recall = tp / (tp + fn)}@
          return (
            2 * (@\colorbox{green}{precision}@ * @\colorbox{skyblue}{recall}@) / 
            (@\colorbox{green}{precision}@ + @\colorbox{skyblue}{recall}@)
          )
        \end{minted}
    \end{minipage}
    \hfill{}
    \begin{minipage}{0.3\textwidth}
        \centering
        \hspace{22pt}\textbf{Prediction:} \verb|precision|
        \begin{minted}[fontsize=\scriptsize, escapeinside=@@]{python}
        def f1_score(pred, label):
          correct = pred == label
          tp = (correct & label).sum()
          fn = (~correct & ~pred).sum()
          @\colorbox{yellow}{recall = tp / (tp + fn)}@
          fp = (~correct & pred).sum()
          precision = tp / (tp + fp)
          return (
            2 * (@\colorbox{green}{precision}@ * @\colorbox{skyblue}{recall}@) / 
            (@\colorbox{green}{precision}@ + @\colorbox{skyblue}{recall}@)
          )
        \end{minted}
    \end{minipage}
    \hspace{20pt}
    \caption{State-of-the-art model on OGB Code2 is susceptible to meaningless permutations (see highlighted text) due to OGB Code2's graph construction. %
    The code was minimally modified for better layout.}
    \label{fig:f1score}
\end{figure*}

\textbf{GNN.} In \autoref{fig:sn_gnn:a}, we study the performance of a GNN (following \citet{battaglia_relational_2018}, see \autoref{sec:source_code} \&  \autoref{sec:appendix_setup} for specifics) with mean readout over the number of message passing steps. Since the improvements diminish for more than two message passing steps, we report the results in \autoref{fig:sn_gnn:b} using three message passing steps. 
We compare a direction-aware ``GNN'' and direction unaware ``GNN (und.)''. Expectedly, the directional information is important for generalization also in the context of GNNs. Note that the directional GNN performs on par with a transformer encoder with the Magnetic Laplacian encodings for sequence length 13. However, the GNN generalizes slightly worse to longer sequences. Motivated by \citet{li_distance_2020}, we additionally pair the GNN with positional encodings and find that the Magnetic Laplacian eigenvectors can also help a GNN's generalization. On the other hand, the random walk encodings struggle to generalize to longer sequences and harm performance. We hypothesize that the Magnetic Laplacian encodings provide complimentary information while the random walk encodings are similar to message passing.

\section{Application: Function Name Prediction}\label{sec:source_code}

We study function name prediction since it is an established task in the graph learning community~\cite{hu_open_2020} where the direction of edges influences the true label, and transformers seem to have an edge over GNNs. Similar to sorting networks, each program represents a specific ordering of statements, and there can be many equivalent programs via reordering statements. Thus, it is surprising that graphs for source code used for machine learning retain the sequential connections between instructions. In other words, these graphs ``only'' enrich sequential source code (here, add a hierarchy). For example, the Open Graph Benchmark Code2 dataset represents the 450,000 functions with its Abstract Syntax Tree (AST) and \emph{sequential connections}. 
Since the input space of sequences can be much larger than the input space of directed graphs (see \autoref{sec:sorting_networks}), for some tasks, such a graph construction is an unfortunate choice.

\textbf{Robustness.} We trained the state-of-the-art model\footnote{Shortly before submission, \citet{luo_dagformer_2022} proposed in their preprint a new model with 0.4\% higher F1 test score.} on the Open Graph Benchmark Code2 dataset, called Structure Aware Transformer (SAT)~\citep{chen_structure-aware_2022}. We then used OGB's code to generate multiple graph representations of functions, where we reordered some statements s.t.\ the functionality is preserved. In \autoref{fig:f1score}, we show that the state-of-the-art model using OGB's graph construction is susceptible to these semantics-preserving permutations of the source code. Moreover, the number of possible reorderings can be surprisingly high. E.g., if constructing a data-flow DAG, the F1 score function of \autoref{fig:f1score} has 16 topological sorts. Further considering commutativity for  \verb|==|, \verb|&|, \verb|+|, and \verb|*|, we find 4,096 possibilities to write this seemingly simple function.

\textbf{Our graph construction} maps all these 4,096 possibilities to the very same directed graph. Our graph construction is greatly inspired by \citet{bieber_library_2022}, although they also connect most instructions sequentially. While we do avoid this sequentialism, we leverage their static code analysis for a graph construction that handles the sharp bits like if-else, loops, and exceptions. The most significant differences to \citet{bieber_library_2022} are: (a) We construct a DAG for each ``block'' (e.g., body of if statement) that reflects the dependencies between instructions and then connect the statements between blocks considering the control flow; %
(b) we address the commutative properties for basic Python operations via edge features (up to the usage of, e.g., two binary operations to add three terms); (c) we do not reference the sequence of tokenized source code; (d) we omit the (in our case) unnecessary ``last read'' edges; (e) we construct the graph similarly to OGB Code2 for comparability. For example, we aggregate child nodes containing only attributes into their parent's node attributes. We provide details and a side-by-side comparison to an OGB Code2 graph in \autoref{sec:appendix_source_code}.

\textbf{Assumptions.} While the right equi-/invariances are task-dependent, we argue that for high-level reasoning tasks, including function name prediction or correctness prediction, the mentioned reorderings should not affect the true label. Nevertheless, e.g., for predicting the runtime of a program, reorderings can have an impact. Moreover, we assume that non-class-member methods are side-effect-free. For example, this includes reordering print statements. Even though this will result in a different output stream, we argue that these differences are typically not vital. Moreover, since we construct the graph with lexicographical static code analysis, we do this on a best-effort basis and do not capture all dynamic runtime effects. Last, our eigenvector-based positional encodings are only permutation equivariant in the absence of repeated eigenvalues (see \autoref{sec:appendix_magnetic_laplacian} for details).

\textbf{Empirical Evaluation.} In \autoref{tab:ogb_main}, we report the results on OGB Code2. Here we also compare to the Structure Aware Transformer (SAT) of~\citet{chen_structure-aware_2022}. SAT is a hybrid transformer w/ GNN for query and key and was the prior state of the art. We illustrate the architecture in \autoref{fig:appendix_architecture:b}. If we omit the GNN, we recover the vanilla transformer encoder \autoref{fig:maglapnet_with_trans:a} (plus degree-sensitive residual). We improve the current state-of-the-art model with a number of small tricks (i.e., no new positional encoding yet). Our SAT++ (w/ GNN) improves the F1-score by 1.66\% (relatively 8.6\%). Besides smaller changes like replacing ReLU with GeLU activations, we most notably (1) add dropout on the sparsely populated node attributes and (2) offset the softmax score to adjust for the class imbalance of the special tokens for \emph{unknown words} as well as \emph{end of sequence}. We also replace the GCN with a three-layer GNN following \citet{battaglia_relational_2018} (w/o global state). The edge and node embeddings are updated sequentially with independently aggregated forward and backward. Then, we concatenate the embeddings we obtained after each message passing step and apply an MLP with two layers. For details on the GNN see \autoref{sec:appendix_setup}.

\textbf{Our graph construction} (``data-flow'' in \autoref{tab:ogb_main}) consistently increases the predictive performance. We do not report results w/o GNN and solely w/ AST depth positional encodings because this approach does not make use of the enhanced graph structure. Our graph construction raises the F1 score by almost 0.58\% (relatively 2.8\%), if using the SAT++ architecture (w/ GNN) with AST depth encodings. Note that the gain partially stems from the improved edge features. In a dedicated experiment, we compare the effect of our data-flow edges with the sequential edges of \citet{bieber_library_2022} and find that our edges contribute to an \(\approx0.1\%\) greater F1 score (model uses Magnetic Laplacian encodings). Nevertheless, we want to emphasize that our graph construction yields robustness gains w.r.t.\ certain reorderings of statements in the source code (see \autoref{fig:f1score}). 

\begin{table}[t]
    \vskip 0.1in
    \caption{Results on the Open Graph Benchmark Code2 dataset. The first two rows correspond to prior work. All other approaches are our contribution. %
    We report the average and error of the mean over 10 reruns. Best is bold.}
    \centering
    \resizebox{\linewidth}{!}{
    \begin{tabular}{llcccc}
        \toprule
                             & \textbf{Position. Enc.} & \textbf{GNN}       &          \textbf{Test F1-Score} &          \textbf{Val. F1-Score} \\
        \midrule
        \multirow{4}{*}{\rotatebox{90}{\textbf{Sequen.}}} & \multirow{4}{*}{AST depth} & \xmark &  16.70$\pm$0.05 &  15.46$\pm$0.06 & \multirow{2}{*}{\rotatebox{90}{\textbf{Prev.}}}\\
                                                                &               & \cmark &  19.37$\pm$0.09 &  17.73$\pm$0.07 \\
        \cline{3-6}\cline{3-6}\cline{3-6}\cline{3-6}\cline{3-6}
        &  & \xmark &  19.09$\pm$0.10 &  17.68$\pm$0.06 & \multirow{7}{*}{\rotatebox{90}{\textbf{Ours}}}  \\
        &  & \cmark &  21.03$\pm$0.07 &  19.38$\pm$0.07   \\
        \cline{1-5}
        \multirow{5}{*}{\rotatebox{90}{\textbf{Data-flow}}} & AST depth  %
                                                        & \cmark &  21.61$\pm$0.12 &  19.79$\pm$0.11 \\
        \cline{2-5}
        & \multirow{2}{*}{Random walk} & \xmark &  19.34$\pm$0.08 &  \textbf{17.96$\pm$0.05} \\
        & & \cmark & 21.82$\pm$0.20 &  20.03$\pm$0.17 \\
        \cline{2-5}
        & \multirow{2}{*}{Magnetic Lap.} & \xmark &  \textbf{19.43$\pm$0.03} &  17.83$\pm$0.05 \\
        &               & \cmark &  \textbf{22.22$\pm$0.10} &  \textbf{20.44$\pm$0.06} \\
        \bottomrule
    \end{tabular}
    }
    \label{tab:ogb_main}
    \vskip -0.1in
\end{table}

\textbf{Hybrid.} The Magnetic Laplacian also helps in the hybrid transformer GNN architecture. \emph{Our SAT++ with Magnetic Laplacian positional encodings (SignNet w/ GNN) marks the new state of the art on the Code2 dataset, outperforming SAT by 2.85\% (relatively 14.7\%)}. The Random Walk positional encodings only slightly improve performance. For the Code2 graphs, the GNN for query and key appears to be of great importance. We hypothesize that this is due to the sparsely populated node features. Only a few nodes are attributed, and additionally, the permitted vocabulary is restrictive. The local message passing might spread the information to neighboring nodes to adjust for this sparseness. Moreover, w/o GNN, we do not make use of edge features.

\textbf{Dataset challenges.} The node attributes (e.g., variable names) and function names are only lightly preprocessed. For example, for perfect performance, one needs to distinguish singular and plural method names. Although singular/plural semantically makes a difference, the naming consistency is lacking for the 450k functions taken from github. %
We refrain from adjusting the dataset accordingly to maintain comparability to prior work.

\FloatBarrier
\section{Related Work}\label{sec:related_work}

Directed graphs appear in various applications and can be crucial to appropriately model the input data, also in well-established domains for GNNs such as citation networks~\citep{rossi_edge_2023}. An important related GNN for directed graphs is MagNet~\citep{zhang_magnet_2021} since it used the Magnetic Laplacian within its message passing. We compare to MagNet in \autoref{sec:appendix_playground} on the playground tasks.

\textbf{Positional encodings.} Prior work on positional encodings includes traditional graph metrics, like shortest path distances~\citep{guo_graphcodebert_2021}. Related to the distance from a node to the AST root node in the OGB Code2 dataset (see \autoref{sec:source_code}), \citet{luo_dagformer_2022} proposes sinusoidal positional encodings for DAGs leveraging their partial order. An alternative form of spectral encodings, based on Singular Value Decomposition (SVD), was used for positional encodings~\citep{hussain_global_2022}. The authors argue that these encodings also include directed graphs; however, they do not verify this choice, and the SVD of the adjacency matrix has undesirable properties (see \autoref{sec:appendix_maglap_svd}). We compare the SVD encodings in \autoref{sec:playground} on the tasks of the positional encodings playground. 

We include a discussion of Laplacians for directed graphs in \autoref{sec:appendix_laplacians_directed_graphs}. For an in-depth overview and a how-to for graph transformers, we refer to \citet{min_transformer_2022}, \citet{muller_attending_2023} and \citet{rampasek_recipe_2022}. They also provide an overview of graph transformers that rethink attention architectures for structure-awareness like~\citep{dwivedi_generalization_2021, mialon_graphit_2021, chen_structure-aware_2022, kim_pure_2022, hussain_global_2022, diao_relational_2022}.

\textbf{Graph construction.} Many works~\citep{allamanis_learning_2018, cummins_programl_2020, guo_graphcodebert_2021, bieber_library_2022} enrich source code in a graph-structured manner for machine learning. However, they all retain the sequentialism of the underlying source code. As we see in \autoref{fig:f1score}, this can lead to a fragile representation w.r.t.\ to semantically meaningless reorderings. Such reorderings are a novel perspective on the robustness of models for code~\citep{yefet_adversarial_2020,bielik_adversarial_2020,jha_codeattack_2022,ramakrishnan_semantic_2022}. However, the relationship between a directed graph and its sequentializations is well-known in task scheduling. Similarly, an appropriate graph construction may improve the robustness of transformers if applied to other combinatorial optimization problems~\cite{ geisler_generalization_2022,kool_attention_2019} than correctness prediction of sorting networks.

\section{Conclusion}\label{sec:conclusion}

We propose positional encodings for directed graphs based on the Magnetic Laplacian and random walks. Both positional encodings can help transformers to gain considerable structure awareness and show complementary strengths in our experiments. We argue that direction-aware positional encodings are an important step towards true multi-purpose transformers universally handling undirected and directed graphs. We show that directedness can be central for the semantics in the target domain and that directed graphs can drastically lower the effective input dimensionality (i.e., many instances map to one graph).

\section*{Acknowledgements}

We thank Kim Stachenfeld, Dimitrios Vytiniotis, Shariq Iqbal, Andrea Michi, Marco Selvi, Daniel Herbst, and Jan Schuchardt for the feedback at various stages of this work.
\newline
This research was supported by the Helmholtz Association under the joint research school “Munich School for Data Science -- MUDS“.

\bibliography{references}
\bibliographystyle{icml2023}

\newpage
\appendix
\onecolumn

\counterwithin{figure}{section}
\counterwithin{table}{section}
\counterwithin{equation}{section}
\counterwithin{algorithm}{section}

\appendix

\section{Example Graphs}\label{sec:appendix_example_graphs}

Additionally to the graphs in \autoref{fig:example_graphs}, here we give further examples (illustrated in \autoref{fig:appendix_example_graphs}). These examples, are used for the more elaborate discussion in the appendix. The construction of most graphs should be self-explanatory, even if we increase the number of nodes. We next provide necessary details. For the two disconnected sequences \autoref{fig:appenidx_example_graphs:i}, we split the sequence after the first \(\lfloor \nicefrac{n}{2} \rfloor\) tokens/nodes. All ``trumpet graphs'' (\autoref{fig:appenidx_example_graphs:d}, \ref{fig:appenidx_example_graphs:h}, and \ref{fig:appenidx_example_graphs:l}) connect the nodes (between) \(\lfloor \nicefrac{3n}{10} \rfloor\) and \(\lfloor \nicefrac{7n}{10} \rfloor\). For \autoref{fig:appenidx_example_graphs:k}, we first construct a fully connected DAG plus self-loops (entries of main diagonal and above are all one). Then, we add the reversed edges for the inner 50\% of nodes (\(\lfloor \nicefrac{n}{4} \rfloor\) to \(\lfloor \nicefrac{3n}{4} \rfloor\)).

\begin{figure}[H]
  \centering
  \makebox[\linewidth][c]{
    \(
    \begin{array}{ccccc}
      \subfloat[Sequence\label{fig:appenidx_example_graphs:a}]{
      \includegraphics[width=0.18\linewidth]{assets/spectrum/small_sequence_graph_0.pdf}} &
      \subfloat[Undirected sequence\label{fig:appenidx_example_graphs:b}]{
      \includegraphics[width=0.18\linewidth]{assets/spectrum/small_undirected_sequence_graph_0.pdf}} &
      \subfloat[Balanced binary tree\label{fig:appenidx_example_graphs:c}]{
      \includegraphics[width=0.18\linewidth]{assets/spectrum/small_reversed_balanced_binary_tree_graph_0.pdf}} &
      \subfloat[Trumpet\label{fig:appenidx_example_graphs:d}]{
      \includegraphics[width=0.18\linewidth]{assets/spectrum/small_trumpet_loop_graph_0.pdf}} \\
      
      \subfloat[Reversed sequence\label{fig:appenidx_example_graphs:e}]{
      \includegraphics[width=0.18\linewidth]{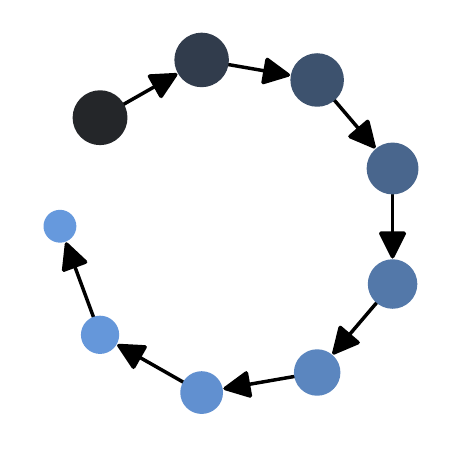}} &
      \subfloat[Cirlce\label{fig:appenidx_example_graphs:f}]{
      \includegraphics[width=0.18\linewidth]{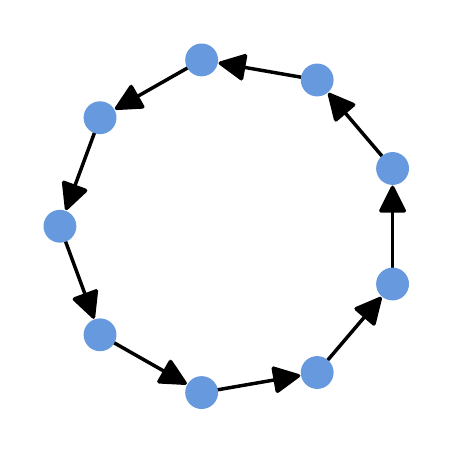}} &
      \subfloat[Reversed bal.\ bin.\ tree\label{fig:appenidx_example_graphs:g}]{
      \includegraphics[width=0.18\linewidth]{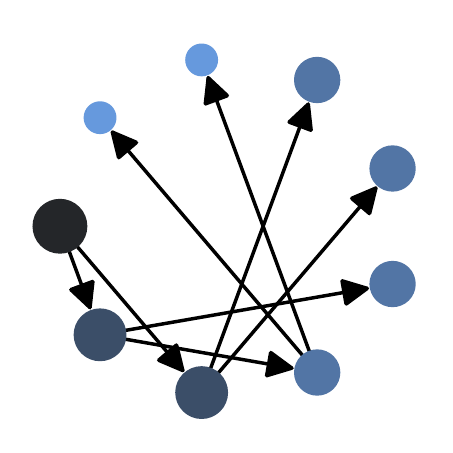}} &
      \subfloat[Trumpet (DAG)\label{fig:appenidx_example_graphs:h}]{
      \includegraphics[width=0.18\linewidth]{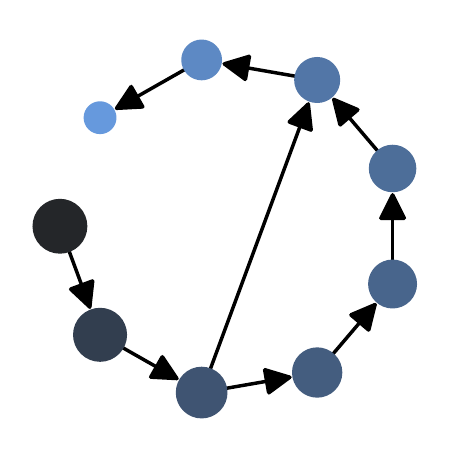}} \\
      
      \subfloat[Disconnected seq.\label{fig:appenidx_example_graphs:i}]{
      \includegraphics[width=0.18\linewidth]{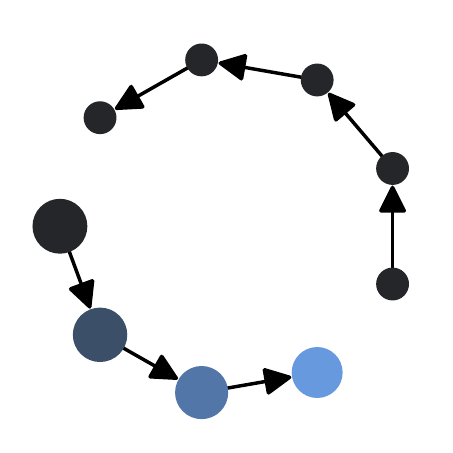}} &
      \subfloat[Fully connected DAG]{
      \includegraphics[width=0.18\linewidth]{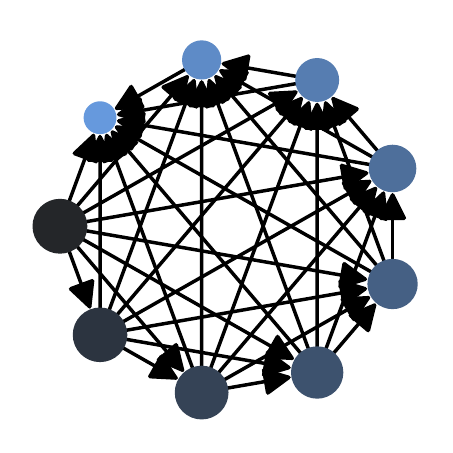}} &
      \subfloat[Mix DAG \& fully con.\label{fig:appenidx_example_graphs:k}]{
      \includegraphics[width=0.18\linewidth]{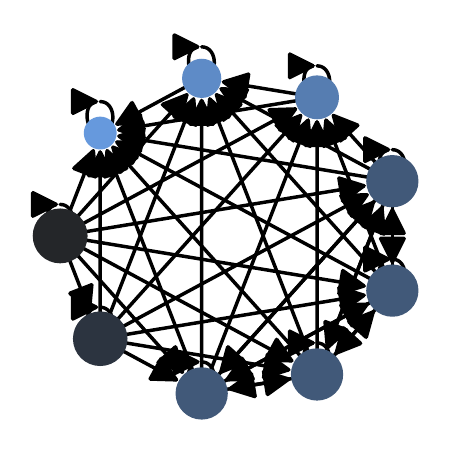}} &
      \subfloat[Trumpet (fully con.)\label{fig:appenidx_example_graphs:l}]{
      \includegraphics[width=0.18\linewidth]{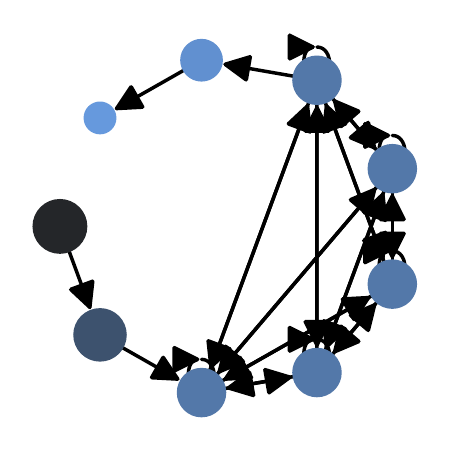}} \\
    \end{array}
    \)
  }
  \caption{First eigenvector of Magnetic Laplacian. Node size encodes the real value and color the imaginary value. %
  }
  \label{fig:appendix_example_graphs}
\end{figure}

\section{Graph Fourier Transformation}\label{sec:appendix_graph_fourier_transformation}

Similarly to the Discrete Fourier Transformation (DFT) \autoref{eq:dft}, a Graph Fourier Transformation (GFT) for \emph{undirected} graphs can be defined with the eigenvectors \(\Eigvec\) of the combinatorial Laplacian (\autoref{eq:laplacian_u} or \autoref{eq:laplacian_s}). Here the graph signal \(\vx \in \R^n\) is mapped to the frequency domain as
\begin{equation}
    \mX \coloneqq \Eigvec^\top \vx
\end{equation}
or for a specific frequency/eigenvalue
\begin{equation}
    \emX(\eigvec_k) = \emX_{k} = \sum_{i=1}^n \evx_i \eEigvec_{i, k}
\end{equation}
The inverse transform is then given as \(\vx = \Eigvec \mX\). Here we assume \(\Eigvec\) to be orthonormal (\(\Eigvec \Eigvec^\top = \mI\)). In analogy to sequences, the lowest eigenvalues and eigenvectors reflect the low frequencies. For undirected and connected graphs, the first (also called trivial) eigenvector is constant across nodes: \(\eEigvec_{u, 0} = \pm \nicefrac{1}{\sqrt{n}}\). This primitive is apparent from the quadratic form of the Laplacian
\begin{equation}\label{eq:appendix_quadratic_form_lap}
    \vx^\top \mL_U \vx = \frac{1}{2} \sum_{(u,v) \in E} (\evx_u - \evx_v)^2 = \frac{1}{2} \sum_{u=1}^n \sum_{v=1}^n \emA_{u,v} (\evx_u - \evx_v)^2
\end{equation}
that is minimized by its smallest eigenvector \(\eigvec_0^\top \mL_U \eigvec_0 = \eigval_0 = 0\). We will come back to a similar relation for the smallest eigenvector of the Magnetic Laplacian in \autoref{sec:appendix_magnetic_laplacian}.

\textbf{Graph convolution.} Particularly important for graph signal processing is the convolution defined on graphs
\begin{equation}\label{eq:appendix_graph_convolution}
    g_\theta \circledast \vx = \Eigvec g_\theta(\Eigval) \underbrace{\Eigvec^\top \vx}_{\text{GFT}} = \Eigvec \underbrace{g_\theta(\Eigval) \mX}_{\text{filter }\times\text{ signal}} = \underbrace{\Eigvec \hat{\mX}}_{\text{inverse GFT}}
\end{equation}
where \(g_\theta(\Eigval)\) is a diagonal matrix parameterized by \(\theta\) that can be understood as a function of the eigenvalues \(\Eigval\) and represents the filter in the frequency domain. For more details see \citet{defferrard_convolutional_2017} or \citet{furutani_graph_2020}.

\textbf{Applications in machine learning.} The eigenvectors of the combinatorial Laplacian are widely used in graph machine learning. For example, they are the workhorse in spectral clustering~\citep{von_luxburg_tutorial_2007} where one can gain helpful insights if dealing with disconnected graphs. Moreover, the Laplacian eigendecomposition also stands at the core of many Graph Neural Networks. For example, ChebyNet~\citep{defferrard_convolutional_2017} approximates \(g_\theta(\Eigval)\) (\autoref{eq:appendix_graph_convolution}) via \(k\)-order polynomial in the spatial domain and, similarly, the popular Graph Convolutional Network  \citep{kipf_semi-supervised_2017} uses an alike linear approximation. For more background, we refer to \citet{bronstein_geometric_2021}.

\section{Laplacians for Directed Graphs}\label{sec:appendix_laplacians_directed_graphs}

It is well known that the combinatorial Laplacian is a discretization of the Laplace-Beltrami operator on Riemannian manifolds \citep{belkin_laplacian_2003}. This insight allows for many connections, including the negligence of direction. In short, here we only encode distances on a manifold where the order of start and end point is irrelevant. In other words, the eigenvectors of the combinatorial Laplacian form an \emph{isotropic} transformation.

\textbf{Combinatorial Laplacian w/o symmetrization.} For directed graphs the Laplacian, e.g., \(\mL = \mD - \mA\) is in general not symmetric. Hence, the eigenvectors as well as eigenvalues are potentially complex, \(\mL\) is not necessarily diagonalizable, and the resulting eigenvectors might not form an orthonormal basis. Especially, the latter criterion is important s.t.\ the different signals do not interfere with each other and that we can go back and forth from the spatial to the frequency domain without complications. Specifically, the eigenvectors might neither be \emph{unitary} \(\bar{\Eigvec}^\top \Eigvec = \mI\) or there might not even be a basis of eigenvectors.

We plot the first 5 eigenvectors of the combinatorial Laplacian without symmetrization in the right column of \autoref{fig:appendix_example_graphs_spectra_part1} and \ref{fig:appendix_example_graphs_spectra_part2} for all the graphs in \autoref{fig:appendix_example_graphs}. It is clear that these eigenvectors are well-suited for positional encodings of nodes for many of the graph topologies. For example, if we would use \(\bar{\Eigvec}^\top\) or \(\Eigvec\) in the GFT or its inverse, respectively, we would potentially ignore the signal of some nodes. Equivalently, a positional encoding would assign the  the zero vector \(\mathbf{0}\) to these nodes. 

\textbf{Magnetic Laplacian.} Various alternatives were proposed that generalize (or substitute) the eigenvectors for the  Laplacian to directed graphs. So far, we have only discussed the Magnetic Laplacian (\autoref{sec:spectral} and \ref{sec:appendix_magnetic_laplacian}) that bridges the gap using complex values, i.e., using a Hermitian matrix for which eigenvectors again form an orthogonal basis. We can use these eigenvectors for a GFT that includes directed graphs~\citep{furutani_graph_2020}. Such a transformation, where the direction does matter, is called \emph{anisotropic}.

\textbf{Jordan Decomposition.} Other approaches include the use of generalized eigendecomposition, like the Jordan decomposion~\citep{sandryhaila_discrete_2013}. However, here we are still left with a potentially non-orthogonal set of eigenvectors, and, most importantly, in this case, the ``low frequencies'' do not necessarily change smoothly over the nodes in the graph~\citep{singh_graph_2016}.

\textbf{Eigenfunctions of random walk operator.} \citet{sevi_harmonic_2021}, on the other hand, use the Dirichlet energy of eigenfunctions of a random walk operator. However, this approach is only applicable to strongly connected graphs and comes with restrictions related to the orthogonality of the Fourier basis.

\textbf{Optimization.} Last, (non-convex) optimization problems with constraints were proposed that typically minimize the Lovász extension of the graph cut size or local variations of the graph signal. Both directions additionally impose orthogonality constraints. A good entry point for the literature can be found in~\citep{marques_signal_2020}. It is worth noting though, that these approaches typically do not preserve all information s.t.\ we can reconstruct the graph structure~\citep{furutani_graph_2020}. Hence, it is not clear if they are well-suited positional encodings.

\section{Magnetic Laplacian}\label{sec:appendix_magnetic_laplacian}

We next give more details on the Magnetic Laplacian~\citep{forman_determinants_1993, shubin_discrete_1994, de_verdiere_magnetic_2013,berkolaiko_nodal_2013,fanuel_magnetic_2016,fanuel_magnetic_2018,furutani_graph_2020} and its eigenvectors. For this recall its definition:
\begin{equation}\label{eq:appendix_unsymmetric_magnetic_laplacian}
\begin{aligned}
    \mL^{(q)}_{U} 
    &\coloneqq \mD_{s} - \mA_{s} \odot \exp \left( i \boldsymbol{\Theta}^{(q)} \right)
\end{aligned}
\end{equation}
with Hadamard product \(\odot\), element-wise \(\exp\), \(\smash{i = \sqrt{\shortminus 1}}\), \(
    \Theta^{(q)}_{u,v} \coloneqq 2\pi q (\emA_{u,v} - \emA_{v,u})
\), and potential \(q \ge 0\). The degree-normalized counterpart is defined as
\begin{equation}
    \mL^{(q)}_{N} 
    \coloneqq \mI - \left( \mD_{s}^{- \nicefrac{1}{2}} \mA_{s} \mD_{s}^{- \nicefrac{1}{2}}  \right) \odot \exp \left( i \boldsymbol{\Theta}^{(q)} \right)
\end{equation}
Here the quadratic form (see \autoref{eq:appendix_quadratic_form_lap}) becomes 
\begin{equation}\label{eq:appendix_quadratic_form_maglap}
\begin{aligned}
    & \frac{1}{2} \sum_{(u,v) \in E_s} | \evx_u - \evx_v \exp(i\Theta^{(q)}_{u, v}) | ^2 \\
    &= \frac{1}{2} \sum_{(u,v) \in E_s} \overline{(\evx_u - \evx_v \exp(i\Theta^{(q)}_{u, v}))} (\evx_u - \evx_v \exp(i\Theta^{(q)}_{u, v})) \\
    &= \frac{1}{2} \sum_{(u,v) \in E_s} \bar{\evx}_u\evx_u - \exp(i \Theta_{u,v}^{(q)}) \bar{\evx}_u\evx_v - \exp(i \Theta_{v,u}^{(q)}) \bar{\evx}_v\evx_u + \underbrace{\exp(i \Theta_{u,v}^{(q)})\exp(i \Theta_{v,u}^{(q)})}_{=1} \bar{\evx}_v\evx_v \\
    &= \frac{1}{2} \sum_{(u,v) \in E_s} \bar{\evx}_u\evx_u - \exp(i \Theta_{u,v}^{(q)}) \bar{\evx}_u\evx_v - \exp(i \Theta_{v,u}^{(q)}) \bar{\evx}_v\evx_u + \bar{\evx}_v\evx_v \\
    &= \sum_{(u,v) \in E_s} \bar{\evx}_u\evx_u - \exp(i \Theta_{u,v}^{(q)}) \bar{\evx}_u\evx_v \\
    &= \underbrace{\sum_{(u,v) \in E_s} \bar{\evx}_u\evx_u}_{=\bar{\vx}^\top \mD_{s} \vx} \,-\, \underbrace{\sum_{(u,v) \in E_s} \exp(i \Theta_{u,v}^{(q)}) \bar{\evx}_u\evx_v}_{=\bar{\vx}^\top (\mA_{s} \odot \exp ( i \boldsymbol{\Theta}^{(q)})) \vx} \\
    &= \bar{\vx}^\top \left( \mD_{s} - \mA_{s} \odot \exp ( i \boldsymbol{\Theta}^{(q)}) \right) \vx \\
    &= \bar{\vx}^\top \mL_U^{(q)} \vx \\
\end{aligned}
\end{equation}
where \(E_s\) is the set of edges of the symmetrized graph. Note that either \(\Theta_{v,u}^{(q)} = - \Theta_{u, v}^{(q)}\) or \(\Theta_{v,u}^{(q)} = 0\). 

Recall that the first eigenvector minimizes the Rayleigh quotient
\begin{equation}\label{eq:appendix_maglap_rayleigh}
    \min_{\vx \in \C^n} \frac{\bar{\vx}^\top \mL_U^{(q)} \vx}{\bar{\vx}^\top \vx} 
    = \frac{\bar{\eigvec}_0^\top \mL_U^{(q)} \eigvec_0}{\bar{\eigvec}_0^\top \eigvec_0} 
    = \eigval_0
\end{equation}
From this, we can obtain the behavior of the first eigenvector of the Magnetic Laplacian (as illustrated in \autoref{fig:exemplary_eigenvecotr_maglap}). The first eigenvector \(\eigvec_0\) is related to the so-called ``angular synchronization problem'' (\autoref{eq:appendix_angular_synchron}). In angular synchronization, we seek the \(\normltwo\)-optimal estimate of \(n\) angles \(\boldsymbol{\alpha}\) given \(m\) (noisy) measurements of phase offsets \(\alpha_u - \alpha_v \mod 2\pi\) where \(u, v \in \{0, 1, \dots, n-1\}\). Formally, the angular synchronization problem~\citep{bandeira_cheeger_2013} is defined as (we drop the normalizing constants as they do not influence the minimum)
\begin{equation}\label{eq:appendix_angular_synchron}
    \angle(\eigvec_0) \in {\arg\min}_{\boldsymbol{\alpha} \in [0, 2\pi)^n} \eta(\boldsymbol{\alpha}) \quad \text{with }
    \eta(\boldsymbol{\alpha}) = \sum\nolimits_{u, v \in E} |\exp(i \alpha_v)- \exp(i \alpha_u + i \Theta_{u,v})|^2
\end{equation}

\begin{figure}[H]
  \centering
  \makebox[\linewidth][c]{
    \(\begin{array}{ccc}
      \subfloat[Sequence]{\includegraphics[width=0.165\linewidth]{assets/spectrum/small_sequence_graph_0.pdf}} & \subfloat[Eigenvectors of Magnetic Laplacian]{\includegraphics[width=0.35\linewidth]{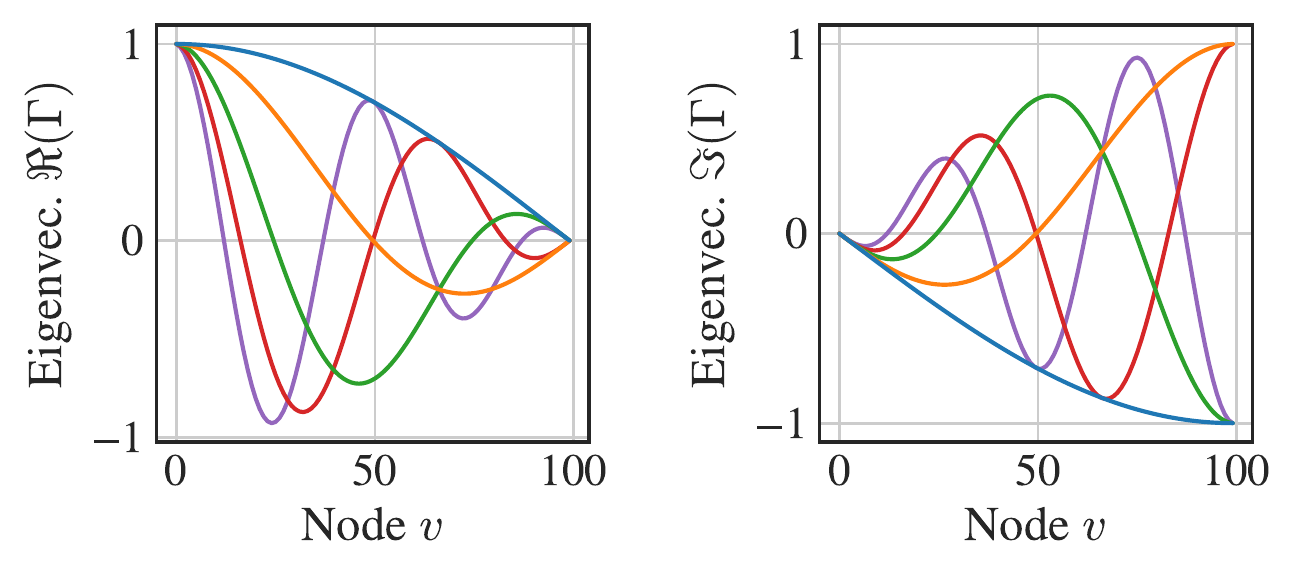}} & \subfloat[Eigenvec. of comb. Lap. w/o symmetrization]{\includegraphics[width=0.35\linewidth]{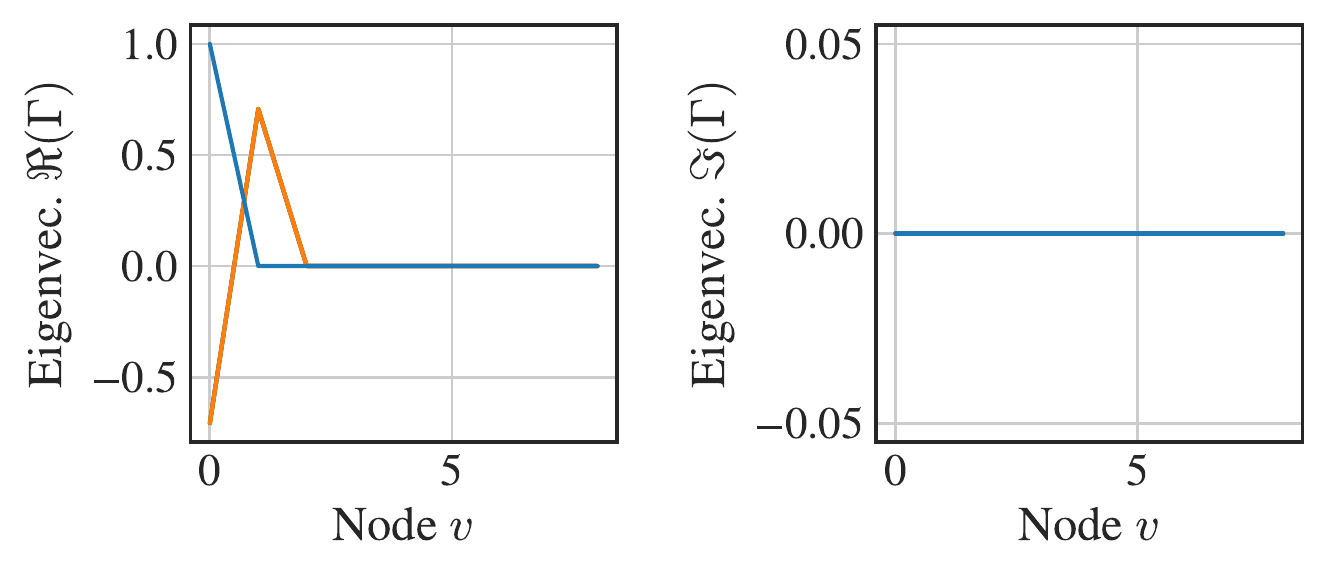}} \\
      
      \subfloat[Reversed sequence]{\includegraphics[width=0.165\linewidth]{assets/spectrum/small_reversed_sequence_graph_0.pdf}} & \subfloat[Eigenvectors of Magnetic Laplacian]{\includegraphics[width=0.35\linewidth]{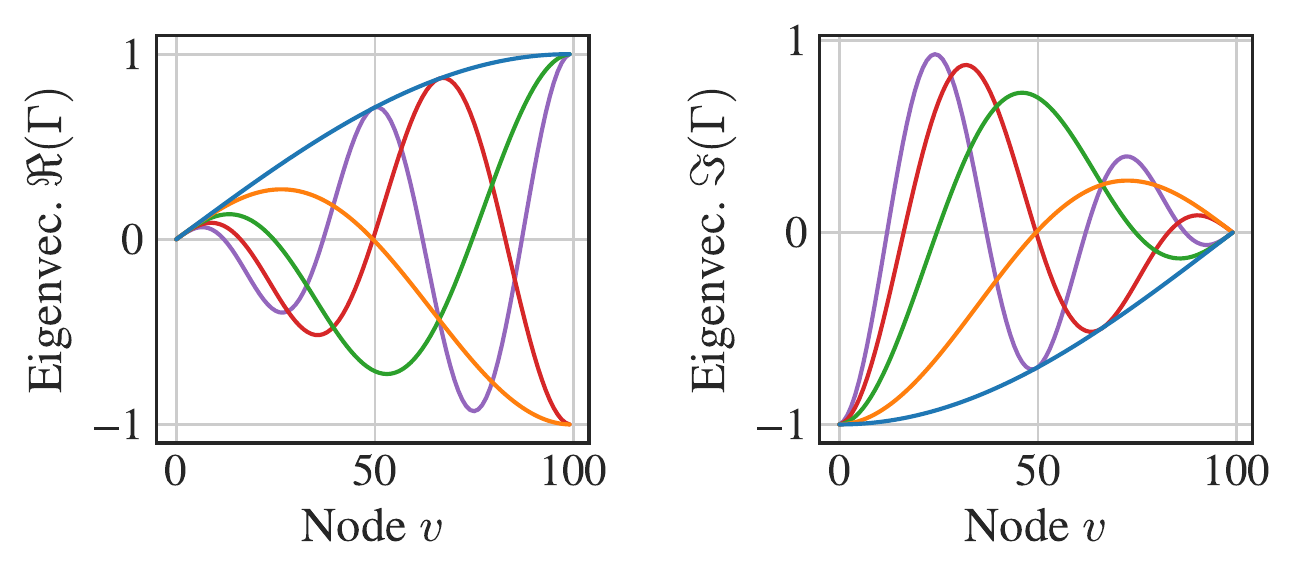}} & \subfloat[Eigenvec. of comb. Lap. w/o symmetrization]{\includegraphics[width=0.35\linewidth]{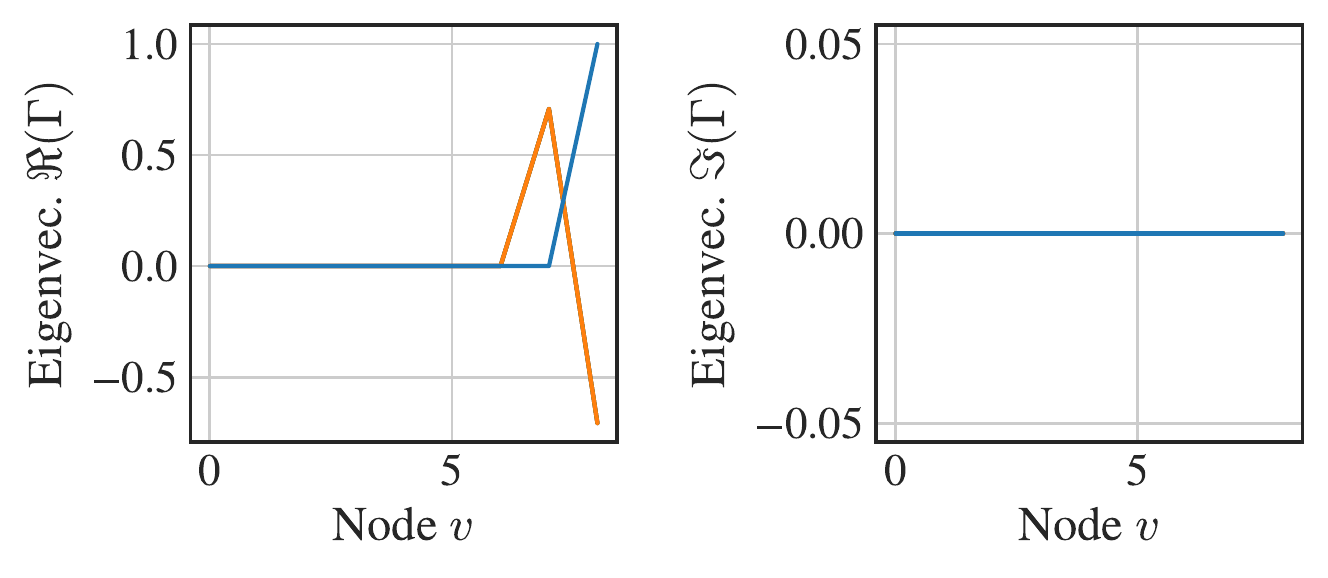}} \\
      
      \subfloat[Undirected sequence]{\includegraphics[width=0.165\linewidth]{assets/spectrum/small_undirected_sequence_graph_0.pdf}} & \subfloat[Eigenvectors of Magnetic Laplacian]{\includegraphics[width=0.35\linewidth]{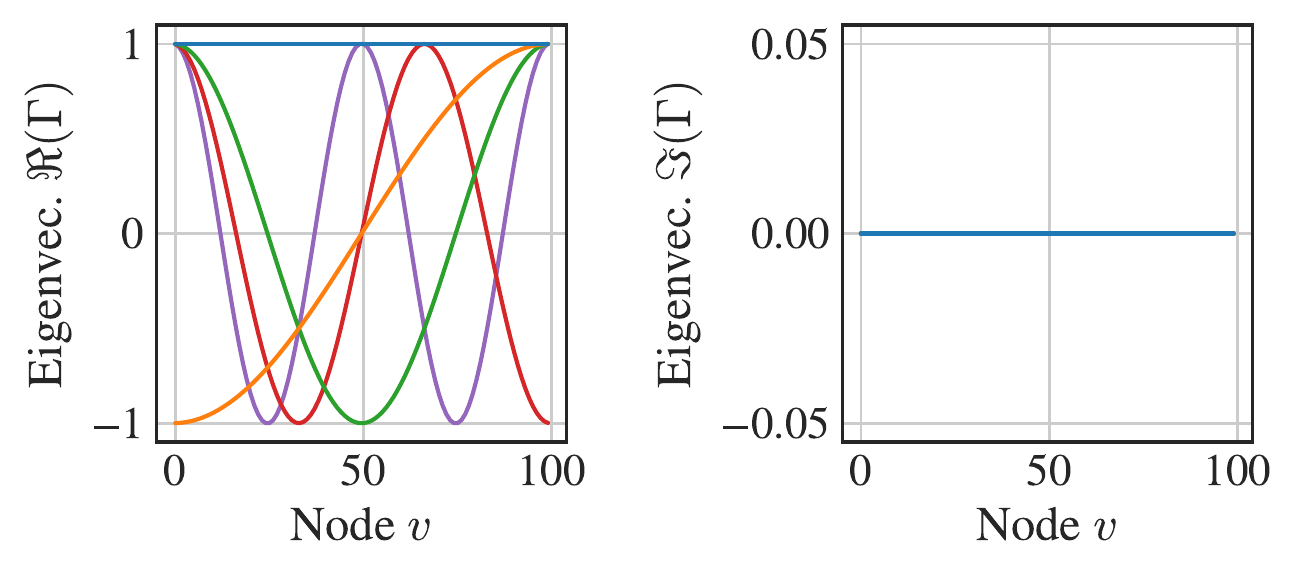}} & \subfloat[Eigenvec. of comb. Lap. w/o symmetrization]{\includegraphics[width=0.35\linewidth]{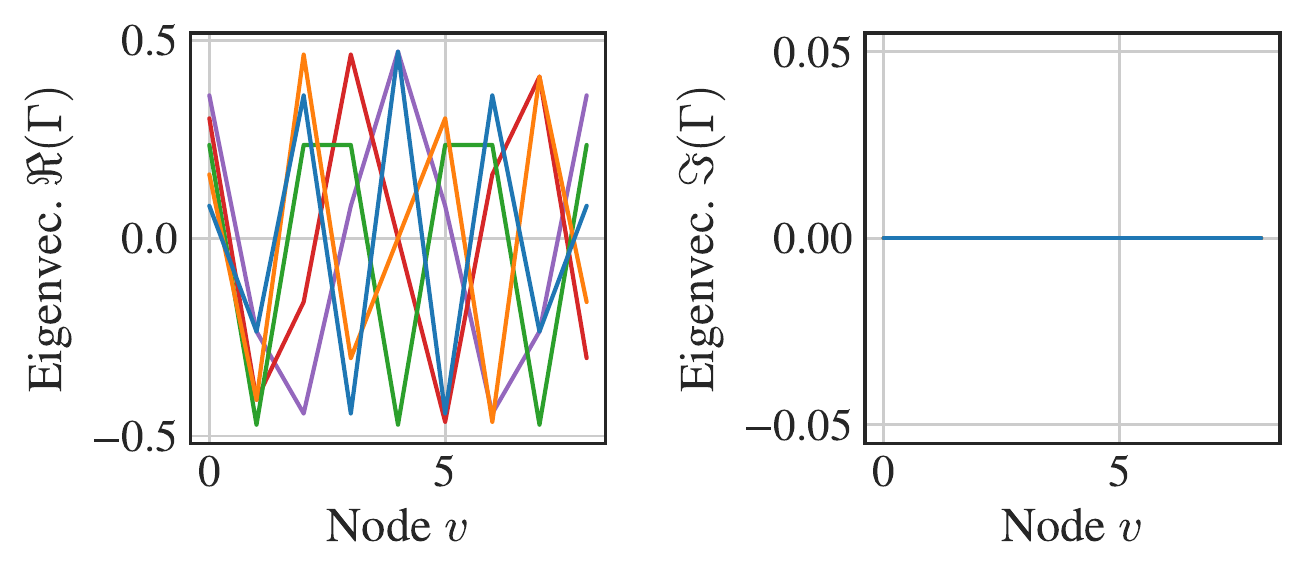}} \\
      
      \subfloat[Circle]{\includegraphics[width=0.165\linewidth]{assets/spectrum/small_circle_graph_0.pdf}} & \subfloat[Eigenvectors of Magnetic Laplacian]{\includegraphics[width=0.35\linewidth]{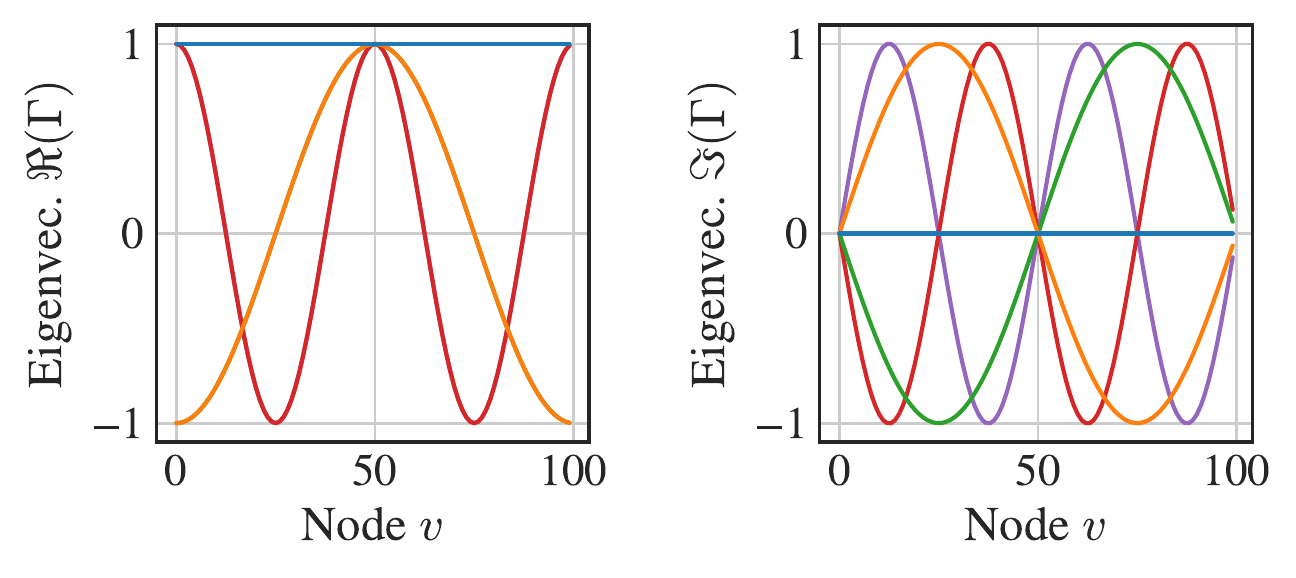}} & \subfloat[Eigenvec. of comb. Lap. w/o symmetrization]{\includegraphics[width=0.35\linewidth]{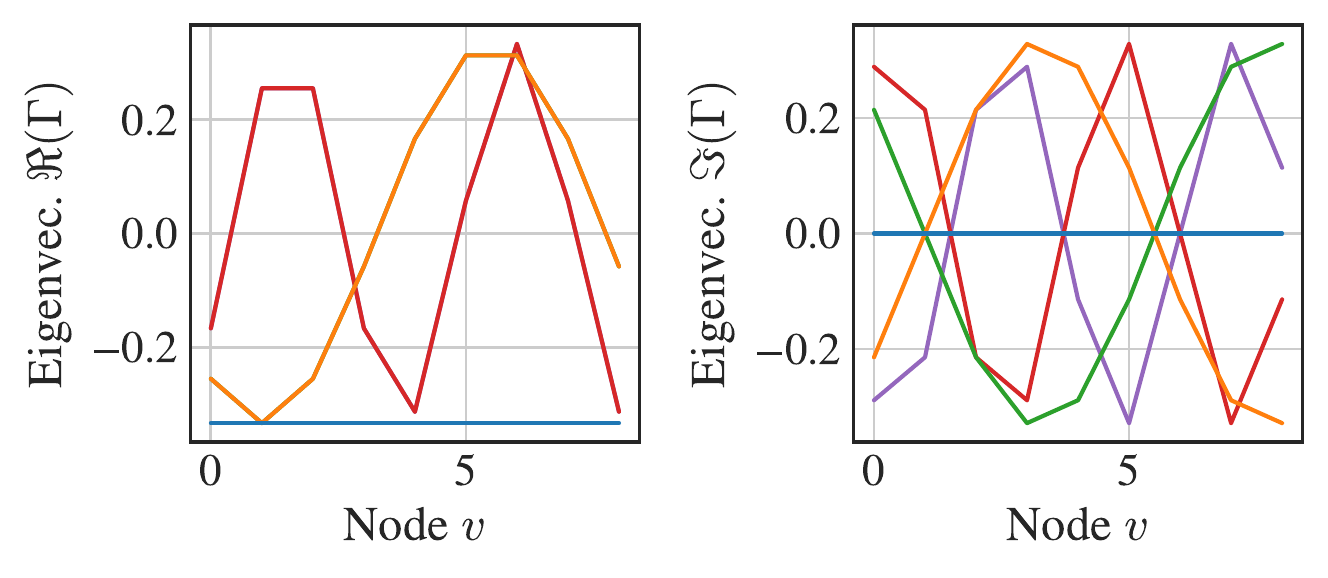}} \\
      
      \subfloat[Disconnected seq. sequence]{\includegraphics[width=0.165\linewidth]{assets/spectrum/small_two_sequences_graph_0.pdf}} & \subfloat[Eigenvectors of Magnetic Laplacian]{\includegraphics[width=0.35\linewidth]{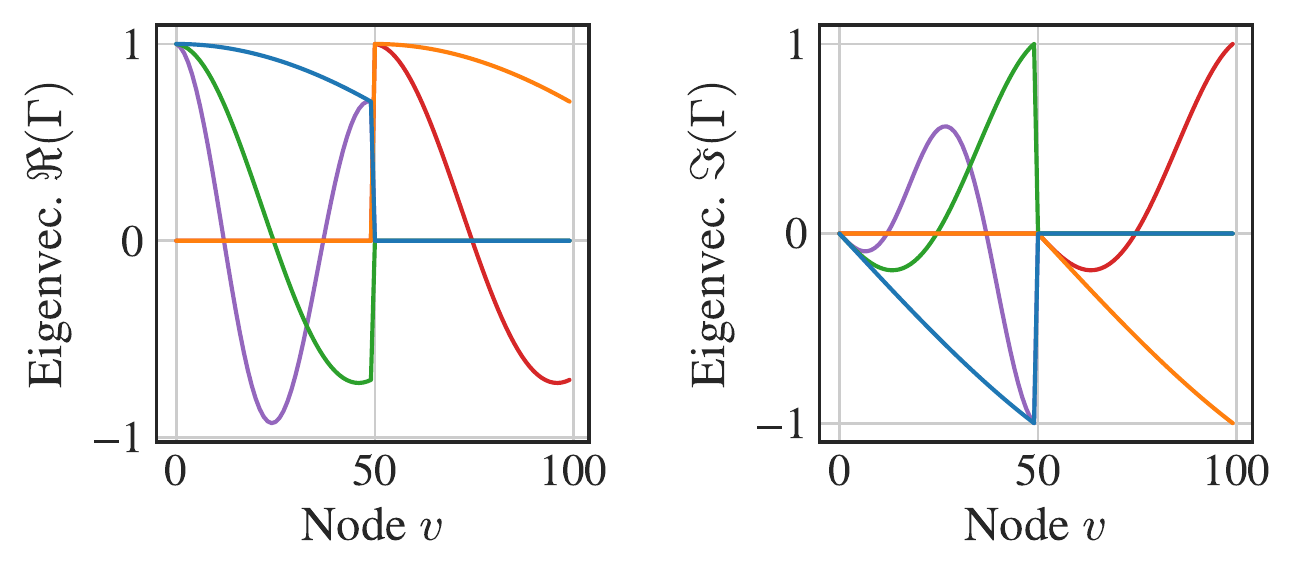}} & \subfloat[Eigenvec. of comb. Lap. w/o symmetrization]{\includegraphics[width=0.35\linewidth]{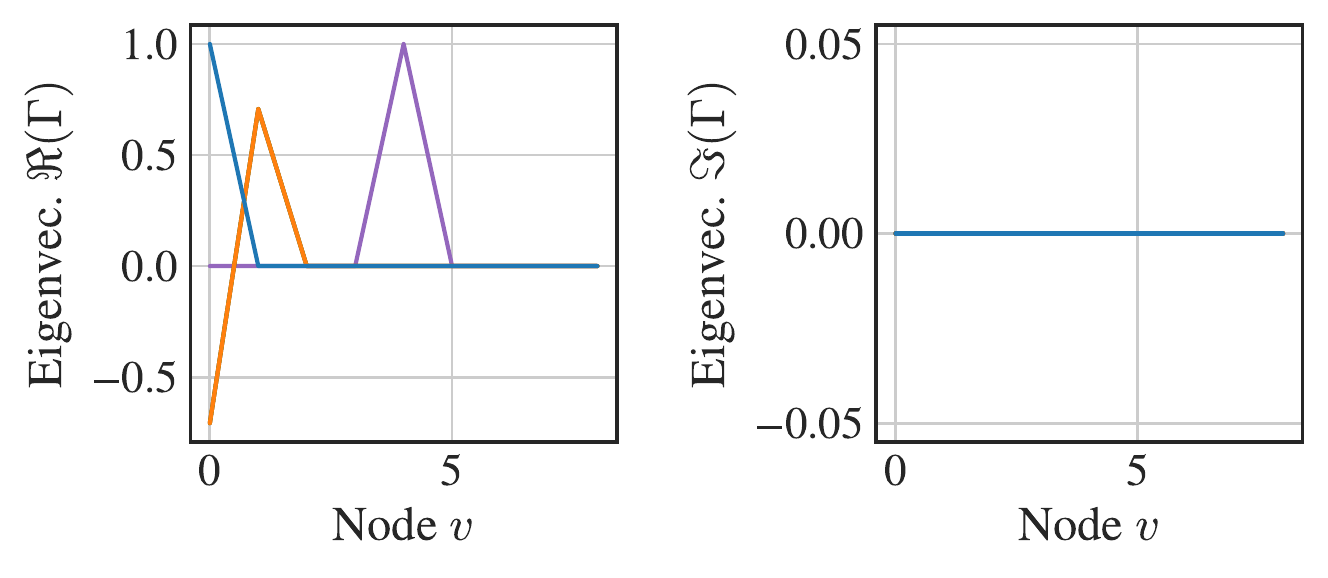}} \\
      
      \subfloat[Binary tree]{\includegraphics[width=0.165\linewidth]{assets/spectrum/small_balanced_binary_tree_graph_0.pdf}} & \subfloat[Eigenvectors of Magnetic Laplacian]{\includegraphics[width=0.35\linewidth]{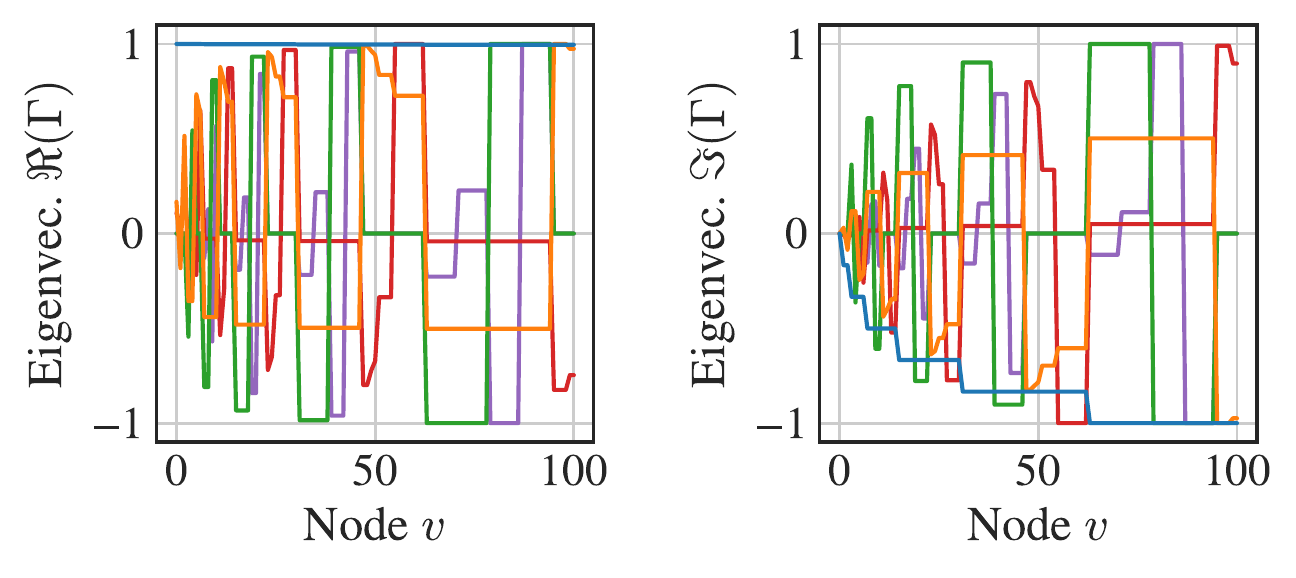}} & \subfloat[Eigenvec. of comb. Lap. w/o symmetrization]{\includegraphics[width=0.35\linewidth]{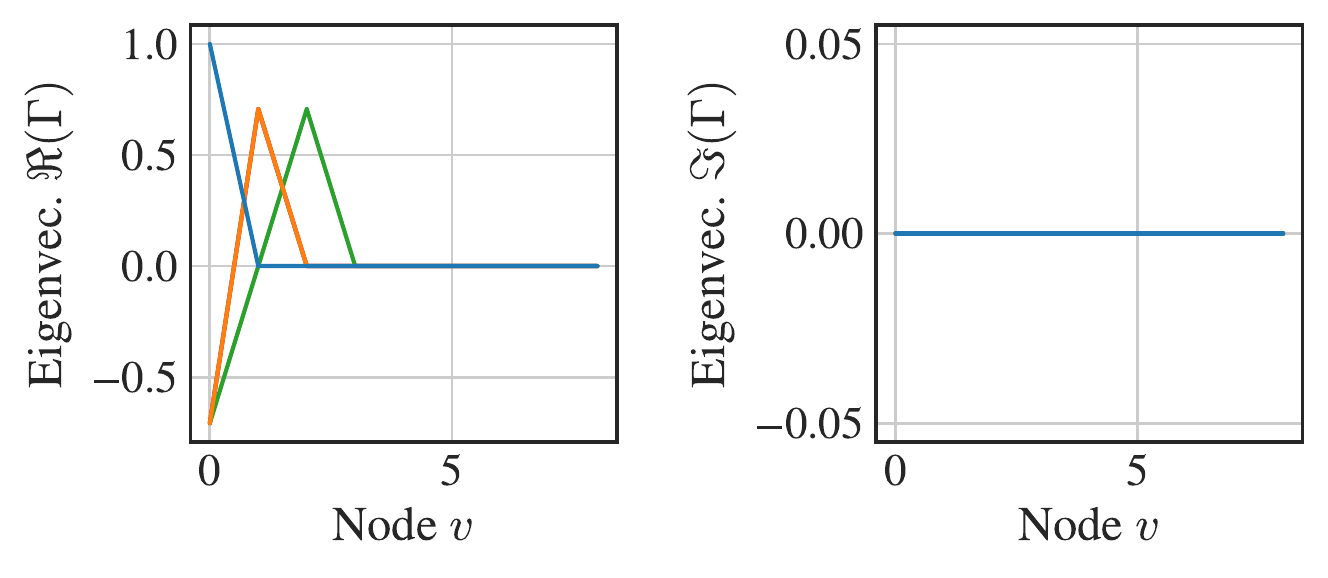}} \\
    \end{array}\)
  }
  \caption{First eigenvector(s) for sample graphs (part 1). In the left column (a, d, g, j, m, p), we show the first eigenvector of the Magnetic Laplacian for \(q=0.25\). The node size encodes the real value and colors the imaginary value. In the middle column (b, e, h, k, n, q), we show the first 5 eigenvectors on a graph with \(n=100\) nodes. In the right column (c, f, i, l, o, r), we show instead the eigenvectors of the Laplacian (\autoref{eq:laplacian_u}) omitting the symmetrization.
  \label{fig:appendix_example_graphs_spectra_part1}}
\end{figure}

\begin{figure}[H]
  \centering
  \makebox[\linewidth][c]{
    \(\begin{array}{ccc}
      \subfloat[Reversed bin. tree]{\includegraphics[width=0.165\linewidth]{assets/spectrum/small_reversed_balanced_binary_tree_graph_0.pdf}} & \subfloat[Eigenvectors of Magnetic Laplacian]{\includegraphics[width=0.35\linewidth]{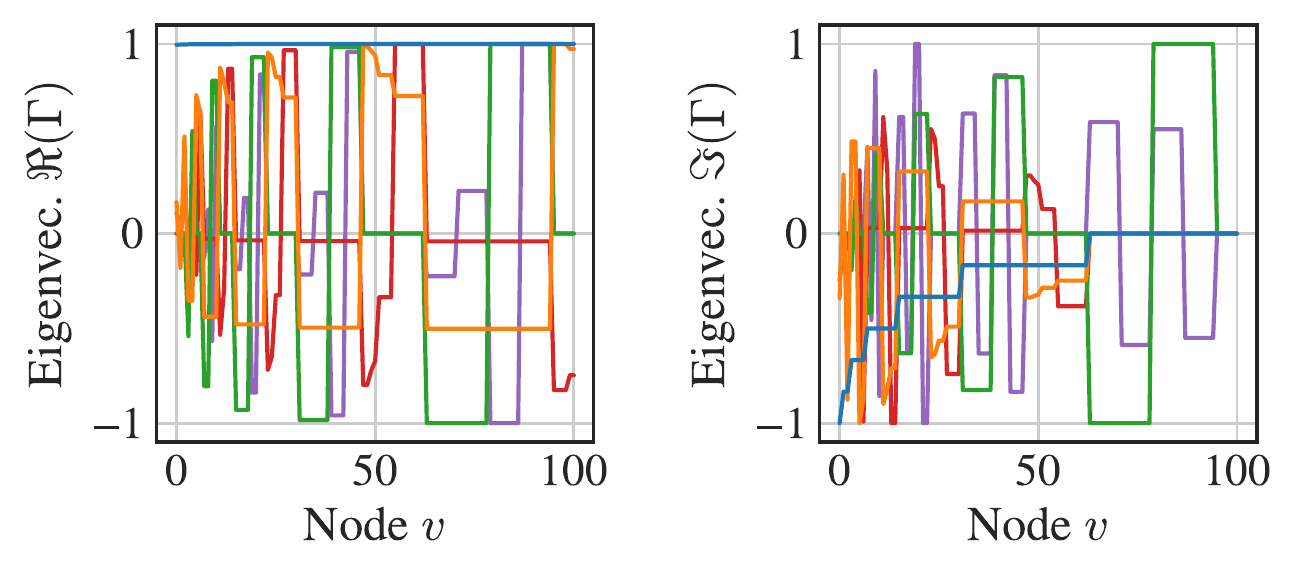}} & \subfloat[Eigenvec. of comb. Lap. w/o symmetrization]{\includegraphics[width=0.35\linewidth]{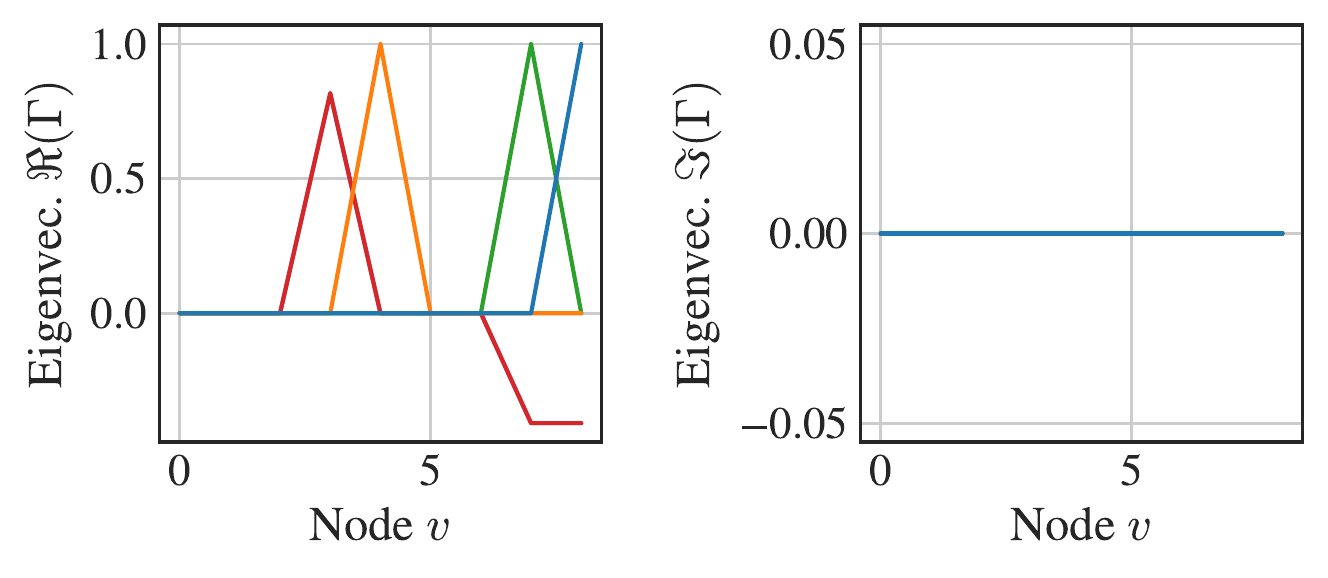}} \\
      
      \subfloat[Trumpet]{\includegraphics[width=0.165\linewidth]{assets/spectrum/small_trumpet_loop_graph_0.pdf}} & \subfloat[Eigenvectors of Magnetic Laplacian]{\includegraphics[width=0.35\linewidth]{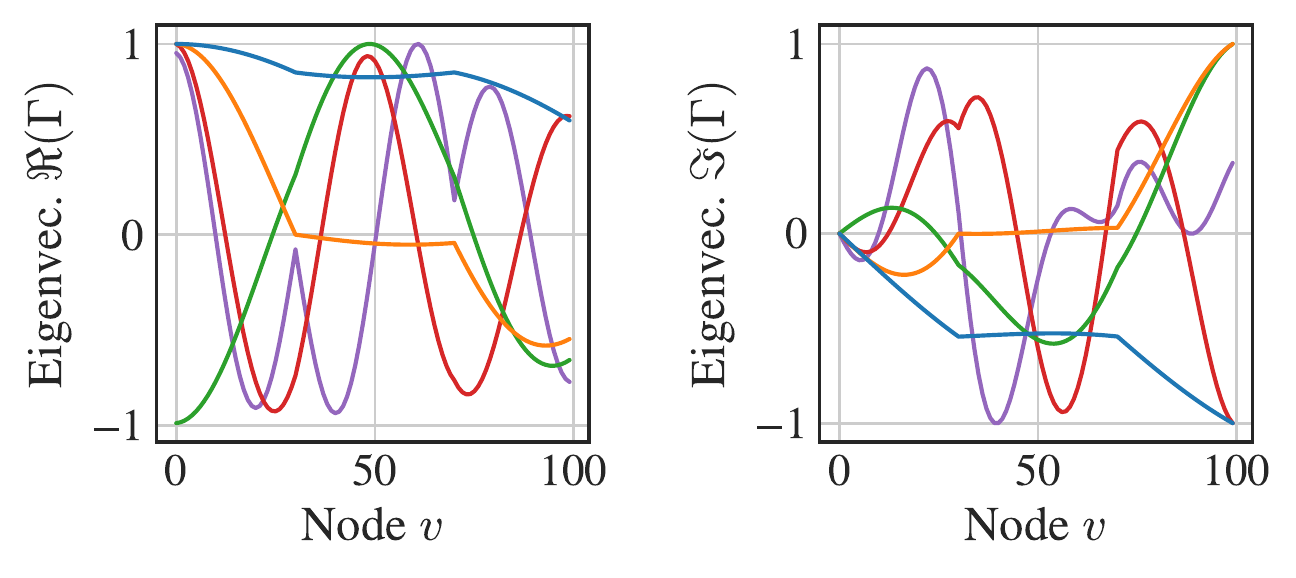}} & \subfloat[Eigenvec. of comb. Lap. w/o symmetrization]{\includegraphics[width=0.35\linewidth]{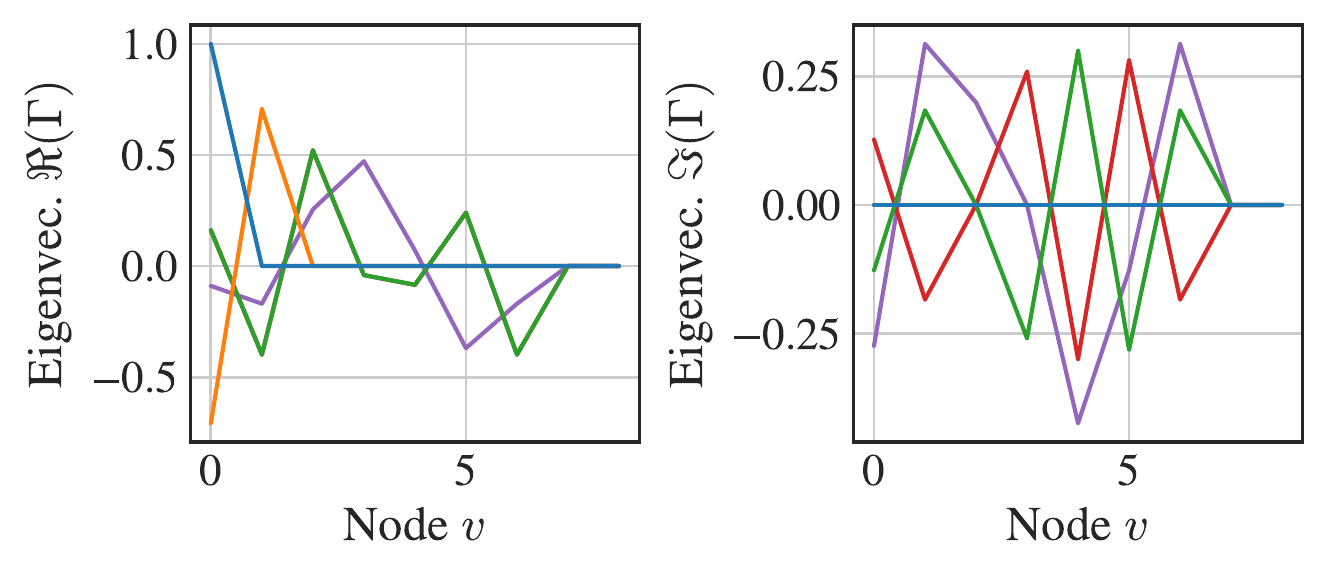}} \\
      
      \subfloat[Trumpet (forward)]{\includegraphics[width=0.165\linewidth]{assets/spectrum/small_trumpet_forward_graph_0.pdf}} & \subfloat[Eigenvectors of Magnetic Laplacian]{\includegraphics[width=0.35\linewidth]{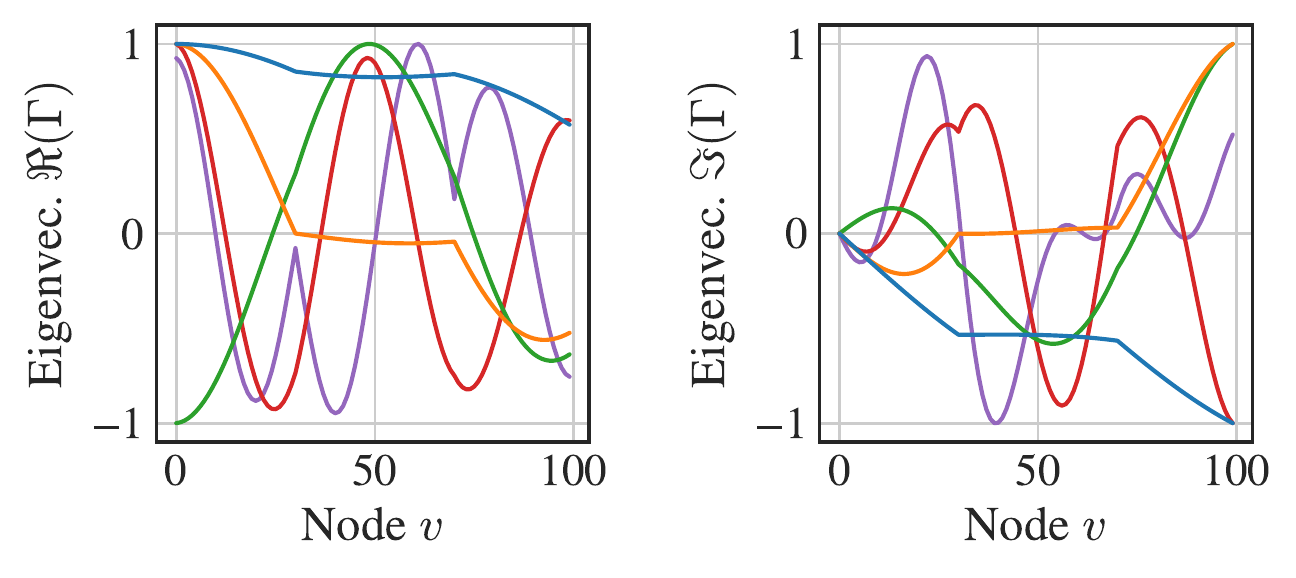}} & \subfloat[Eigenvec. of comb. Lap. w/o symmetrization]{\includegraphics[width=0.35\linewidth]{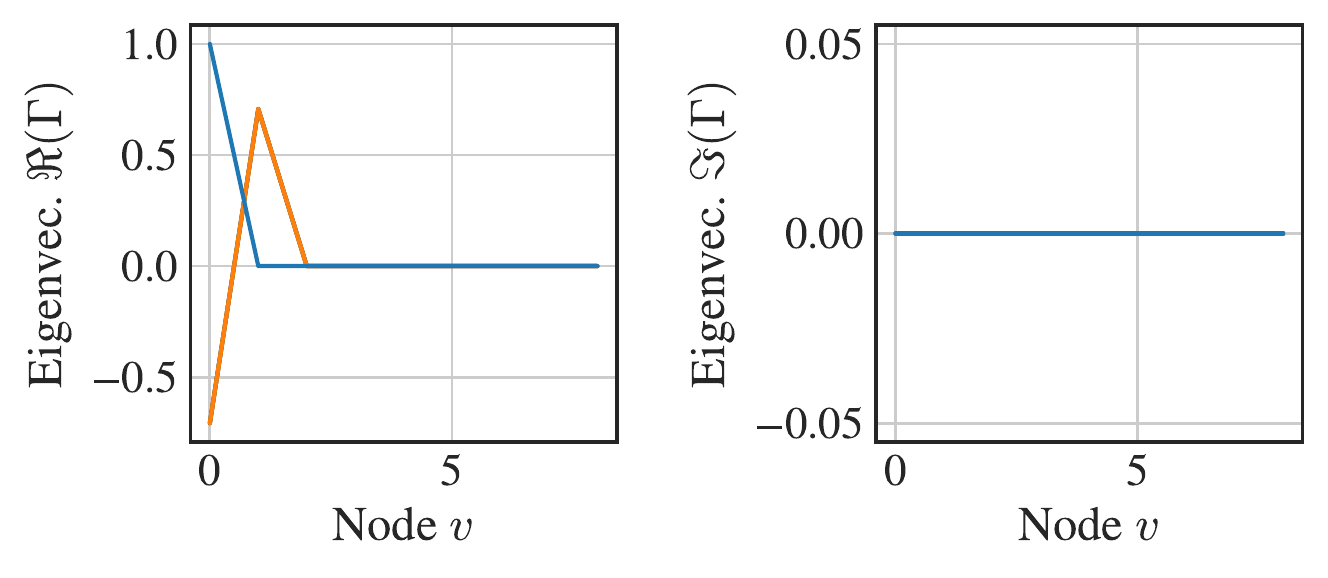}} \\
      
      \subfloat[Trumpet (DAG)]{\includegraphics[width=0.165\linewidth]{assets/spectrum/small_belly_snake_graph_0.pdf}} & \subfloat[Eigenvectors of Magnetic Laplacian]{\includegraphics[width=0.35\linewidth]{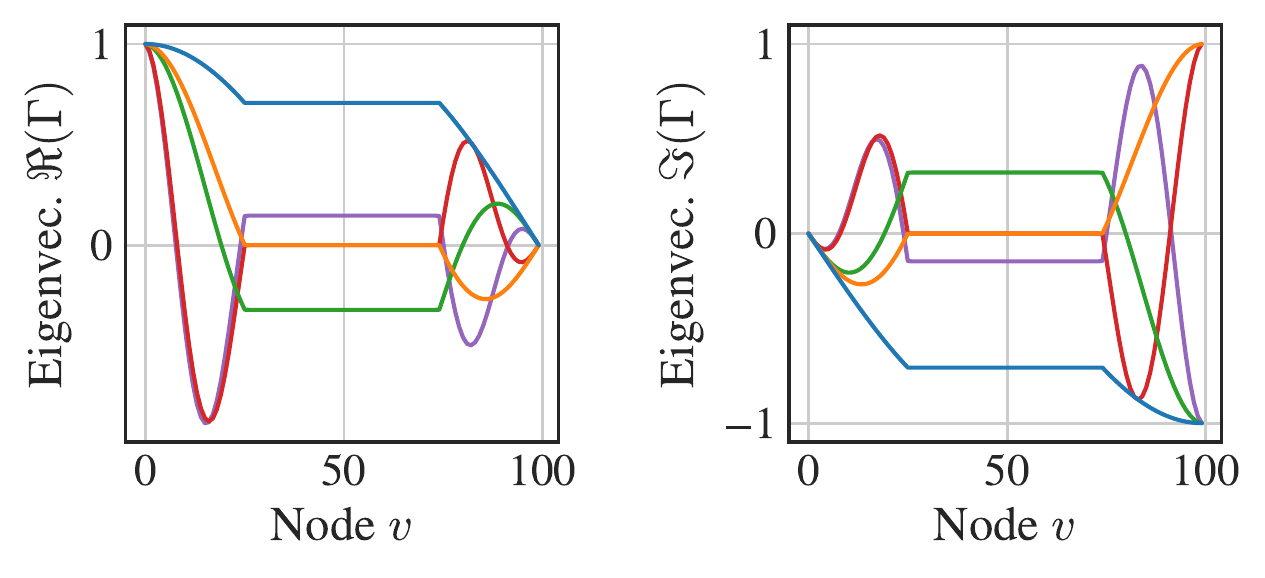}} & \subfloat[Eigenvec. of comb. Lap. w/o symmetrization]{\includegraphics[width=0.35\linewidth]{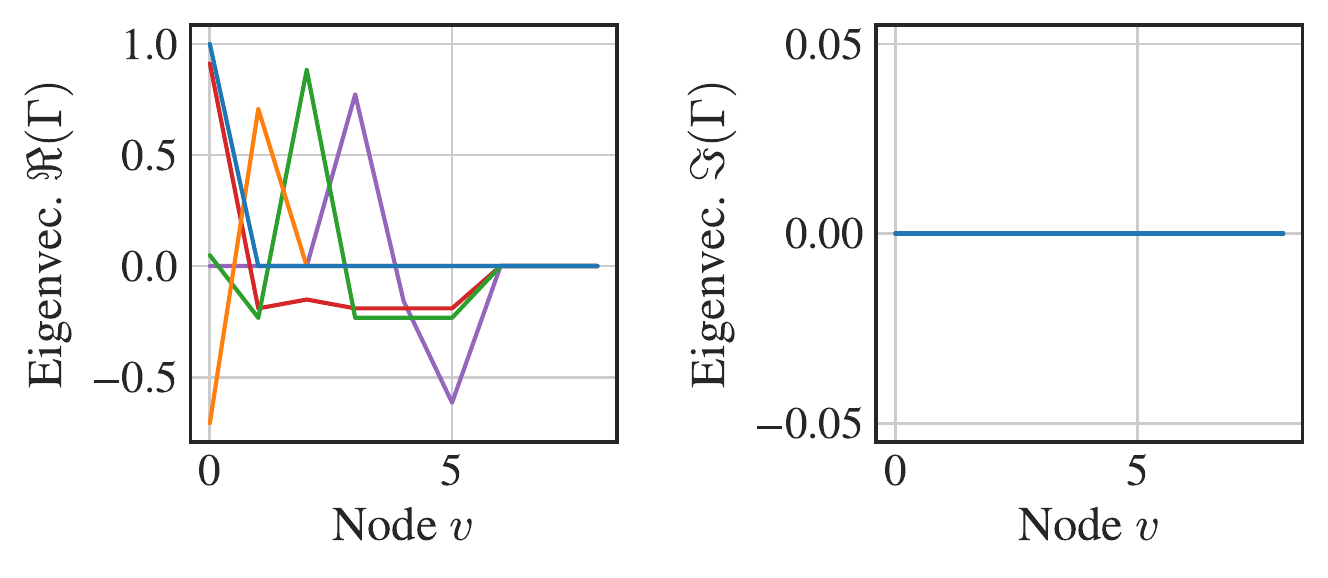}} \\
      
      \subfloat[Fully con.\ DAG]{\includegraphics[width=0.165\linewidth]{assets/spectrum/small_fully_connected_dag_graph_0.pdf}} & \subfloat[Eigenvectors of Magnetic Laplacian]{\includegraphics[width=0.35\linewidth]{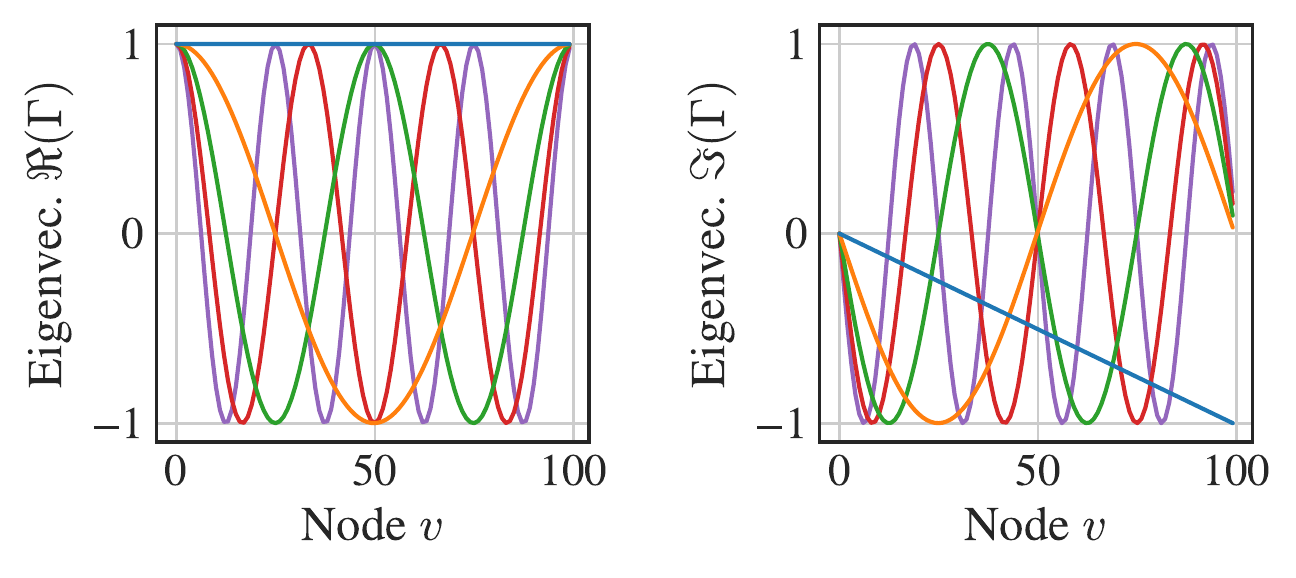}} & \subfloat[Eigenvec. of comb. Lap. w/o symmetrization]{\includegraphics[width=0.35\linewidth]{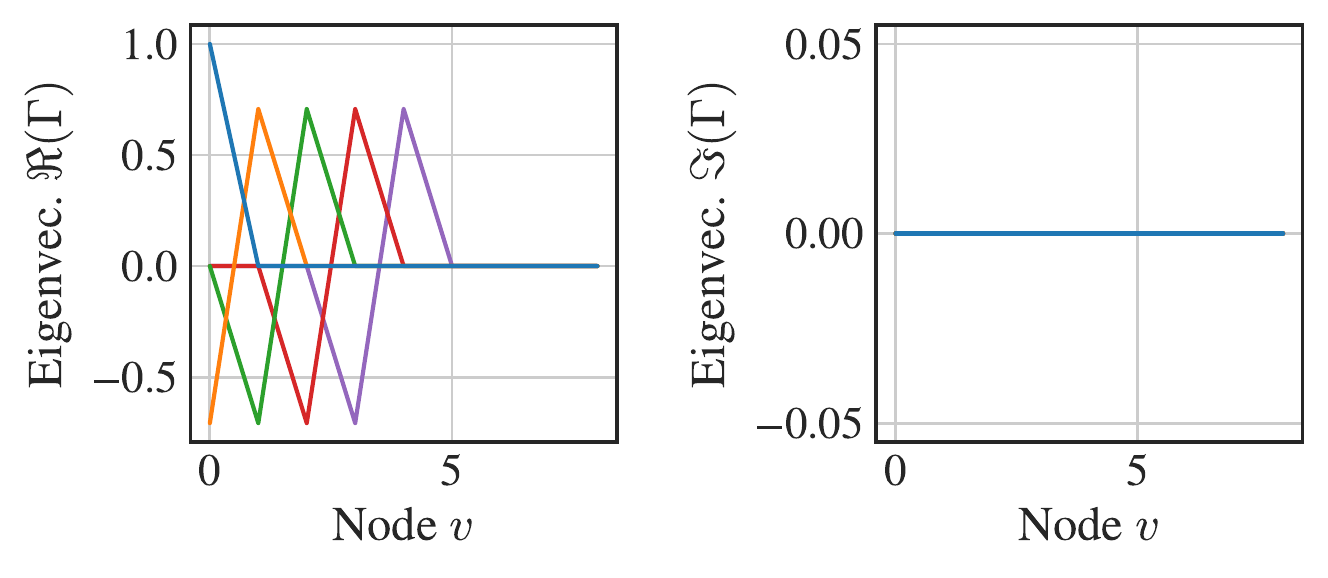}} \\
      
      \subfloat[Mix DAG \& f.\ con.]{\includegraphics[width=0.165\linewidth]{assets/spectrum/small_fully_connected_graph_dag_mix_graph_0.pdf}} & \subfloat[Eigenvectors of Magnetic Laplacian]{\includegraphics[width=0.35\linewidth]{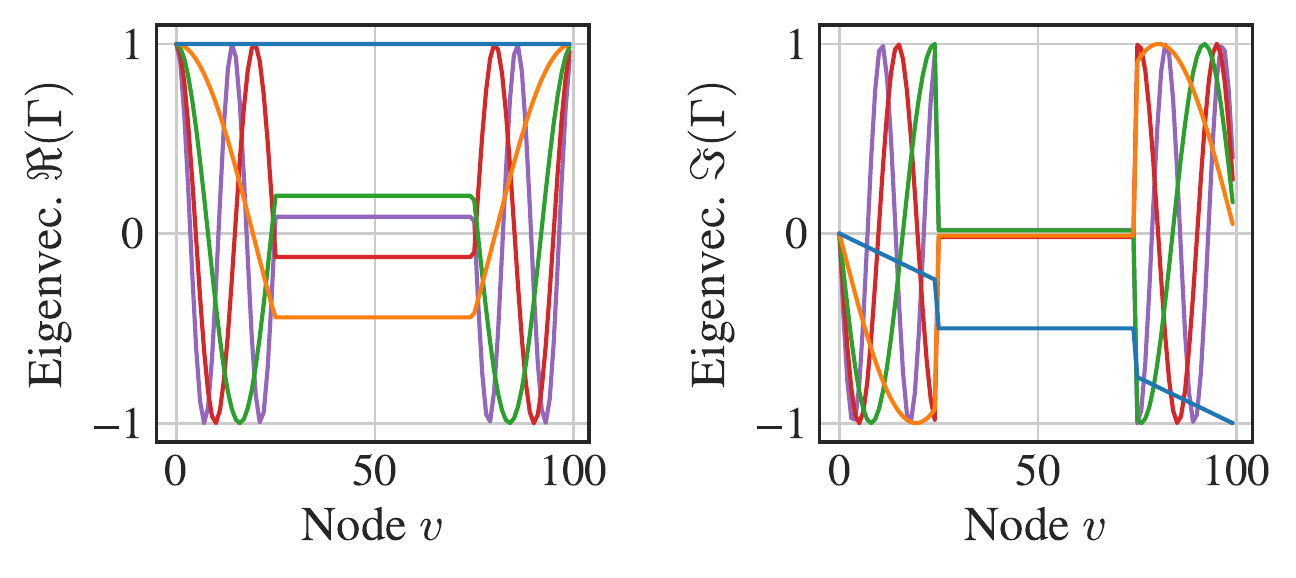}} & \subfloat[Eigenvec. of comb. Lap. w/o symmetrization]{\includegraphics[width=0.35\linewidth]{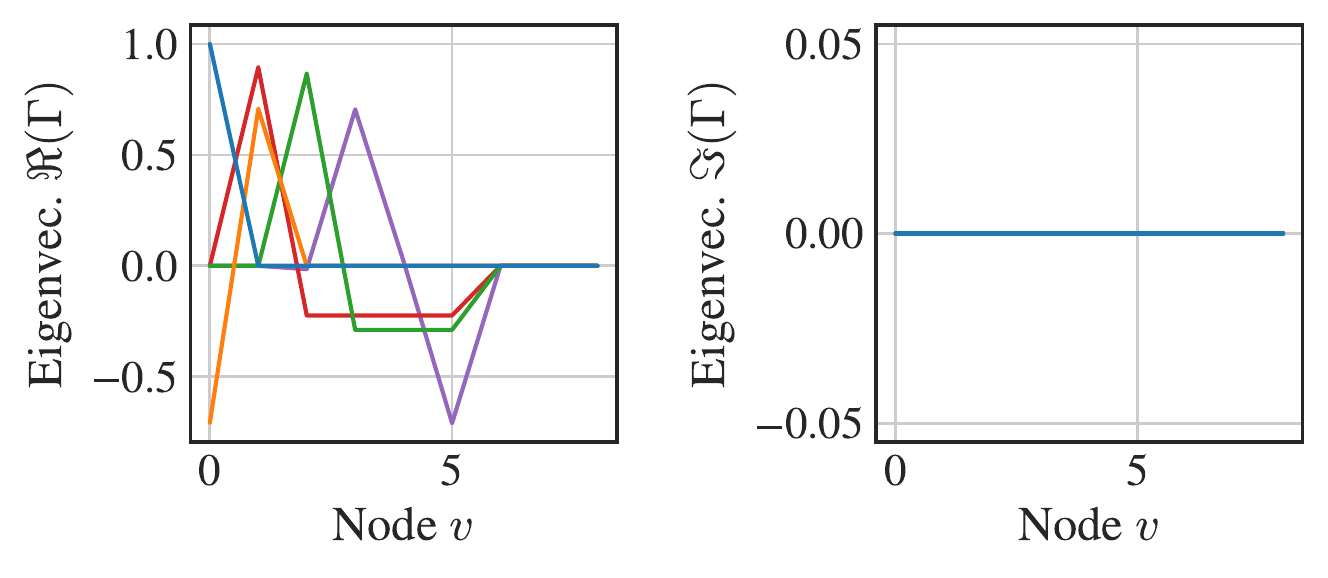}} \\
    \end{array}\)
  }
  \caption{First eigenvector(s) for sample graphs (part 1). In the left column (a, d, g, j, m, p), we show the first eigenvector of the Magnetic Laplacian for \(q=0.25\). The node size encodes the real value and colors the imaginary value. In the middle column (b, e, h, k, n, q), we show the first 5 eigenvectors on a graph with \(n=100\) nodes. In the right column (c, f, i, l, o, r), we show instead the eigenvectors of the Laplacian (\autoref{eq:laplacian_u}) omitting the symmetrization.
  \label{fig:appendix_example_graphs_spectra_part2}}
\end{figure}

In this analogy, each \emph{directed} edge \(\emA_{u,v} \ne \emA_{v,u}\) encourages a difference in the angle / phase in the first eigenvector \(\eEigvec_{u, 0}\) and \(\eEigvec_{v, 0}\), while an \emph{undirected} edge  \(\emA_{u,v} = \emA_{v,u} = 1\) supports them to be equal. We give an example in \autoref{fig:exemplary_eigenvecotr_maglap}. Here each directed edge induces a phase shift in \(\eigvec_0\) of \(2\pi q h \mod 2\pi\) and the undirected edges connect to nodes of equal phase.

In contrast to the combinatorial Laplacian, the first eigenvalue can only be equal to zero if the angles of the first eigenvector \(\eigvec_0\) (also see \autoref{eq:appendix_angular_synchron}) are ``conflict-free'', i.e., \(| \eEigvec_{u, 0} - \eEigvec_{v, 0} \exp(\Theta^{(q)}_{u, v}) | ^2 = 0\) for all \((u,v) \in \E\) (this term also appears in \autoref{sec:appendix_maglap_influence_q}). We plot the first eigenvector(s) of the Magnetic Laplacian in the first two columns of \autoref{fig:appendix_example_graphs_spectra_part1} and \ref{fig:appendix_example_graphs_spectra_part2}. For an eased comparability, here we normalize the eigenvectors s.t.\ the maximum absolute value is equal to one (\(\max_u |\eigvec_u| = 1\)).

\textbf{Relationship between combinatorial and Magnetic Laplacian.} For certain graphs we can relate the eigenvectors of the Magnetic Laplacian to the eigenvectors of the combinatorial Laplacian: \(\eEigvec^{(q)}_{v, j} = c \evs_v \eEigvec^{(0)}_{v, j}\) for node \(v\), the \(j\)-th eigenvector, normalizer \(c \in \C \setminus \{0\}\), and vector \(\vs \in \C^{n}\). Moreover, we define \(\mS\) as the diagonal matrix with \(\vs\) on its diagonal. Then, if we choose \(\mS\) s.t.\ \(\mL^{(q)} \mS = \mS \mL^{(0)}\) it follows that \(\mL^{(q)} \mS \eigvec_j^{(0)} = \mS \mL^{(0)} \eigvec_j^{(0)} = \mS (\mL^{(0)} \eigvec_j^{(0)})\). Since, \(\mL^{(0)} \eigvec_j^{(0)} = \eigval_j^{(0)} \eigvec_j^{(0)}\) we conclude \(\mL^{(q)} \mS \eigvec_j^{(0)} =  \mS (\eigval_j^{(0)} \eigvec_j^{(0)}) = \eigval_j^{(0)} \mS \eigvec_j^{(0)}\). Thus, the eigenvectors of the Magnetic Laplacian can be calculated form the eigenvectors of the combinatorial Laplacian \(\eigvec_j^{(q)} = \mS \eigvec_j^{(0)}\) if \(\mS\) exists and is known. For example, for trees and sequences it is trivial to construct \(\mS\). Here the elements on the diagonal can be chosen to \(\emS_{v,v} = \exp(-i 2 \pi \evd_v)\), where \(\evd_v\) is the distance from the root node to node \(v\).

\textbf{Repeated eigenvalues.}  A source of ambiguity in the eigenvectors \(\Eigvec\) emerges in the case of \(l\) repeated eigenvalues (also called multiple eigenvalues) of a connected component. Then, the respective eigenvectors can be chosen from an \(l\)-dimensional subspace as long as they are orthogonal. We refer to \citep{lim_sign_2022} and \citet{wang_equivariant_2022} on how to address it.

\textbf{Permutation equivariance.} Notably, in the presence of repeated eigenvalues, the eigenvectors calculated by the eigensolvers are generally not equivariant to node-permutations. Following, also our eigenvector encodings are not permutation equivariant for graphs with repeated eigenvalues. Similarly, permutation equivariance is affected by the arbitrary factor \(c \in \C \setminus \{0\}\) we can apply to eigenvectors \(c \mL \eigvec = c \eigval \eigvec \implies \mL (c \eigvec) = \eigval (c \eigvec)\). Thus, on needs also to be careful about modelling the scale, sign and rotation of \(c\) to retain permutation equivariance. If using a sign-net like encoder with our convention of normalizing the eigenvectors, we obtain equivariance w.r.t.\ scale, sign and rotation.

\textbf{Disconnected components.} In the case of disconnected components, the eigenvectors and eigenvalues resolve as if we would decompose each component independently, where the components of the eigenvectors for the other components are set to zero. For example, for two disconnected sequences \autoref{fig:appenidx_example_graphs:i} with an even number of nodes, we obtain two equal disconnected components. Here, we will obtain each eigenvalue \(\eigval\) twice. Also, \(\Eigvec\) contains two full sets of equivalent eigenvectors, except that they are zero for the other component and vice versa.

\textbf{Co-spectrality}~\cite{stevanovic_research_2007} describes the phenomenon that there exist (potentially non-isomorphic) graphs with identical eigenvalues. Co-spectrality is of lesser concern for our architecture since we also process the eigenvectors. Similarly to co-spectrality, many well-known facts for the combinatorial Laplacian (e.g., the Cheeger inequality) extend to the Magnetic Laplacian. For the interested reader, we refer to \cite{bandeira_cheeger_2013}.

\subsection{Influence of Potential \(q\)}\label{sec:appendix_maglap_influence_q}

The potential \(q\) seems to be a crucial choice for successfully encoding direction. For example, in \autoref{fig:sn_qsweep}, we see that for too large potentials \(q\), the performance degrades on the sorting network tasks. Moreover, \citet{furutani_graph_2020} show that for too large values of \(q\), the eigenvectors of the Magnetic Laplacian do not order from low to high frequencies. In other words, then the eigenvalue order is not predictive for the variation of the graph signal.

In applications of the Magnetic Laplacian stemming from physics, \(q\) is typically given since, e.g., it represents the electric charge in a magnetic field. However, in general, it is not clear how to derive appropriate \(q\) values from the domain. Although the Magnetic Laplacian has been used for a Spectral GNN~\citep{zhang_magnet_2021}, Community Detection~\citep{fanuel_magnetic_2016} or the visualization of directed graphs~\citep{fanuel_magnetic_2018}, there is not much guidance on how to choose \(q\) in these works. These works treat \(q\) simply as a (hyper-) parameter.

\begin{figure}[t]
  \centering
  \includegraphics[width=0.7\linewidth]{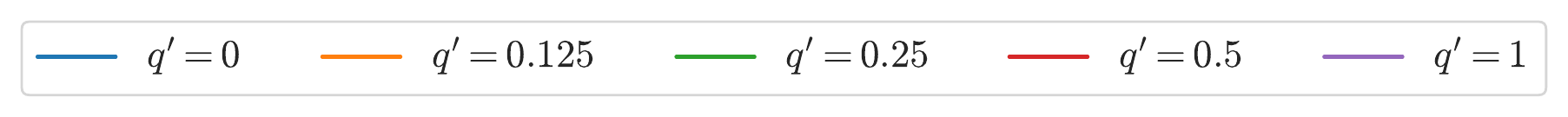}
  \resizebox{1.0\linewidth}{!}{
      \subfloat[Sequence]{
      \includegraphics[height=2.5cm]{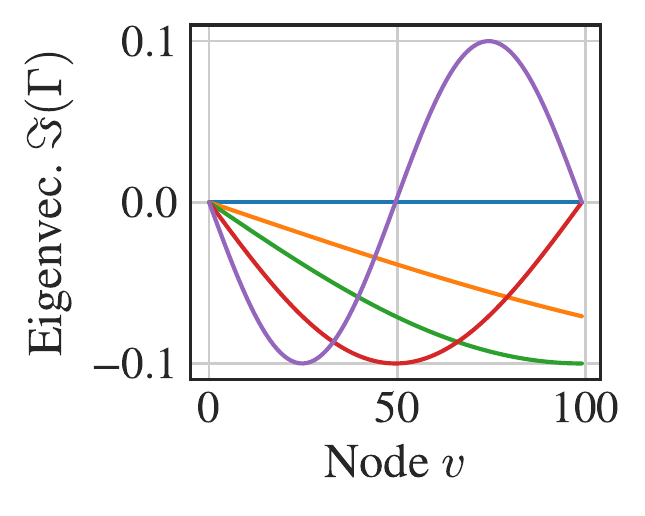}}
      \subfloat[Undirected Seq.]{
      \includegraphics[height=2.5cm]{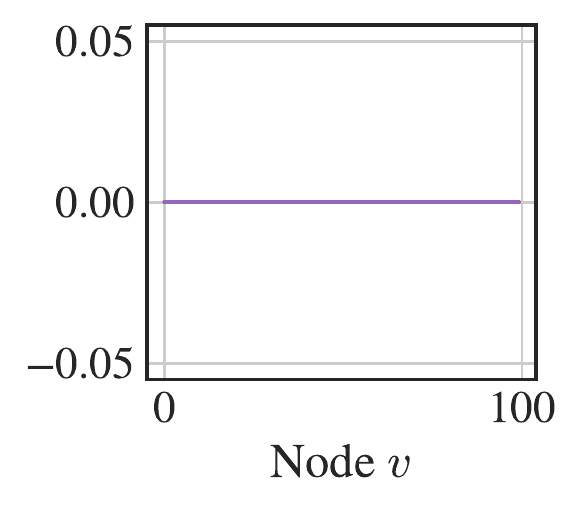}}
      \subfloat[Binary Tree]{
      \includegraphics[height=2.5cm]{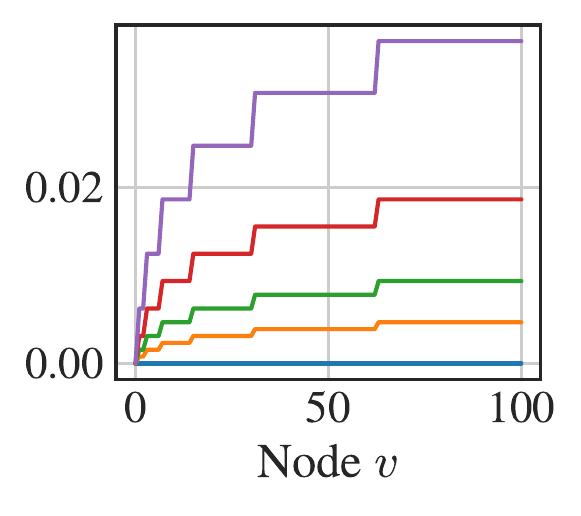}}
      \subfloat[Trumpet]{
      \includegraphics[height=2.5cm]{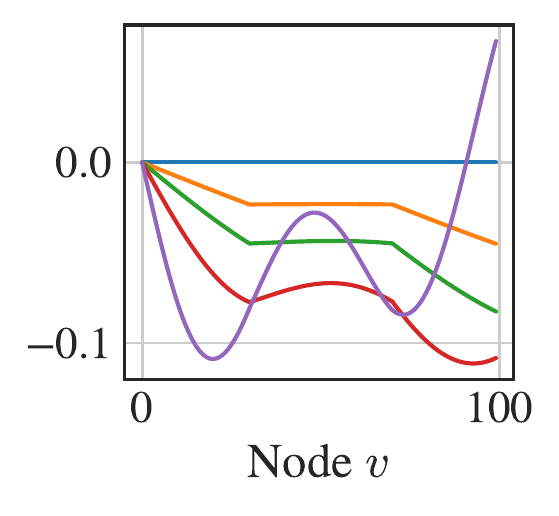}}
      \subfloat[FC DAG]{
      \includegraphics[height=2.5cm]{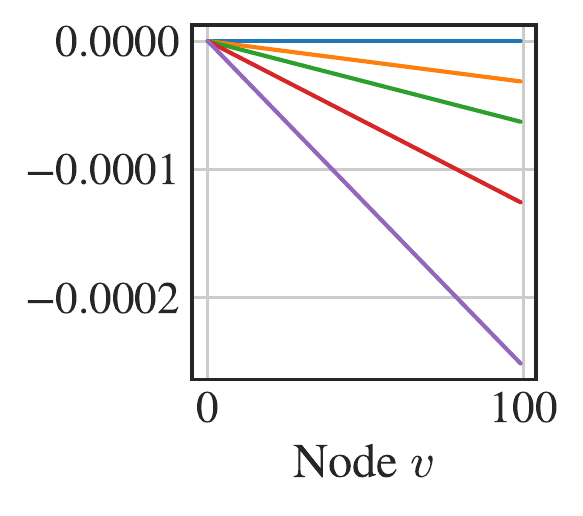}}
  }
  \caption{The influence of \(q\) on the imaginary part of the eigenvectors for the graphs in \autoref{fig:example_graphs}.}
  \label{fig:appenidx_phase_shift}
\end{figure}

\citet{fanuel_magnetic_2018} open a perspective on \(q\) that is perhaps not well-suited for positional encodings but conveys an interesting intuition. They argue to choose \(q\) as a rational number. For example, \(q = \nicefrac{1}{3}\) is particularly well-suited for visualizing graphs that consist of directed triangles. In such a directed triangle, each edge can induce a shift of \(\nicefrac{2}{3} \pi\). Consequently, a triangle would induce a cumulative shift of a full 360 degrees.

In graph signal processing, \citet{furutani_graph_2020} propose to choose potential \(q\) using insights from eigenvector perturbation theory. However, they choose potential \(q\) based on the average node degree which is not related to the directedness of the graph. Additionally, it does not scale with \(n\) and, hence, the maximum phase shift between a source and target node is not bounded for larger \(n\).

\textbf{Conflict-free graphs.} We argue that for positional encodings it is particularly helpful to bound the total phase shift (here for \autoref{eq:unsymmetric_magnetic_laplacian}) to avoid degenerate cases. For this we first discuss graphs we call \emph{conflict-free}, i.e., if its first eigenvalue is \(\bar{\eigvec}_0^\top \mL^{(q)} \eigvec_0 = \eigval_0 = 0\) for all \(0 < q \le \nicefrac{1}{4}\). It is easy to see that for graphs without conflicting edges the phase shift between at least one source and sink nodes is \(2 \pi q l\), where \(l\) is the maximum number of \emph{purely directed edges} (\(\{(u, v) \in E \,|\, (v, u) \notin E\}\)) in a path accounting for their direction (\autoref{eq:appendix_maglap_rayleigh}). That is, we increment f \(l\) for every \emph{purely directed edge} that we traverse in its direction and decrement \(l\) if we traverse such an edge against its direction. Bidirectional edges do not affect \(l\). This can be understood as a weighted longest simple path problem. In the following, we call \(l\) simply the longest simple path.

\begin{figure}[H]
  \centering
  \subfloat[Absolute potential \(q=0\) (relative \(q'=0\))]{
  \includegraphics[width=0.48\linewidth]{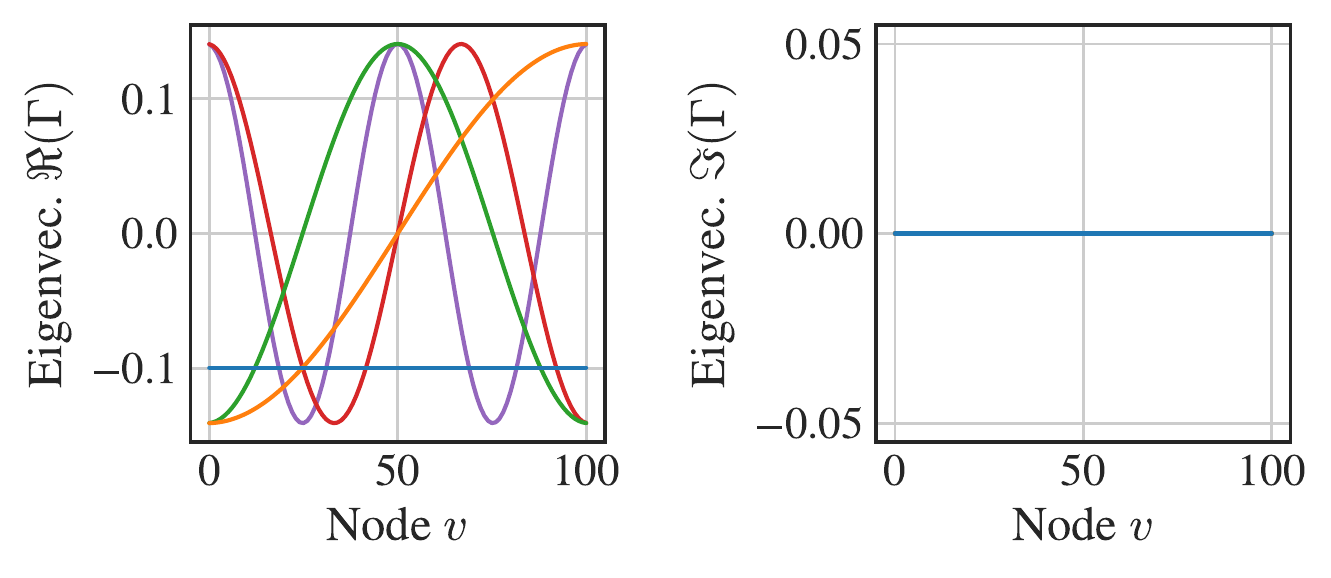}}
  \subfloat[Absolute potential \(q=\num{2.5e-3}\) (relative \(q'=2.5e-1\))]{
  \includegraphics[width=0.48\linewidth]{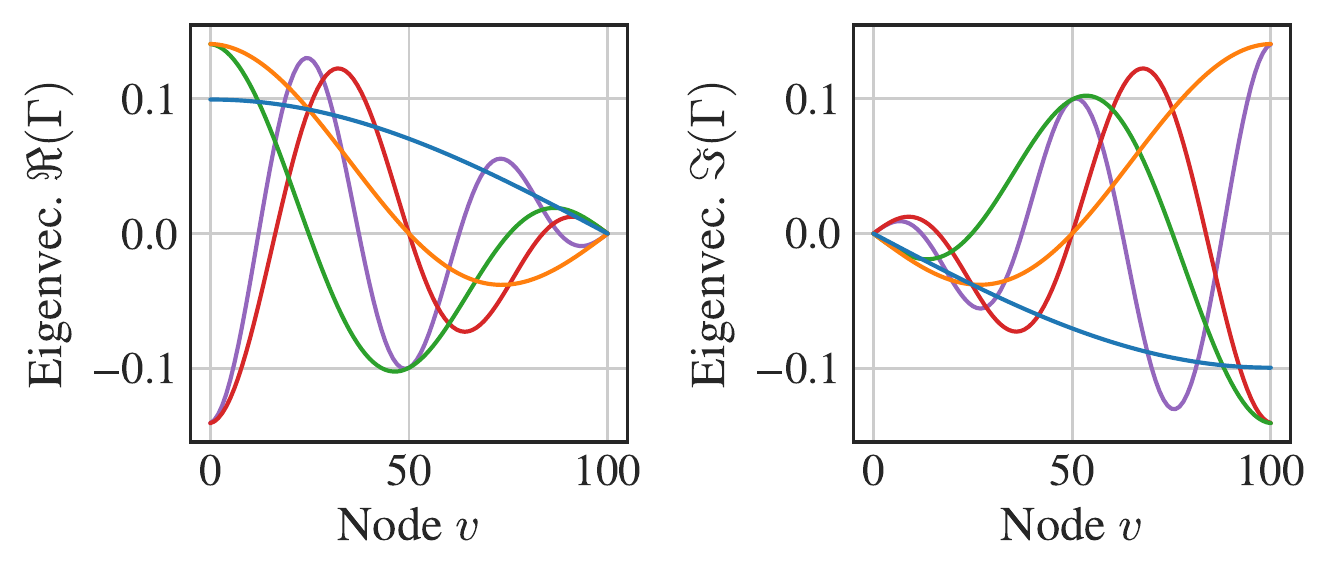}} \\
  
  \subfloat[Absolute potential \(q=\num{2.5e-2}\) (relative \(q'=2.5\))]{
  \includegraphics[width=0.48\linewidth]{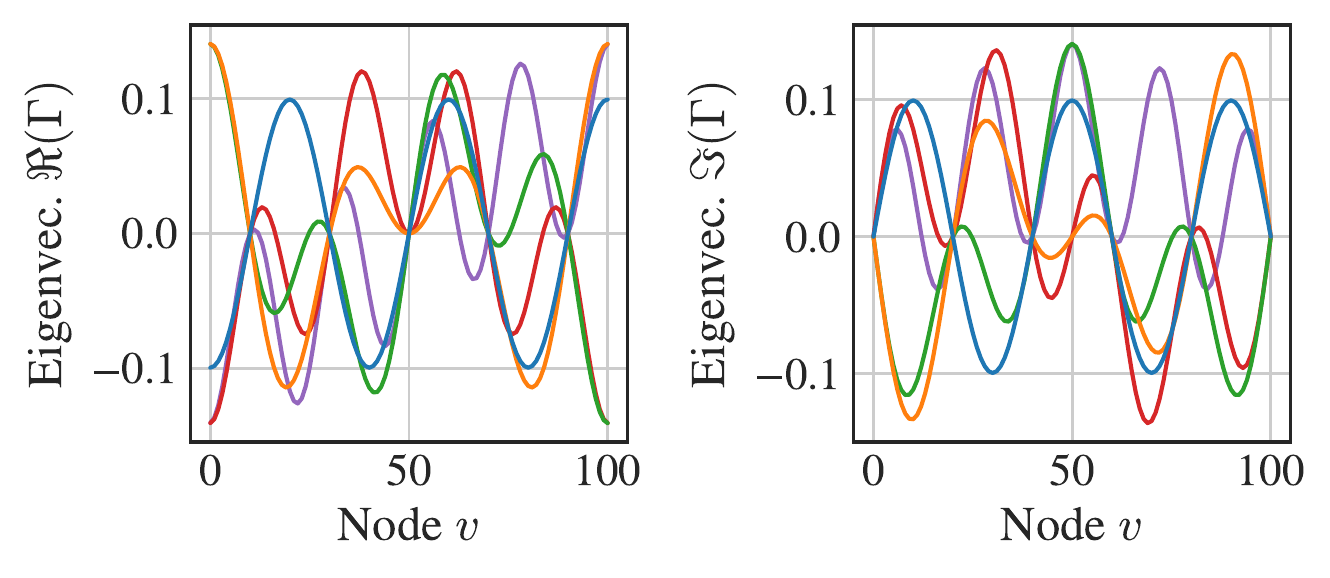}}
  \subfloat[Absolute potential \(q=\num{2.5e-1}\) (relative \(q'=25\))]{
  \includegraphics[width=0.48\linewidth]{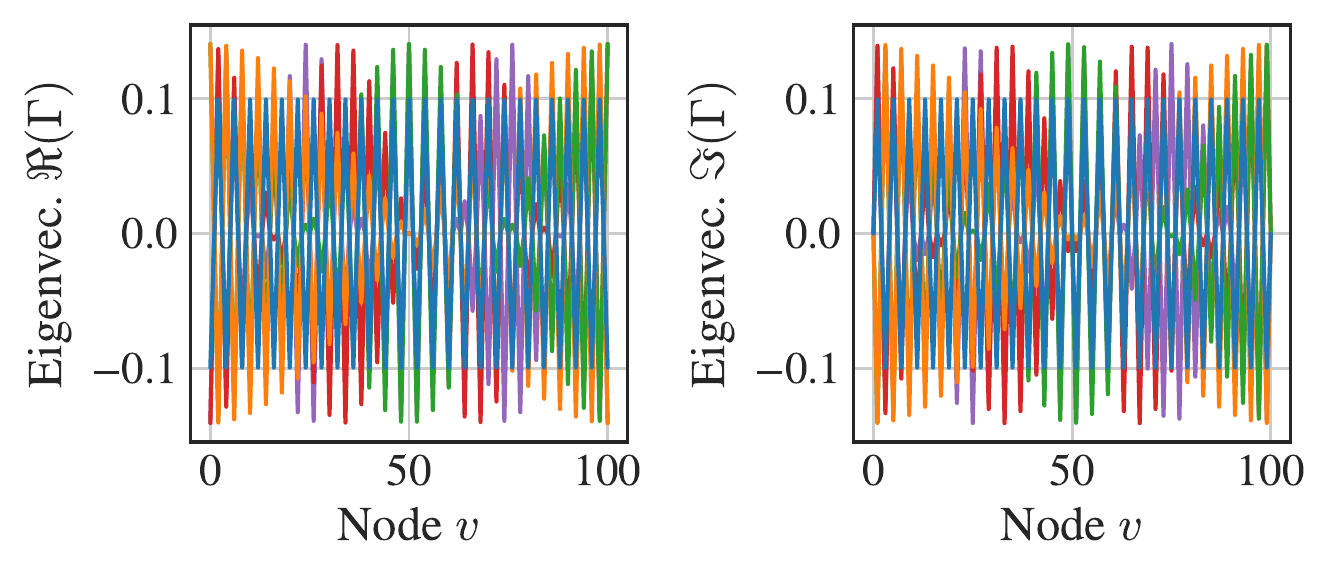}}
  \caption{Eigenvectors of Magnetic Laplacian (\autoref{eq:appendix_unsymmetric_magnetic_laplacian}) for a sequence \autoref{fig:example_graphs:a} with \(n=101\) nodes where we also include particularly large \(q\) values (see subcaptions).
  \label{fig:appendix_maglap_high_q}}
\end{figure}

All directed trees are conflict-free, but there also exist conflict-free graphs with cycles. For example, we can construct a conflict-free graph with cycles, if we add self-loops. Alternatively, a graph remains conflict-free, if we add bidirectional edges between (some) pairs of nodes \(u\) and \(v\) that have the same phase, i.e., if \(\angle(\eEigvec_{u, 0}) = \angle(\eEigvec_{v, 0})\).

Moreover, except for numerical issues of the eigendecomposition, also for general graphs (\(\eigval_0 > 0\)) the phase shift is bounded by \(2 \pi q l\). The argument is as the following: Choose an arbitrary pair of nodes \(u\) and \(v\) for which the longest simple path distance is \(l\) (see above). Although there might exist other paths between \(u\) and \(v\), they have at most a simple path length of \(l\). If there exist alternative paths of smaller length \(o < l\), then \(\bar{\eigvec}_0^\top \mL^{(q)} \eigvec_0\) is optimal for a maximum shift in the open interval \((2 \pi q o, 2 \pi q l)\) (here with accounting for overflows in the value range).

\textbf{Visualizations.} We next validate our choice for \(q = \nicefrac{q'}{d_\gG}\) with \emph{relative potential} \(q'\) (see \autoref{sec:spectral}) and graph specific normalizer \(d_\gG = \max(\min(\vec{m}, n), 1)\). In \autoref{fig:appenidx_phase_shift}, we show the imaginary value of the first eigenvector for different exemplary graphs. We see that for the conflict-free sequence with longest path distance \(\vec{m}\), \(\nicefrac{2 \pi}{q'}\) is the total induced phase shift. For all shown graphs, with \(q' \le \nicefrac{1}{4}\) we could reorder the graph nodes using a simple sort operation (up to ties and cycles). We argue that this is a desirable property for directional positional encodings. We demonstrate this ability in \autoref{sec:appendix_maglap_reorder}. Moreover, in \autoref{fig:appendix_maglap_high_q}, we contrast the eigenvectors of the combinatorial Laplacian (a) as well as Magnetic Laplacian with reasonable \(q'=\nicefrac{1}{4}\) (b) to very high values for potential \(q\) (c-d). For the large potentials \(q\), the resulting oscillations in the direction can be of high frequency (relatively to \(n\)). Consequently, it seems neither helpful nor necessary to choose high potential values \(q\). Our graph-specific normalization avoids such degenerate cases.

\subsection{Sign, Scale and Rotation}\label{sec:appendix_maglap_scale_rotate}

As discussed in \autoref{sec:spectral}, if \(\eigvec\) is an eigenvector of \(\mL\) then so is \(c \eigvec\), even if \(c \in \C\) with \(|c| > 0\) (proof: \(c \mL \eigvec = c \eigval \eigvec \implies \mL (c \eigvec) = \eigval (c \eigvec)\)). One way to cope with this is to fix \(c\) s.t.\ the neural network does not need to be invariant w.r.t.\ to arbitrary factors \(c\). We note that this is much more challenging for complex eigenvectors as than it is for real-valued eigenvectors. We give an algorithmic description in \autoref{algo:appendix_norm}.

For the subsequent procedure it is important that the maximum relative phase shift is small enough. Moreover, the used eigensolver chooses a zero imaginary component for the first node (in terms of the adjacency matrix). With out choice of  potential \(q\), the rotation to the first node is within \([-\nicefrac{\pi}{2}, \nicefrac{\pi}{2}]\).

\begin{figure}[H]
\centering
\begin{minipage}{.5\linewidth}
\begin{algorithm}[H]
  \caption{Normalize Eigenvectors}
  \label{algo:appendix_norm}
  \begin{algorithmic}[1]
    \STATE {\bfseries Input:} Eigenvectors \(\Eigvec \in \C^{n \times k}\)
    \STATE \(\vj \leftarrow \operatorname{argmax}(|\Re(\Eigvec)|,\, \operatorname{axis}=0)\)  \hspace{10pt} // shape \(k\)
    \STATE \(\vs \leftarrow \operatorname{sign}(\Re(\Eigvec)[0:n-1, \vj])\)  \hspace{20pt} // shape \(k\)
    \STATE \(\Eigvec \leftarrow \vs \odot \Eigvec\)
    
    \STATE \(j \leftarrow \operatorname{max}(\Im(\Eigvec[:, 0]))\) \hspace{53pt} // scalar
    \STATE \(\alpha \leftarrow \angle(\Eigvec[j, :])\)
    \STATE \(\Eigvec \leftarrow \exp(-i \alpha)^\top \odot \Eigvec\)
    \STATE Return \(\Eigvec\)
  \end{algorithmic}
\end{algorithm}
\end{minipage}
\end{figure}

\textbf{Sign and scale.} For the scale we choose \(c\) for all eigenvectors s.t.\ \(\Eigvec\) is unitary. However, this still allows for choosing the sign / direction as well as rotation. For the sign, we choose maximum real component for each \(\eigvec\) to be positive.

\textbf{Rotation.} If there is an application specific origin, e.g., as in function name prediction, we also use this to choose the rotation relatively to it (i.e., replace line 5 or abort if \(j = 0\)). Otherwise, as in the playground and sorting network tasks, we choose the foremost source node as origin. That is, we search first for the source node with greatest phase \(\argmax_v \angle(\eEigvec_{v, 0}) = u\). Then we use this node as origin. That is, we rotate \emph{all} eigenvectors: \(\angle(\eEigvec_{u, j}) = 0\,,\,\,\forall\,j \in \{0, 1, \dots, n-1\}\).

\subsection{Reorder Permuted Graphs}\label{sec:appendix_maglap_reorder}

\begin{figure}[H]
  \centering
  \makebox[\linewidth][c]{
    \(\begin{array}{cccccc}
      \subfloat[Sequence]{
      \includegraphics[height=2.5cm]{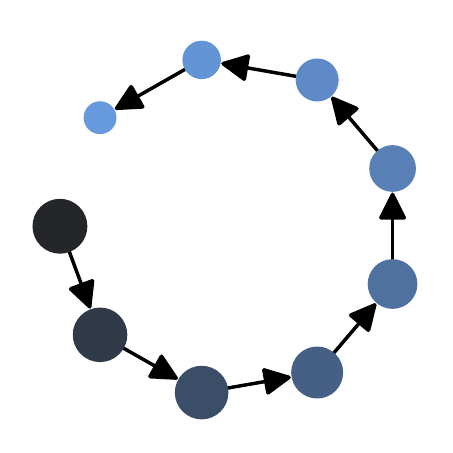}} &
      \subfloat[Rand. permuted]{
      \includegraphics[height=2.5cm]{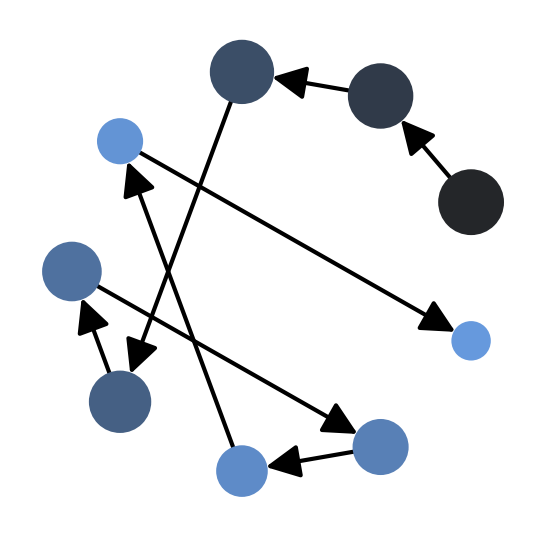}} &
      \subfloat[Rand. permuted]{
      \includegraphics[height=2.5cm]{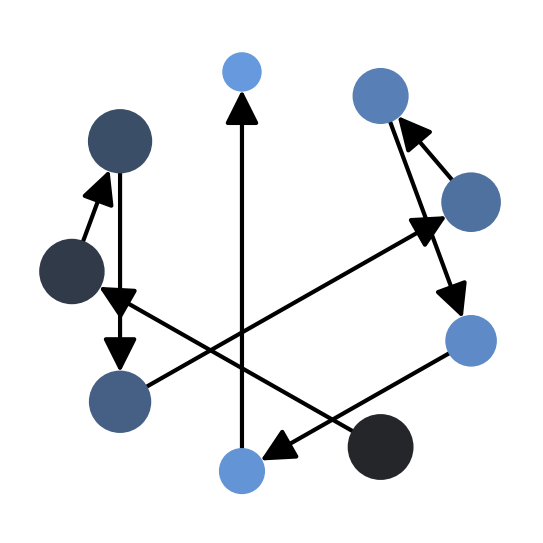}} &
      \subfloat[Rand. permuted]{
      \includegraphics[height=2.5cm]{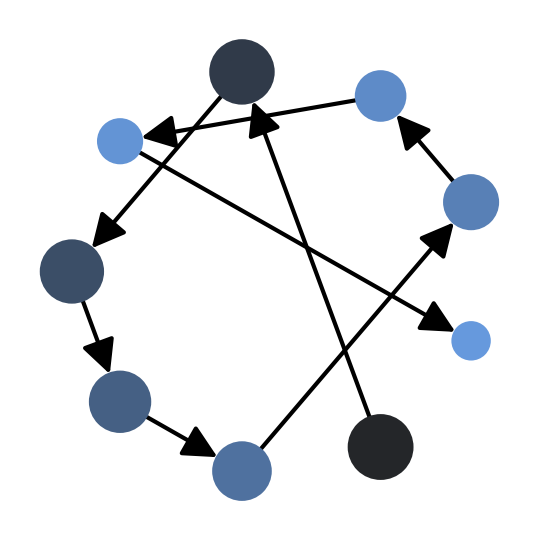}} &
      \subfloat[Rand. permuted]{
      \includegraphics[height=2.5cm]{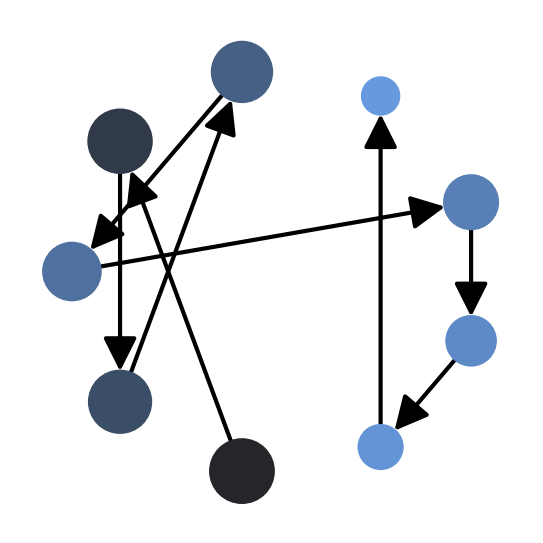}} &
      \subfloat[Rand. permuted]{
      \includegraphics[height=2.5cm]{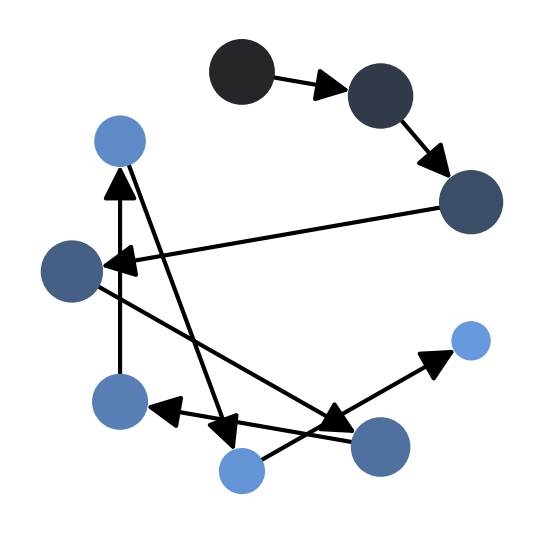}} \\
      
      \subfloat[Sequence]{
      \includegraphics[height=2.5cm]{assets/permuted_png/raw_permuted_sequence_n9_q0.027777777777777776_rel_u_graph_0.pdf}} &
      \subfloat[Reordered]{
      \includegraphics[height=2.5cm]{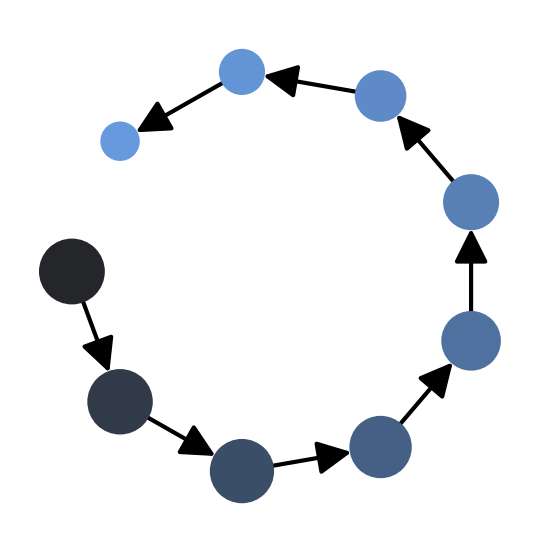}} &
      \subfloat[Reordered]{
      \includegraphics[height=2.5cm]{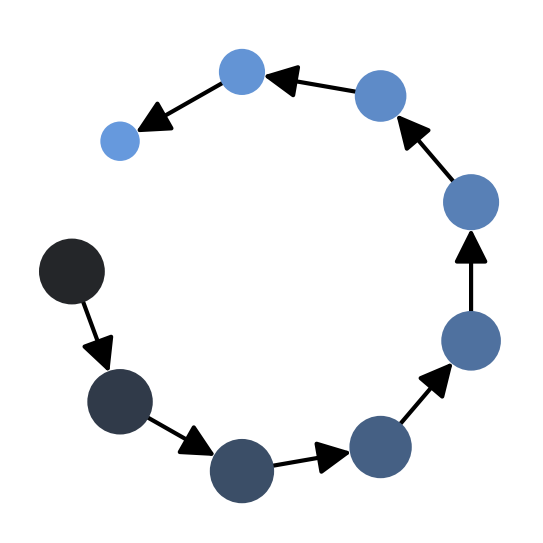}} &
      \subfloat[Reordered]{
      \includegraphics[height=2.5cm]{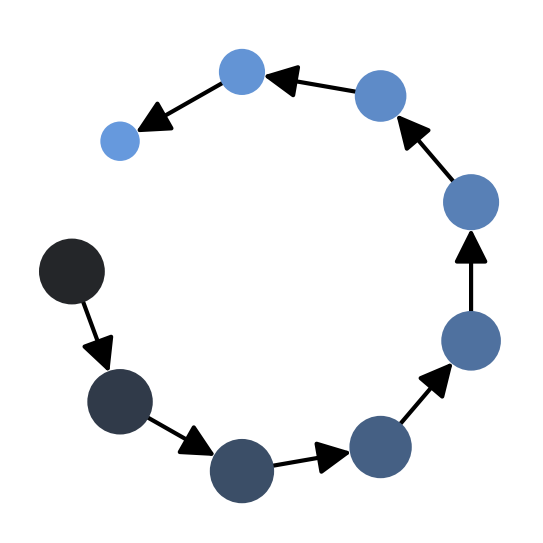}} &
      \subfloat[Reordered]{
      \includegraphics[height=2.5cm]{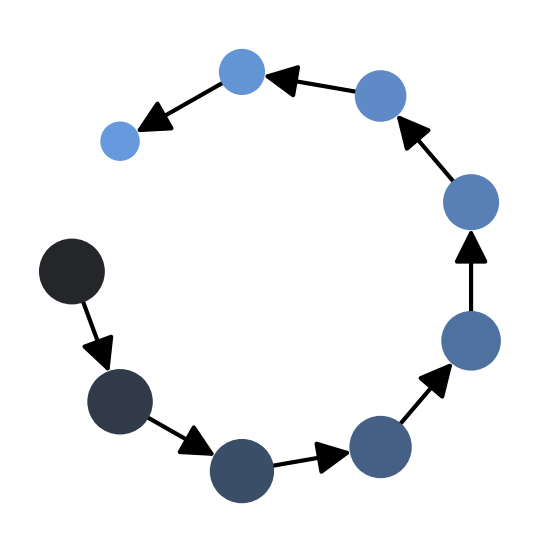}} &
      \subfloat[Reordered]{
      \includegraphics[height=2.5cm]{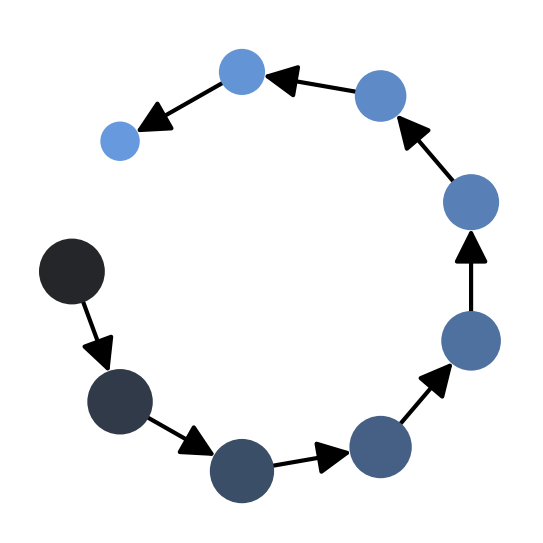}} \\
      \end{array}\)
  }
  \caption{Reordering of randomly permuted sequence \autoref{fig:appenidx_example_graphs:a} with \(\operatorname{idx} = \operatorname{argsort}(\Im(\eigvec_0))\). \label{fig:appendix_maglap_perm_sequence}}
\end{figure}

\begin{figure}[H]
  \centering
  \makebox[\linewidth][c]{
    \(\begin{array}{cccccc}
      \subfloat[Binary tree]{
      \includegraphics[height=2.5cm]{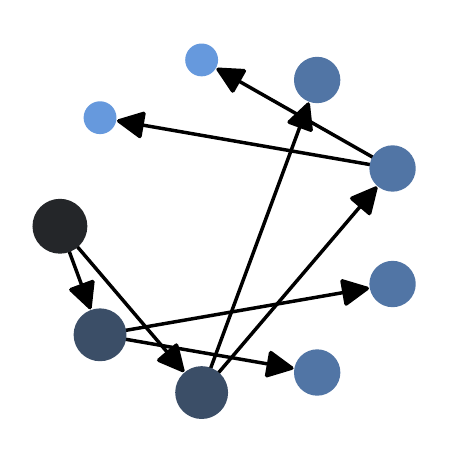}} &
      \subfloat[Rand. permuted]{
      \includegraphics[height=2.5cm]{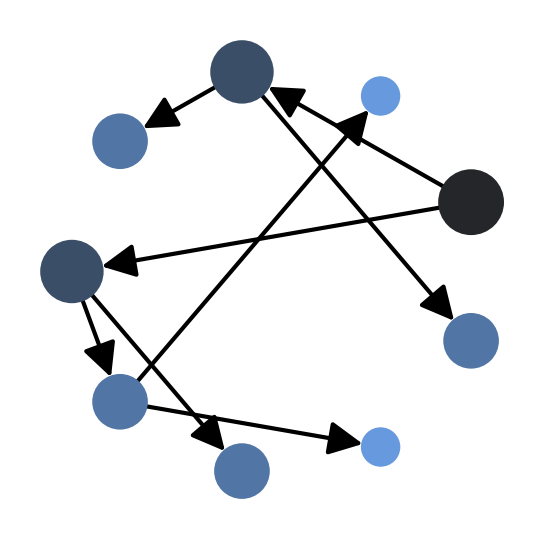}} &
      \subfloat[Rand. permuted]{
      \includegraphics[height=2.5cm]{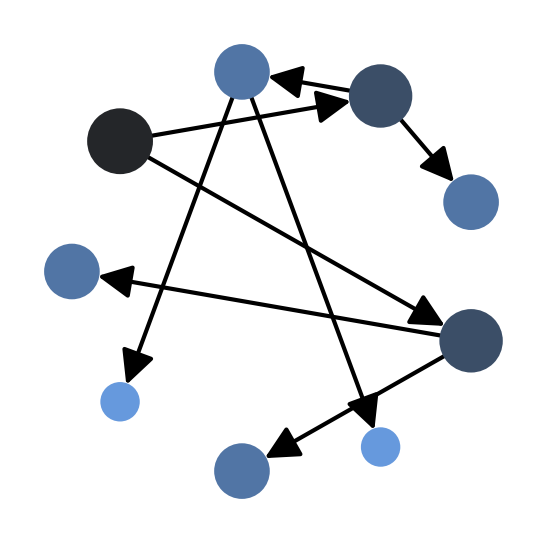}} &
      \subfloat[Rand. permuted]{
      \includegraphics[height=2.5cm]{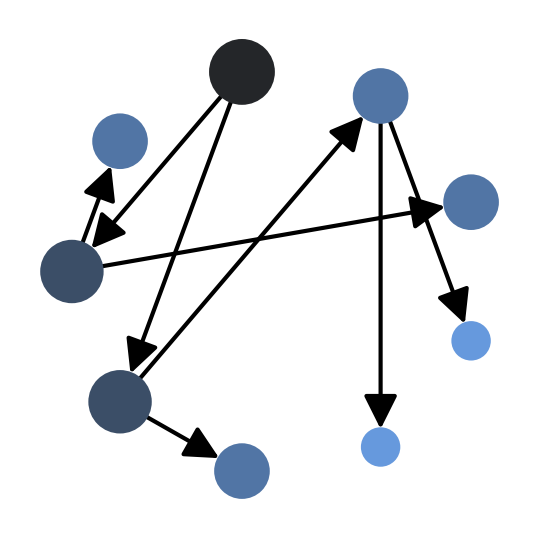}} &
      \subfloat[Rand. permuted]{
      \includegraphics[height=2.5cm]{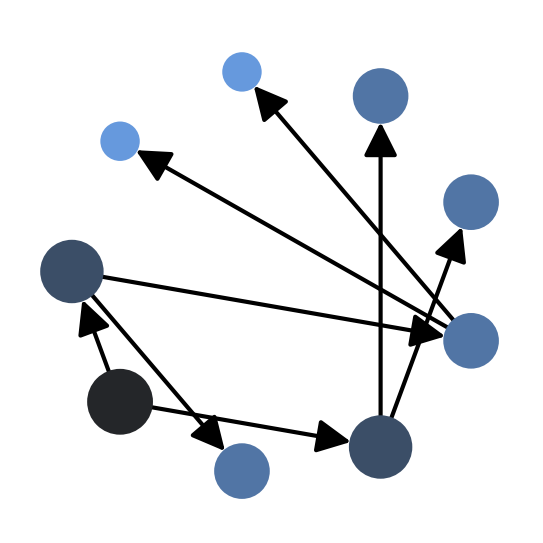}} &
      \subfloat[Rand. permuted]{
      \includegraphics[height=2.5cm]{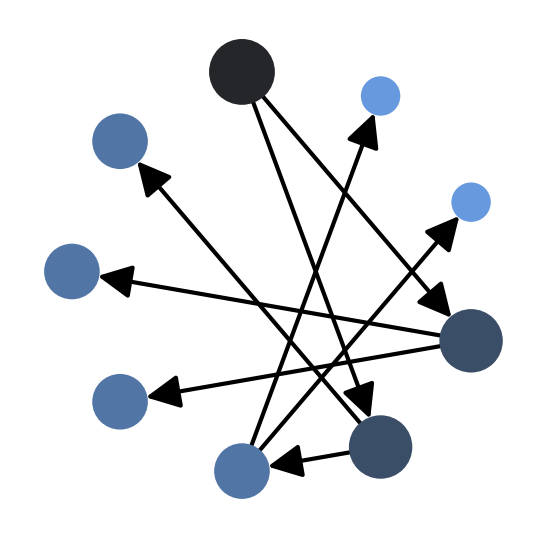}} \\
      
      \subfloat[Binary tree]{
      \includegraphics[height=2.5cm]{assets/permuted_png/raw_permuted_balanced_binary_tree_n9_q0.027777777777777776_rel_u_graph_0.pdf}} &
      \subfloat[Reordered]{
      \includegraphics[height=2.5cm]{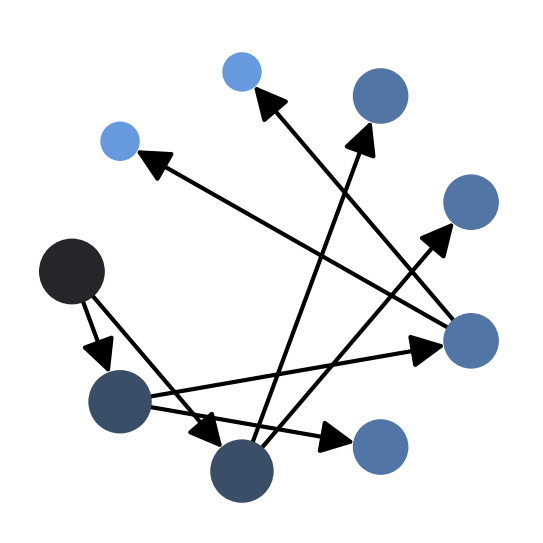}} &
      \subfloat[Reordered]{
      \includegraphics[height=2.5cm]{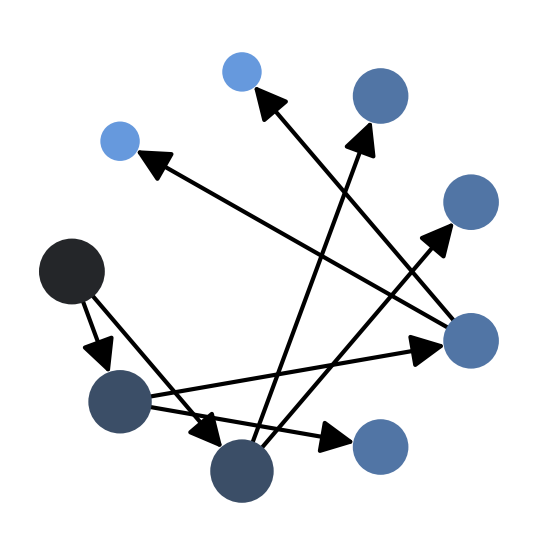}} &
      \subfloat[Reordered]{
      \includegraphics[height=2.5cm]{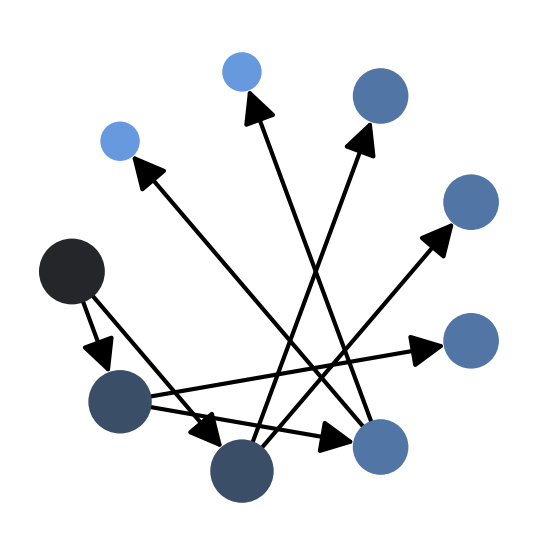}} &
      \subfloat[Reordered]{
      \includegraphics[height=2.5cm]{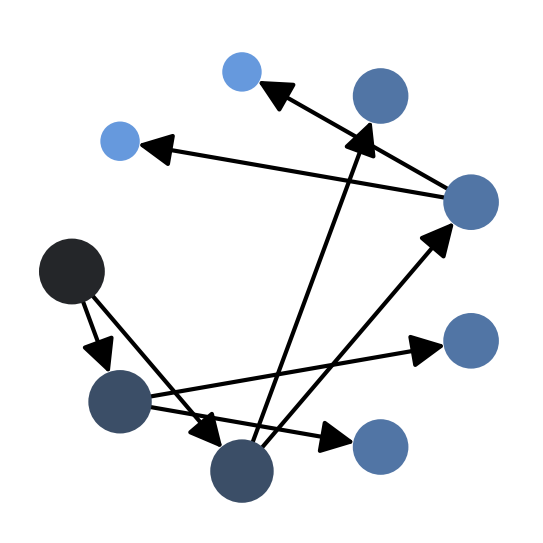}} &
      \subfloat[Reordered]{
      \includegraphics[height=2.5cm]{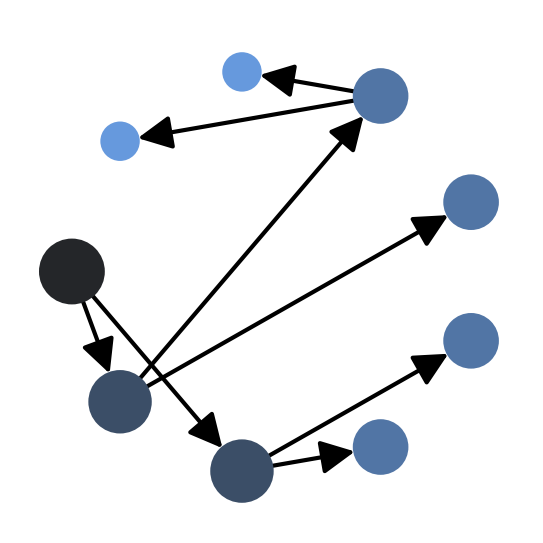}} \\
      \end{array}\)
  }
  \caption{Reordering of randomly permuted binary tree \autoref{fig:appenidx_example_graphs:c} with \(\operatorname{idx} = \operatorname{argsort}(\Im(\eigvec_0))\). \label{fig:appendix_maglap_perm_binary_tree}}
\end{figure}

\begin{figure}[H]
  \centering
  \makebox[\linewidth][c]{
    \(\begin{array}{cccccc}
      \subfloat[Trumpet (fully c.)]{
      \includegraphics[height=2.5cm]{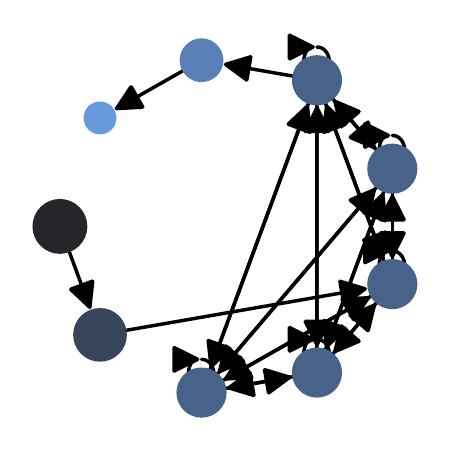}} &
      \subfloat[Rand. permuted]{
      \includegraphics[height=2.5cm]{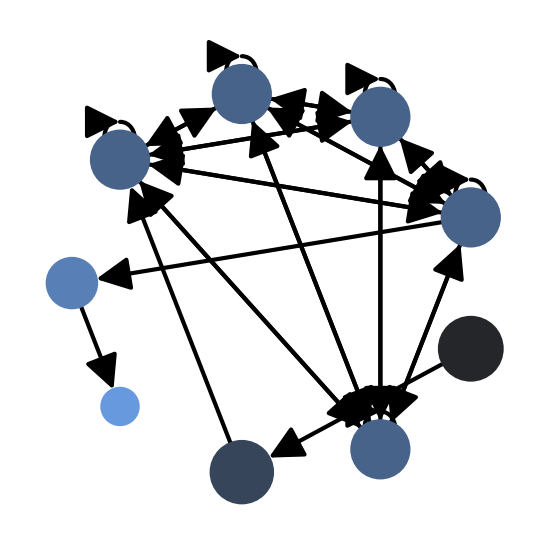}} &
      \subfloat[Rand. permuted]{
      \includegraphics[height=2.5cm]{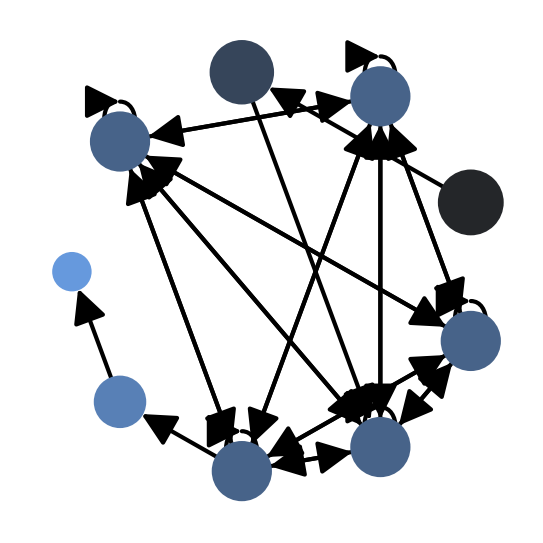}} &
      \subfloat[Rand. permuted]{
      \includegraphics[height=2.5cm]{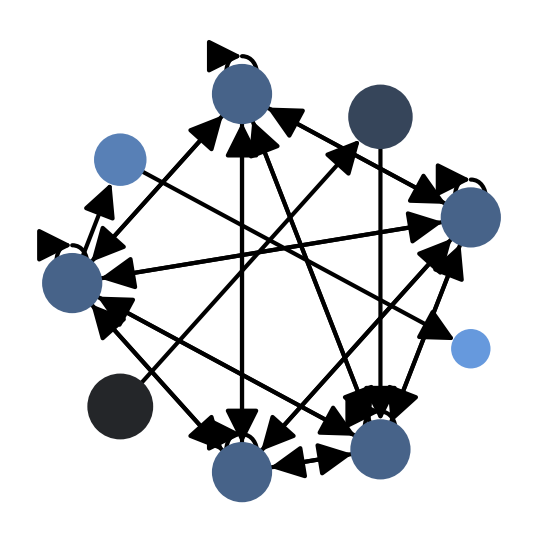}} &
      \subfloat[Rand. permuted]{
      \includegraphics[height=2.5cm]{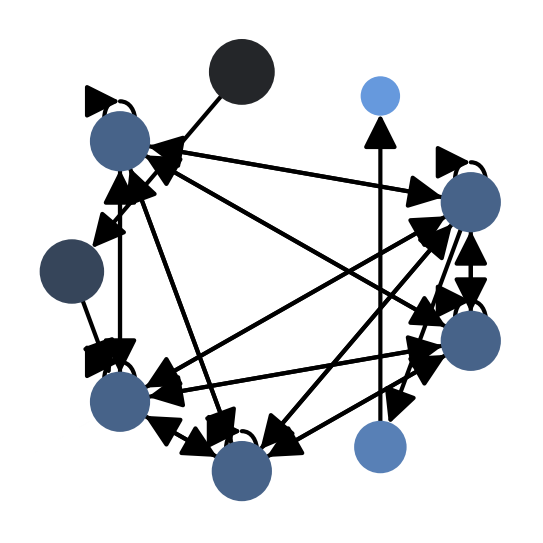}} &
      \subfloat[Rand. permuted]{
      \includegraphics[height=2.5cm]{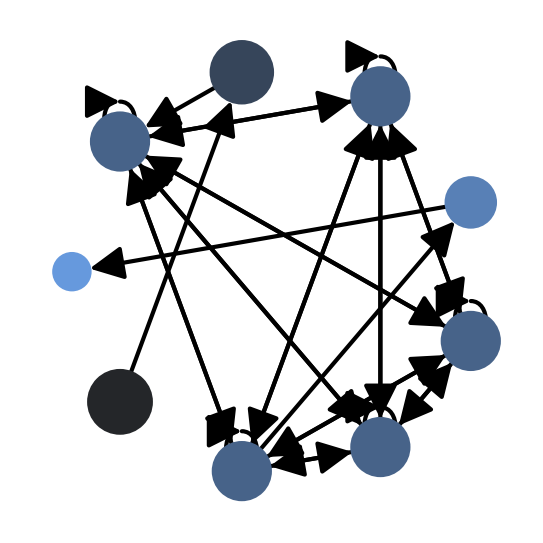}} \\
      
      \subfloat[Trumpet (ful. c.)]{
      \includegraphics[height=2.5cm]{assets/permuted_png/raw_permuted_belly_snake_n9_q0.027777777777777776_rel_u_graph_0.pdf}} &
      \subfloat[Reordered]{
      \includegraphics[height=2.5cm]{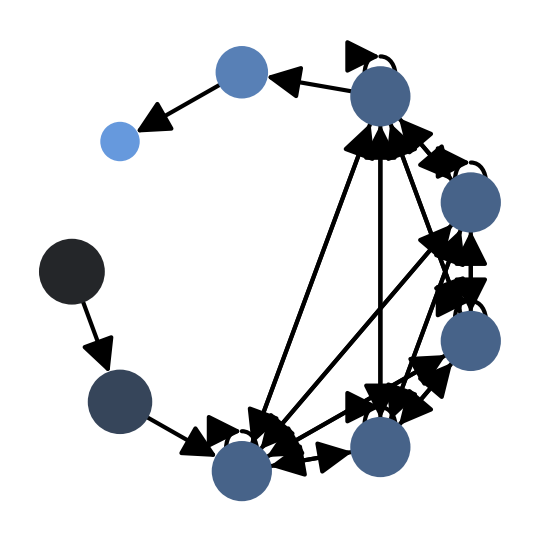}} &
      \subfloat[Reordered]{
      \includegraphics[height=2.5cm]{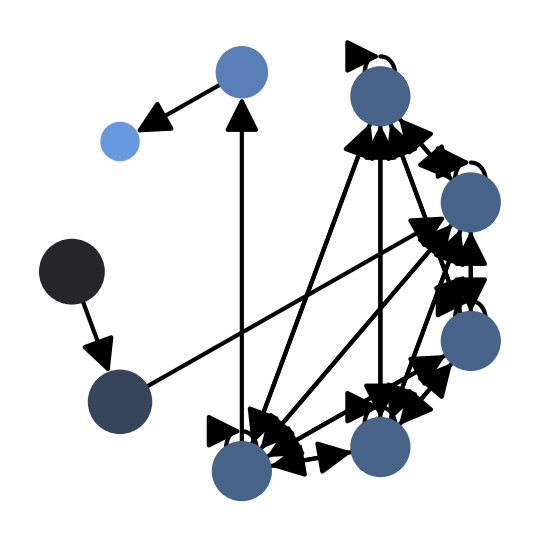}} &
      \subfloat[Reordered]{
      \includegraphics[height=2.5cm]{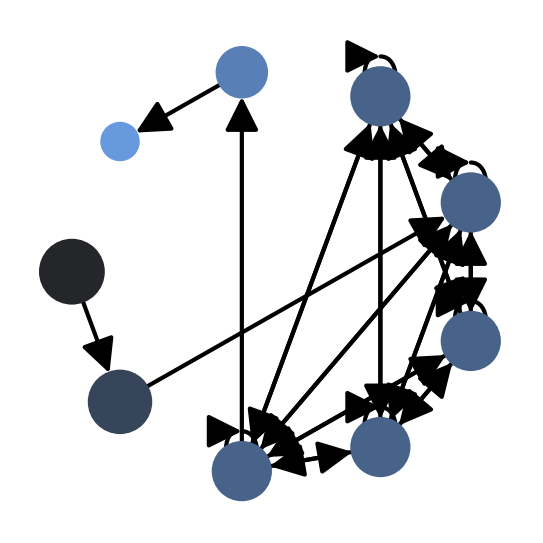}} &
      \subfloat[Reordered]{
      \includegraphics[height=2.5cm]{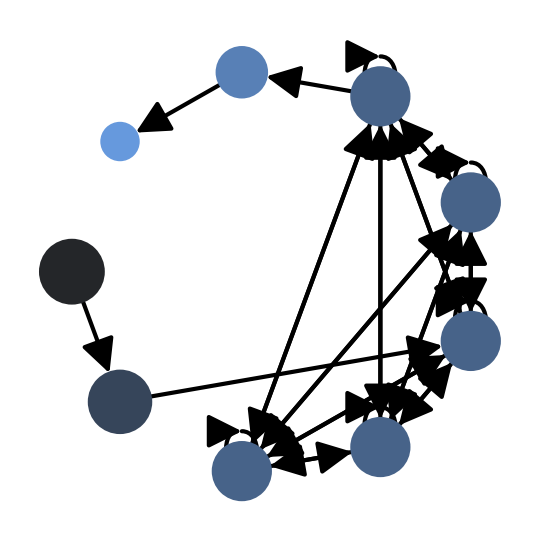}} &
      \subfloat[Reordered]{
      \includegraphics[height=2.5cm]{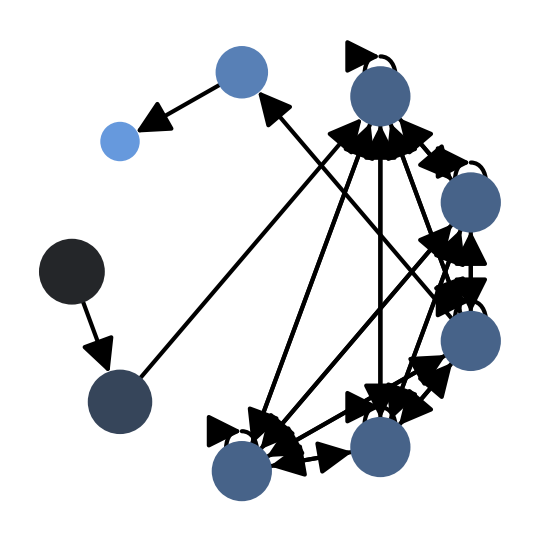}} \\
      \end{array}\)
  }
  \caption{Reordering of randomly permuted trumpet with fully connected middle part \autoref{fig:appenidx_example_graphs:l} with \(\operatorname{idx} = \operatorname{argsort}(\Im(\eigvec_0))\). \label{fig:appendix_maglap_perm_belly_snake}}
\end{figure}

We can also use the Magnetic Laplacian for reordering permuted graphs, up to ties. For example, we have ties in a balanced binary tree (\autoref{fig:appenidx_example_graphs:c} or \ref{fig:appenidx_example_graphs:g}) stemming from the equal distance to the root node. For the three exemplary graphs in \autoref{fig:appendix_maglap_perm_sequence},  \ref{fig:appendix_maglap_perm_binary_tree}, and  \ref{fig:appendix_maglap_perm_belly_snake} we show that the eigenvectors of the Magnetic Laplacian can be used to reorder directed graphs. First, we permute the nodes arbitrarily, perform the eigendecomposition, and, lastly reorder the graphs after applying our normalization scheme for the eigenvectors (\autoref{sec:appendix_maglap_scale_rotate}). We na\"ively recover the order by \(\operatorname{idx} = \operatorname{argsort}(\Im(\eigvec_0))\).

\subsection{Comparison to Singular Value Decomposition}\label{sec:appendix_maglap_svd}

One could also use the Singular Value Decomposition (SVD) \(\mU \boldsymbol{\Sigma} \mV\) to obtain structure-aware positional encodings. Specifically, \citet{hussain_global_2022} argue that a low-rank approximation (via SVD) of the adjacency matrix yields general positional encodings that are also suitable for directed graphs (see \autoref{sec:appendix_playground} for an empirical comparison). In contrast to the eigendecomposition, the singular values and singular vectors are real also if decomposing asymmetric matrices. However, it is questionable if the SVD of the adjacency matrix has desirable properties.

For example, a low-rank approximation of the adjacency matrix of a directed sequence \autoref{fig:appenidx_example_graphs:a} simply filters out some of the edges. For example, a 2-rank approximation for a directed sequence with \(n=5\) nodes is
\begin{equation}
    \mA = \resizebox{!}{0.7\height}{\(\begin{bmatrix}
    0 & \colorbox{yellow}{\(\mathbf{1}\)} & 0 & 0 & 0 \\
    0 & 0 & 1 & 0 & 0 \\
    0 & 0 & 0 & 1 & 0 \\
    0 & 0 & 0 & 0 & \colorbox{yellow}{\(\mathbf{1}\)} \\
    0 & 0 & 0 & 0 & 0
    \end{bmatrix} \)} \approx \resizebox{!}{0.7\height}{\(
    \begin{bmatrix}
    0 & 0 \\
    1 & 0 \\
    0 & 1 \\
    0 & 0 \\
    0 & 0 \\
    \end{bmatrix}
    \begin{bmatrix}
    1 & 0 \\
    0 & 1 \\
    \end{bmatrix}
    \begin{bmatrix}
    0 & 0 & 1 & 0 & 0 \\
    0 & 0 & 0 & 1 & 0 \\
    \end{bmatrix} \)} = \resizebox{!}{0.7\height}{\( \begin{bmatrix}
    0 & \colorbox{yellow}{\(\mathbf{0}\)} & 0 & 0 & 0 \\
    0 & 0 & 1 & 0 & 0 \\
    0 & 0 & 0 & 1 & 0 \\
    0 & 0 & 0 & 0 & \colorbox{yellow}{\(\mathbf{0}\)} \\
    0 & 0 & 0 & 0 & 0
    \end{bmatrix}\)}
\end{equation}

One could obtain appropriate results if using well-suited matrices like the combinatorial Laplacian for undirected graphs or the Magnetic Laplacian for directed graphs. However, then the alleged advantage of the SVD for directed graphs vanishes. That is, the singular vectors of a Hermitian Matrix are complex \(\mU \in \C^{n \times n}\) and \(\mV \in \C^{n \times n}\) and not real. Moreover, one needs to account then for the (slightly different) properties of the SVD, like its sign ambiguity \(\mU \boldsymbol{\Sigma} \mV = (\shortminus \mU) \boldsymbol{\Sigma} (\shortminus \mV)\).

\section{Weighted Graphs}\label{sec:appendix_weighted}

Our positional encodings based on random walks as well as the Magnetic Laplacian straight-forwardly generalize to weighted graphs. Here the adjacency matrix \(\mA \in \R_{\ge 0}^{n \times n}\) contains positive real-valued weights. For the extension of the Magnetic Laplacian to signed graphs, i.e., also allowing negative edge weights, we refer to \citet{he_msgnn_2022}.

For the random walk encodings, all equations in \autoref{sec:random_walks} are directly applicable, however, for the (Magnetic) Laplacian one needs to adjust the strategy for symmetrizing the graph and potentially how to choose potential \(q\). Then, for weighted graphs, the smallest eigenvalue of the Magnetic Laplacian solves
\begin{equation}\label{eq:appendix_maglap_raleigh}
    \min_{\vx \in \C} \frac{\bar{\vx}^\top \mL^{(q)} \vx}{\bar{\vx}^\top \vx}
    = \frac{1}{2} \sum_{(u,v) \in E_s} w_{u, v} | \eEigvec_{u, 0} - \eEigvec_{v, 0} \exp(\Theta^{(q)}_{u, v}) | ^2
\end{equation}
where \(w_{u, v}\) is the weight for edge \((u, v)\) in the then weighted symmetrized adjacency matrix \(\mA_s\).

\section{Overview of Our Positional Encodings}\label{sec:appendix_overview}

We next provide a concise overview over the positional encodings. In Algorithm~\ref{algo:appendix_maglapnet_complete}, we provide an overview for the Magnetic Laplacian (\autoref{sec:spectral}). Thereafter, we list the full equation for the random walk encodings (\autoref{sec:random_walks}). In both cases, we add the obtained positional encodings to the node features. For the relative random walk encodings see \autoref{sec:appendix_playground}.

\begin{figure}[H]
\centering
\begin{minipage}{.7\linewidth}
\begin{algorithm}[H]
  \caption{Magnetic Laplacian Positional Encodings}
  \label{algo:appendix_maglapnet_complete}
  \begin{algorithmic}[1]
    \STATE {\bfseries Input:} Adjacency matrix \(\mA \in \{0, 1\}^{n \times n}\), potential \(q\), number of eigenvectors \(k\)
    \STATE Calculate \(\mL^{(q)} \leftarrow \operatorname{Laplacian}(\mA, q)\)   \hspace{67pt} // e.g.\ \autoref{eq:unsymmetric_magnetic_laplacian}
    \STATE Decompose \(\Eigval, \Eigvec \leftarrow \operatorname{eigh}(\mL^{(q)}, k)\)
    \STATE Normalize \(\Eigvec, \Eigval \leftarrow \operatorname{norm}(\Eigvec, \Eigval)\) according to Algorithm~\ref{algo:appendix_norm}
    \STATE Obtain preprocessed \(\hat{\Eigvec} \leftarrow \operatorname{MagLapNet}(\Eigvec, \Eigval)\) \hspace{20pt} // see \autoref{fig:maglapnet_with_trans:b}
    \STATE Return \(\hat{\Eigvec}\)
  \end{algorithmic}
\end{algorithm}
\end{minipage}
\end{figure}

The finite-horizon random walk and Personalized Page Rank landing probabilities are processed together as
\begin{equation}
    \zeta(v | u) = f^{(2)}_{\text{rw}}[\Pi(\mR)_{v,u}, (\mR^k)_{v,u}, \dots, (\mR^2)_{v,u}, \emR_{v,u}, \emT_{v,u}, (\mT^2)_{v,u}, \dots, (\mT^k)_{v,u}, \Pi(\mT)_{v,u}]
\end{equation}
where \(\Pi(\mT) = p_r (\mI - (1 - p_r) \mT)^{-1}\). Then, the node encodings (here for node \(v\)) are given as 
\begin{equation}
\zeta(v | \gG) = f_{\text{rw}}^{(1)}(\operatorname{AGG}(\{\zeta(v | u)\,|\,u \in V\}))
\end{equation}

\section{Experimental Setup}\label{sec:appendix_setup}

We use the AdamW optimizer~\citep{loshchilov_decoupled_2019} and employ early stopping, using the validation data. Moreover, we apply adaptive gradient clipping (AGC) as proposed by \citet{brock_high-performance_2021} and decay the learning rate with cosine annealing~\citep{loshchilov_sgdr_2017}. We report peak learning rate in \autoref{tab:appendix_hyperparameters}. For the playground classification tasks \autoref{sec:playground}, we train on one Nvidia GeForce GTX 1080TI with 11 GB RAM. Regression as well as sorting network results are obtained with a V100 with 40 GB RAM. For training the models on function name prediction dataset, we used four Google Cloud TPUv4 (behaves like 8 distributed devices). For eased reproduction of results, we also provide configuration for a single V100. In the single device setup, training with precomputed eigenvectors and eigenvalues of the Magnetic Laplacian requires less than 4 hours.

\begin{table}[t]
\centering
\caption{Most important hyperparameters for tasks and models\label{tab:appendix_hyperparameters}}
\resizebox{\linewidth}{!}{
\begin{tabular}{cccl}
    \toprule
     & \multicolumn{2}{c}{\textbf{Model}} & \textbf{Hyperparameter} \\
    \midrule
    \multirow{4}{*}{\rotatebox[origin=c]{90}{\begin{tabular}[c]{@{}c@{}}\textbf{Playground and}\\\textbf{Sorting Network}\end{tabular}}} & \multicolumn{2}{l}{Base} & 
    \begin{tabular}[l]{@{}l@{}}
    15 epochs sorting network / 30 epochs playground, \\
    learning rate \(\eta \approx \num{8.3e-6} \times \text{batch size}\), weight decay \(\alpha = \num{6e-5}\),  \\
    Adam \(\beta_1=0.7\), \(\beta_2=0.9\), AGC clipping \num{7.5e-2}
    \end{tabular} \\  \cline{3-4}
     & + & \begin{tabular}[c]{@{}c@{}}Combinatorial\\ Laplacian\end{tabular} &  
     \begin{tabular}[l]{@{}l@{}}Top \(k=25\) eigenvectors (degree normalized~\autoref{eq:laplacian_s}) for sorting network and \(k=16\) otherwise, \\ dropout \(p_{\text{pos}} = 0.15\), w/o SignNet \end{tabular}
     \\  \cline{3-4} 
     & + & \begin{tabular}[c]{@{}c@{}}Magnetic\\  Laplacian\end{tabular} &  
     \begin{tabular}[l]{@{}l@{}}Top \(k=25\) eigenvectors (deg.\ norm.) for sorting network\ and \(k=16\) otherwise,\\rel. potential \(p' = \nicefrac{1}{4}\),  dropout \(p_{\text{pos}} = 0.15\), w/o SignNet \end{tabular}
     \\ \cline{3-4} 
     & + & \begin{tabular}[c]{@{}c@{}}Random\\ Walk\end{tabular} &  
     \(k=3\) random walk steps, personalized page rank restart probability \(p_r = 0.05\)
     \\ \hline

    \multirow{6}{*}{\rotatebox[origin=c]{90}{\textbf{OGB Code2}}} & \multicolumn{2}{l}{Sequential data} & 
    \begin{tabular}[l]{@{}l@{}}
    15 epochs, learning rate \(\eta /approx \num{5.4e-5} \times \text{batch size}\), weight decay \(\alpha = \num{7.5e-5}\),  \\
    Adam \(\beta_1=0.75\), \(\beta_2=0.935\), AGC clipping \num{0.1}, 3 mess. passing steps
    \end{tabular} \\  \cline{3-4}
     & \multicolumn{2}{l}{Our dataset constr.} & 
    \begin{tabular}[l]{@{}l@{}}
    32 epochs, learning rate \(\eta = \num{5.4e-5} \times \text{batch size}\), weight decay \(\alpha = \num{6e-5}\),  \\
    Adam \(\beta_1=0.9\), \(\beta_2=0.95\), AGC clipping \num{5e-2}, 3 mess. passing steps
    \end{tabular} \\  \cline{3-4}
     & + & \begin{tabular}[c]{@{}c@{}}Magnetic\\ Laplacian\end{tabular} & 
     Top \(k=25\) eigenvectors (deg. norm.), rel. potential \(p' = \nicefrac{1}{4}\),  dropout \(p_{\text{pos}} = 0.15\), SignNet w/ GNN 
     \\ \cline{3-4} 
     & + & \begin{tabular}[c]{@{}c@{}}Random\\ Walk\end{tabular} &  
     \(k=3\) random walk steps, personalized page rank restart probability \(p_r = 0.05\)
     \\
    \bottomrule
\end{tabular}
}
\end{table}

\begin{wrapfigure}[21]{r}{0.55\textwidth}
  \centering
  \vspace{-20pt}
  \subfloat[W/o GNN\label{fig:appendix_architecture:a}]{\includegraphics[height=4.8cm]{assets/VanillaTransformerArchitectureColored.pdf}}
  \subfloat[W/ GNN\label{fig:appendix_architecture:b}]{\includegraphics[height=4.8cm]{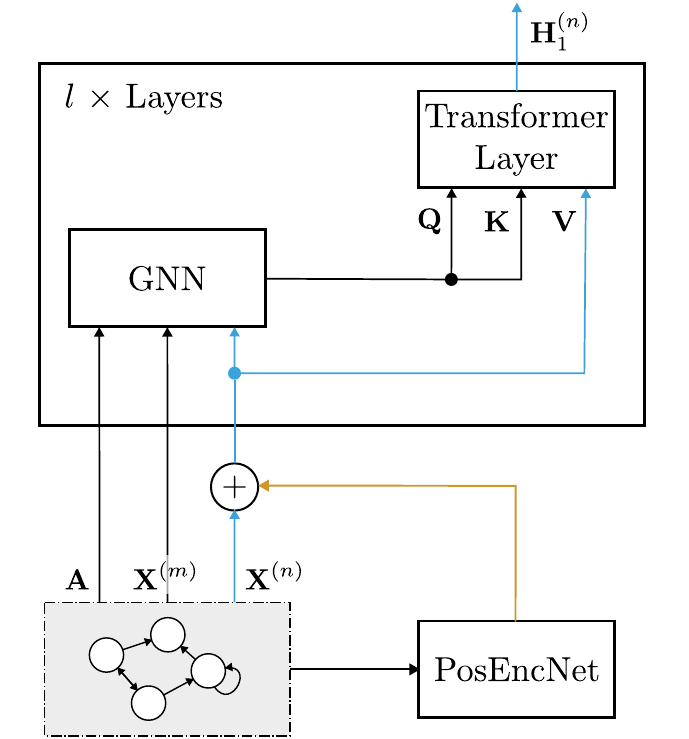}}
  \caption{Architectures of structure-aware transformers that we study in this work. (a) solely relies on positional encodings for structure awareness and (b) resembled the Structure Aware Transformer (SAT) \citep{chen_structure-aware_2022}. Analogously, to their architecture for the OGB Code2 dataset, here we omit the subgraph sampling. We only show the first of \(l\) layers. Subsequent layers \(\mH^{(j)}\) for \(1 \le j \le l\) operate on the input data besides the node embeddings. \(\mH^{(l)}\) is used for the downstream task.}
  \label{fig:appendix_architecture}
\end{wrapfigure}
\textbf{Models.} We study two architectures: \autoref{fig:appendix_architecture:a} a standard transformer encoder that relies on the positional encodings for structure awareness and \autoref{fig:appendix_architecture:b} a hybrid GNN transformer architecture, also called Structure Aware Transformer~\citep{chen_structure-aware_2022}. The latter is motivated via the connection of the self-attention to kernels and the GNN resembles here a learnable graph kernel. Additionally, the Structure Aware Transformer uses the node degree for weighting the residual connection around self-attention in each transformer layer.

\textbf{GNN architecture} We follow the generic GNN ``framework'' of \citet{battaglia_relational_2018} w/o a global state. Their framework subsumes most major spatial message-passing schemes. We tested various variants but only report the best-performing model. The used GNN alternatingly updates edge embeddings $\mathbf{e}_p^{(l)}$ and node embeddings $\mathbf{v}_j^{(l)}$, with layer $l$, node index $v \in V$, and edge index $p$ from the set of edges $(u, v) \in E$. Specifically, $\mathbf{e}_p^{(l)} = f_e(\mathbf{e}_p^{(l-1)} || \sum_{k \in \mathcal{V}_{\leftarrow}(p)} \mathbf{v}_k^{(l-1)} || \sum_{k \in \mathcal{V}_{\rightarrow}(p)} \mathbf{v}_k^{(l-1)})$ and $\mathbf{v}_j^{(l)} = f_v(\mathbf{v}_j^{(l-1)} || \sum_{k \in \mathcal{E}_{\leftarrow}(j)} \mathbf{e}_k^{(l)} || \sum_{k \in \mathcal{E}_{\rightarrow}(j)} \mathbf{e}_k^{(l)})$ with concatenation $||$ and the sets of incoming nodes $\mathcal{V}_{\leftarrow}(p)$ as well as outgoing nodes $\mathcal{V}_{\rightarrow}(p)$ of edge $p$. Respectively, $\mathcal{E}_{\leftarrow}(j)$ and $\mathcal{E}_{\rightarrow}(j)$ are the sets of incoming and outgoing edges of node $j$. $f_e$ and $f_v$ are MLPs. For the undirected GNN we sum up forward and backward messages instead of concatenating them.

\textbf{Batching.} The maximum batch size per device is 48, except for the tasks in \autoref{sec:playground}, where we use a batch size of 96. Since we use JAX for our experiments, we have constraints on the tensor shape variations. Thus, in each batch, we consider graphs of similar size to avoid excessive padding. Moreover, we increase the batch size \(4 \times\) and \(2 \times\) if \(n < 256\) and \(n < 512\), respectively.

\textbf{Hyperparameters.} We choose the hyperparameters for each model based on a random search over the important learning parameters like learning rate, weight decay, and the parameters of AdamW (30 random draws for the sorting network and 50 for OGB). Due to the small and mostly insignificant differences, we consolidated the parameters for both architectures (\autoref{fig:appendix_architecture}) and all positional encodings. That is, hyperparameters unspecific to the respective positional encoding are shared. Moreover, for the results in \autoref{sec:playground}, we use the parameters for the sorting network task (\autoref{sec:sorting_networks}) without additional tuning. We list the important hyperparameters in \autoref{tab:appendix_hyperparameters}.

\section{Scalability}\label{sec:appendix_scalablity}

Both positional encodings, namely random walks (\autoref{sec:random_walks}) and eigenvectors of Magnetic Laplacian (\autoref{sec:spectral}) can be calculated efficiently. Although, for the random walk encodings one has to be cautious since even for a small number of steps \(k\) the transition matrix becomes practically dense (complexity \(\mathcal{O}(n^2)\)) if not using some sparsification technique. For the scalability of personalized page rank encodings within a neural netowork, we refer to \citep{bojchevski_scaling_2020}. 

Moreover, we only study the standard self-attention that scales with \(\mathcal{O}(n^2)\). Scalable alternatives were extensively studied, e.g., \citep{choromanski_rethinking_2020, kitaev_reformer_2020}, also covering the graph domain \citep{dwivedi_generalization_2021, rampasek_recipe_2022, wu_nodeformer_2022}. It is straightforward to apply our positional encodings to most of these approaches.

\begin{figure}[H]%
    \centering
    \includegraphics[width=0.7\linewidth]{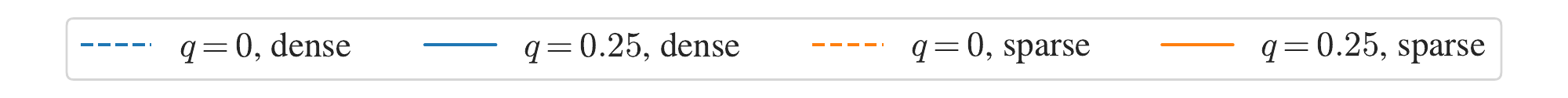}\\
    \subfloat[Intel(R) Xeon(R) CPU @ 2.20GHz]{
    \includegraphics[width=0.35\linewidth]{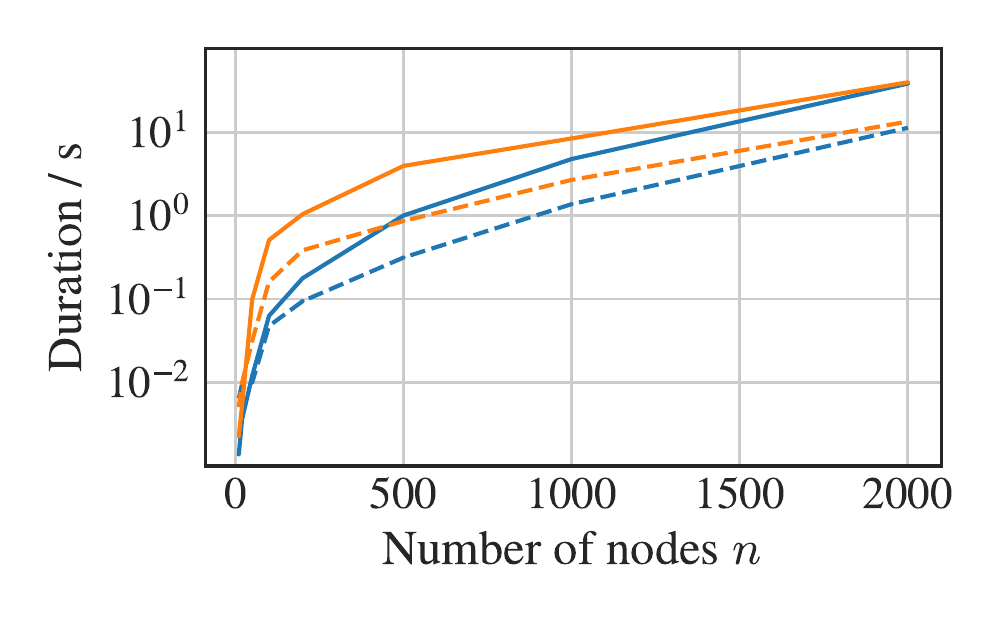}}
    \subfloat[Nvidia Tesla T4 GPU]{
    \includegraphics[width=0.35\linewidth]{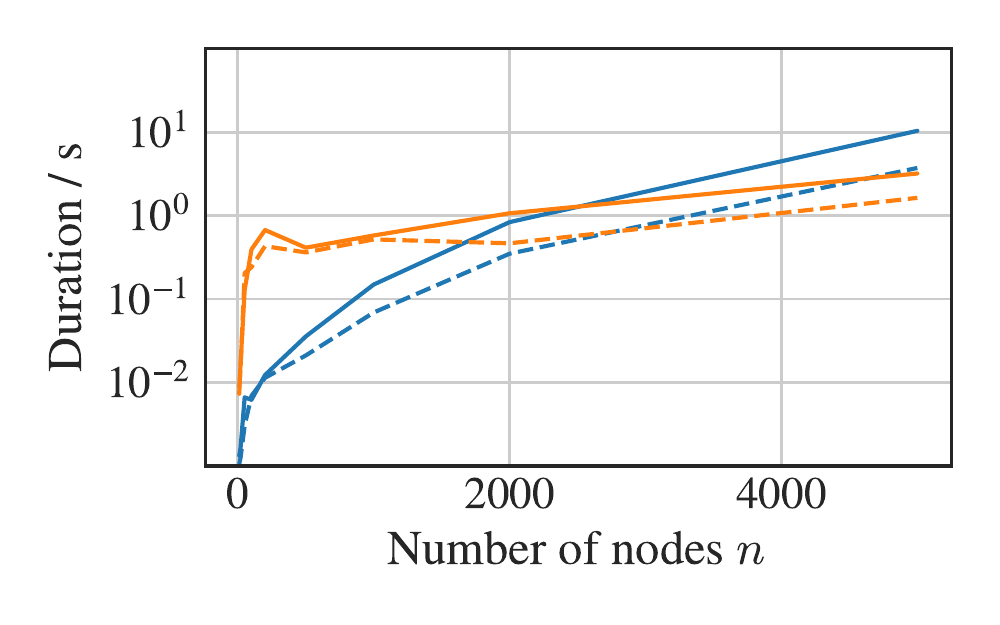}}
    \caption{Runtime required for eigendecomposition on random Erd\H{o}s-R\`enyi graphs.}
    \label{fig:appendix_runtime_eigendecomposition}
\end{figure}

In \autoref{fig:appendix_runtime_eigendecomposition}, we study the duration of the eigendecompositon on a CPU (\verb|scipy|) and GPU (\verb|cupy|) over graphs of different size. We also contrast the overhead of having a Hermitian matrix (\(q=0.25\)) in contrast to the decomposition of a real matrix (\(q=0\)) and in which cases a sparse calculation (with \(k=25\)) is beneficial over its dense equivalent. For this benchmark, we draw random Erd\H{o}s-R\`enyi graphs with an average degree of 5 (similar to the positional encodings playground \autoref{sec:playground}) and report the average over 10 trials via \verb|timeit| or \verb|cupyx.profiler.benchmark|. 

In any case, once \(q\) and other hyperparameters have been chosen it is beneficial to precompute the eigenvectors for training. With precomputed eigenvectors, the training on OGB Code2 (32 epochs) finishes within 4 hours using a single V100.

\clearpage
\section{Positional Encodings Playground}\label{sec:appendix_playground}

\begin{wrapfigure}[35]{r}{0.675\textwidth}
    \vspace{-10pt}
    \centering
    \includegraphics[width=\linewidth]{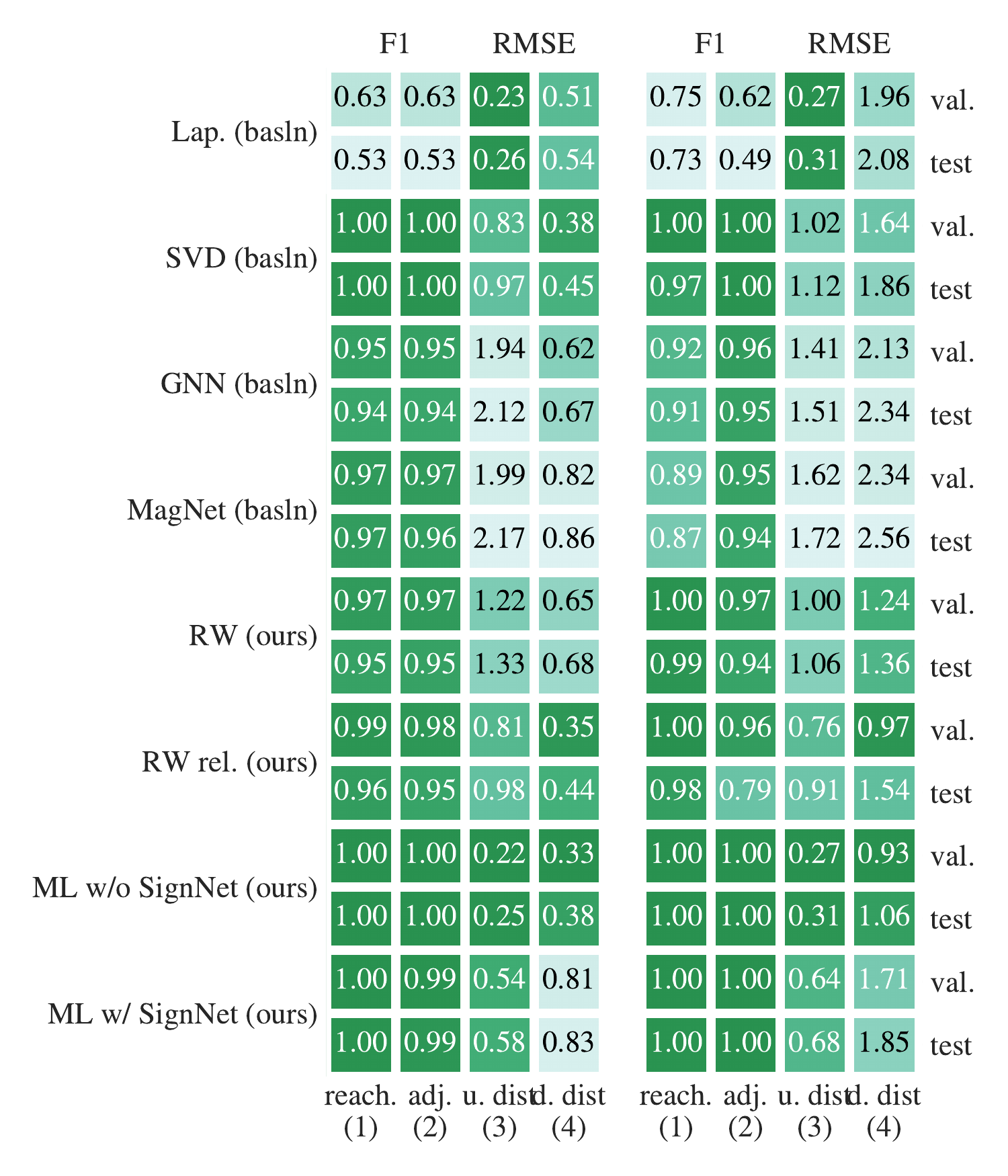}
    \caption{Complimentary results to \autoref{fig:playground} for \emph{(1) reachability}, \emph{(2) adjacency}, \emph{(3) undirected distance}, and \emph{(4) directed distance}.}
    \label{fig:appendix_playground_all}
\end{wrapfigure}

\textbf{GNNs.} Additionally to the baselines of Laplacian and SVD encodings, we also compare to our default GNN architecture (see \autoref{sec:appendix_setup}) and MagNet~\citep{zhang_magnet_2021}. MagNet is a GNN for directed graphs that uses the Magnetic Laplacian in its message passing. Specifically, they formulate each message passing step as a polynomial of the Magnetic Laplacian. We follow the default parameters of the authors and choose (absolute) potential \(q = 0.1\), which is performs best out of \(q \in \{0, 0.05, 0.1, 0.15, 0.2, 0.25\}\). We find that both GNNs are predictive for classifying reachability and adjacency, however, fall behind in the distance regression tasks (see \autoref{fig:appendix_playground_all}).

\textbf{Relative random walk encodings} can be constructed following the recipe in \autoref{sec:random_walks}. However, instead of aggregating the encodings to a node-level, we keep the \(n \times n\) encodings \(\zeta(v | u)\) using the a different linear transformation \(f^{(2)}_{\text{rw}}\) for each transformer layer and head. The attention mechanism for each head becomes \(\operatorname{softmax}(\nicefrac{\mQ\mK^\top}{\sqrt{d}} + \boldsymbol{\zeta}) \mV\), where \(\boldsymbol{\zeta} \in \R^{n \times n}\) is the matrix containing the one-dimensional \(\zeta(v | u)\). Note tha these positional encodings can be understood as a generalization of the pair-wise shortest path distances used by \citet{ying_transformers_2021} and \citet{guo_graphcodebert_2021}, if using a sufficiently large number of random walk steps.

In \autoref{fig:appendix_playground_all}, we show that these relative positional random walk encodings (RW rel.) consistently outperform node-level random walk encodings (RW). However, the relative positional encodings seem to be more prone to overfitting and training becomes less stable. This can, e.g., be seen in the comparably large differences between validation and test scores. Recall that the validation set is more similar to the training set than the test set due to the different number of nodes. Due to the brittleness and the additional tuning required, we do not study the relative positional encodings in the other tasks.

\textbf{SignNet with MLP.} Additionally to the Magnetic Laplacian encodings w/o SignNet (as presented in \autoref{fig:playground}), we report results w/ SignNet \(f_{\text{elem}}(-\eigvec_j) + f_{\text{elem}}(\eigvec_j)\) using using an MLP for \(f_{\text{elem}}\). As we can see in \autoref{fig:appendix_playground_all}, the encodings w/ SignNet perform considerably worse than the encodings w/o SignNet. As hypothesised in \autoref{sec:spectral}, one reason could be that our convention for choosing the eigenvectors sign resolves the sign ambiguity to a sufficient extend. Moreover, SignNet with an element-wise MLP behaves similarly to processing the absolute value of the eigenvectors. If using solely the absolute values, we loose the information about relative differences between different nodes that include sign changes. Note that this finding crucially rely on the usage of a point-wise MLP and if using a GNN (e.g., as we do for function name prediction in \autoref{sec:source_code}) SignNet appears to help achieving better performance.

\clearpage
\section{Random Walk Hyperparameter Study}\label{sec:appendix_rw_ablation}

We next study our most important design choices along the impact of the number of random walk steps \(k\). We find that these decision are rudimental for random walk positional encodings on directed graphs and that two or three random walk steps \(k\) are sufficient for this task. 

\begin{figure}[H]
  \centering
  \includegraphics[width=0.9\linewidth]{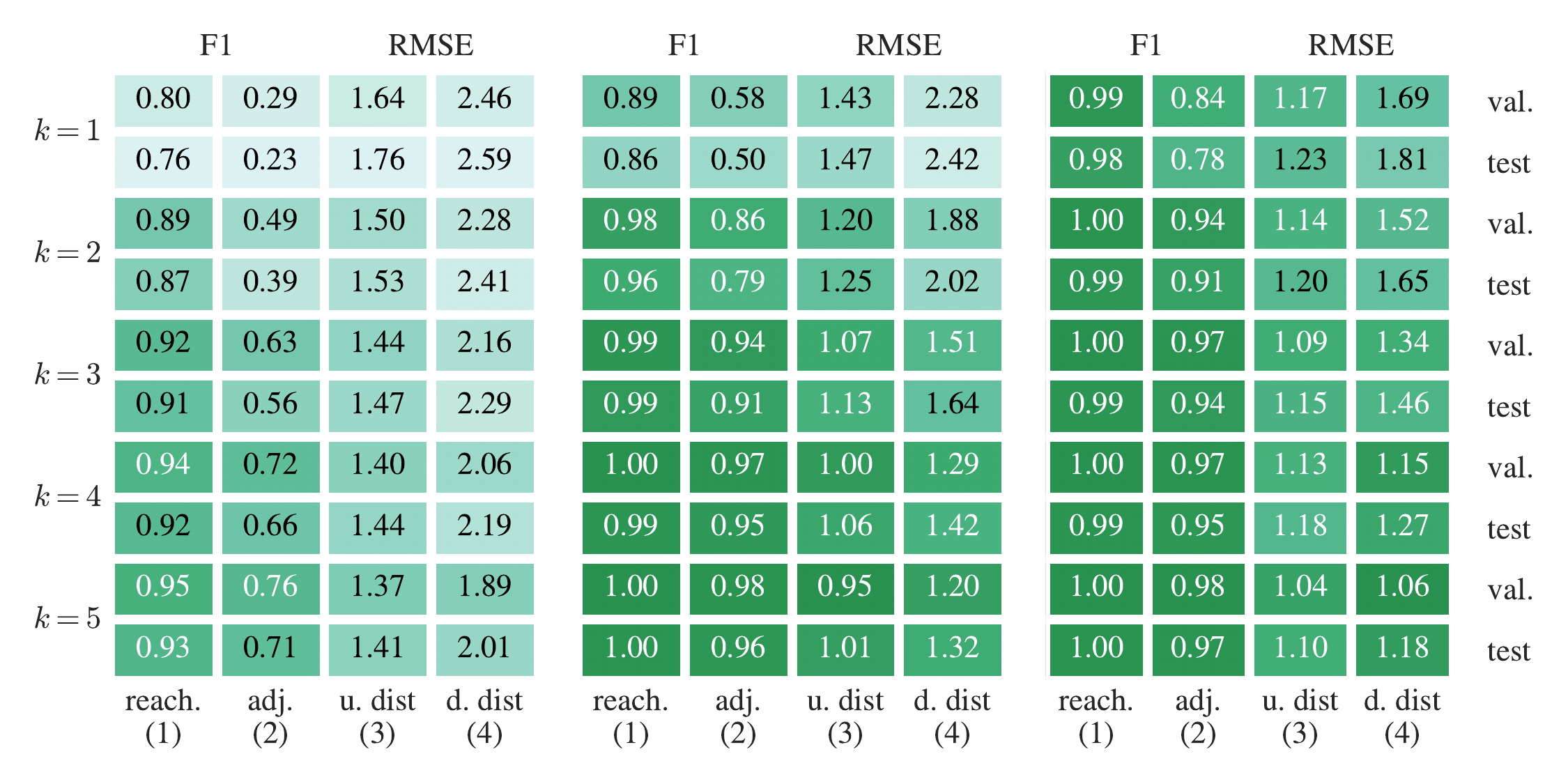}\\[-4ex]
  \subfloat[W/o rev., w/o PPR \citep{li_distance_2020}\label{fig:appendix_playground_rw:a}]{\hspace{0.3\linewidth}}
  \subfloat[W/ reversal, w/o PPR\label{fig:appendix_playground_rw:b}]{\hspace{0.25\linewidth}}
  \subfloat[W/ reversal, w/ PPR\label{fig:appendix_playground_rw:c}]{\hspace{0.3\linewidth}}
  \caption{Hyperparemter study for random walk encodings on the playground tasks: \emph{(1) reachability}, \emph{(2) adjacency}, \emph{(3) undirected distance}, and \emph{(4) directed distance} Dark green encodes the best scores and bright green bad ones. For F1 score high values are better and for RMSE low values.}
  \label{fig:appendix_playground_rw}
  \vspace{-0.12in}
\end{figure}

In \autoref{fig:appendix_playground_rw:a}, we show the random walk encodings alike \citet{li_distance_2020}. That is, solely relying on the transition matrix \(\mT = \mA \mD_{\text{out}}^{-1}\), without backward direction and without Personalizd Page Rank (PPR). In \autoref{fig:appendix_playground_rw:b}, we show the results with reversed random walk (transition matrix \(\mR = \mA^\top \mD_{\text{in}}^{-1}\)). In \autoref{fig:appendix_playground_rw:c}, we study the random walk encodings as presented in \autoref{sec:random_walks}, including backward random walk and PPR. Both, choices substantially boost performance, although, for PPR the gains diminish for \(k > 3\) (on the admittedly small graphs in this task).

\section{Sorting Networks Dataset Construction}\label{sec:appendix_sorting_networks}

\begin{figure}[t]
\centering
\hfill
\begin{minipage}{.6\linewidth}
\centering
\begin{algorithm}[H]
  \caption{Generate Sorting Network}
  \label{algo:appendixsortingnetwork}
  \begin{algorithmic}[1]
    \STATE {\bfseries Input:} Number of nodes set \(N\)%
    \STATE Sample uniformly \(n \sim U(N)\)
    \STATE Empty list of comparators \(C \leftarrow []\)
    \STATE Unsorted sequences \(S \leftarrow \operatorname{all\,\,possible\,\,sequences}(\{0, 1\}, n)\)
    \STATE Unsorted locations \(L \leftarrow \{0, 1, \dots, n-1\}\)
    \WHILE{\(|S| \ge n + 1\)}%
        \STATE Sample uniformly \((u, v) \sim U(L)\) without replacement
        \IF{\((u, v) \ne C[-1]\)}
            \STATE \(C \leftarrow C + [(u, v)]\)
            \STATE \(S \leftarrow \operatorname{sort\,\,positions}(S, u, v)\)
            \STATE \(L \leftarrow \operatorname{determine\,\,unsorted}(S)\)
        \ENDIF
    \ENDWHILE

    \STATE Return \(C\)
  \end{algorithmic}
\end{algorithm}
\end{minipage}
\hfill
\begin{minipage}{.3\linewidth}
    \centering
    \includegraphics[width=0.9\linewidth]{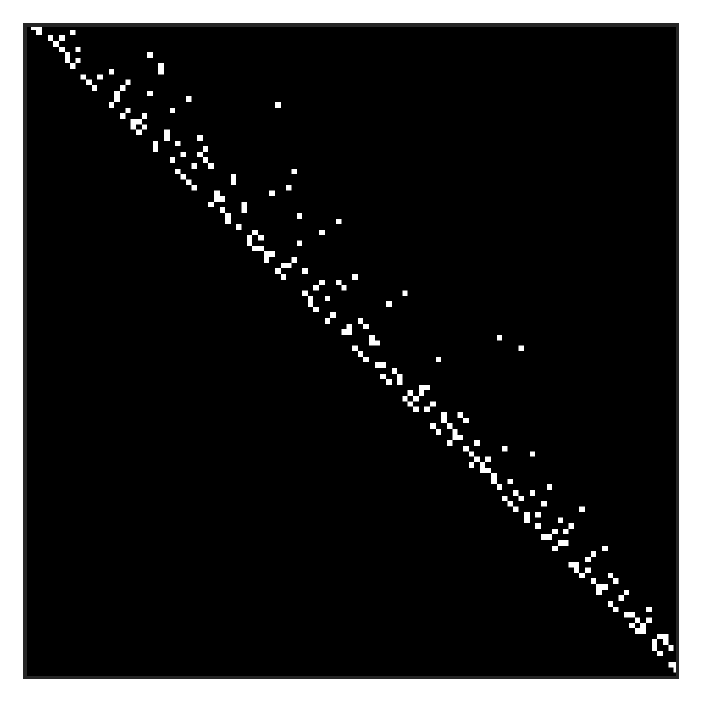}
    \caption{%
    Sorting networks are \emph{near-sequential} since the edges align well with main diagonal.}
    \label{fig:appenidx_sorting_adj}
\end{minipage}
\hfill
\end{figure}

We give the data generation process for a single sorting network in \hyperref[algo:appendixsortingnetwork]{Algorithm~\ref{algo:appendixsortingnetwork}}. Additionally, we only consider sorting networks with less than 512 comparators and abort early if this bound is exceeded. Since adding a comparator cannot diminish any progress of the sorting network, we only need to generate all possible test sequences once in the beginning and sort ti successively. Moreover, we make use of the so-called ``0-1-principle''~\citep{knuth_art_1973}. That is, if a sorting network sorts all possible sequences in \(\{0, 1\}^p\), then it sorts all possible sequences consisting of comparable elements. Once a valid sorting network was constructed, we drop the last comparator \(C[:-1]\) for an instance of an incorrect sorting network. Moreover, for test and validation, we include another (typically incorrect) sorting network via swapping the order of comparators \(C[::-1]\).

\section{Sorting Networks Are Near-Sequential}\label{sec:appendix_sorting_networks_near_seq}

All the comparators perform a data-dependent in-place swap operation. In other words, there are no buffers to store, e.g., a value at the beginning of the program and retrieve it at the end. Consequently, edges in the resulting directed graph (e.g.\ \autoref{fig:sorting_graphs}) are typically connected to close ancestor nodes and align well with the main diagonal of the adjacency matrix. We give an example in \autoref{fig:appenidx_sorting_adj}. For this reason, we call the graphs in the sorting network task \emph{near-sequential}.

Due to the near-sequentiality, sinusoidal positional encodings are intuitively an appropriate choice (see \autoref{sec:sinlap}). However, even these near-sequential graphs can have a huge amount of topological sorts. Enumerating all topological orders for the training instances rarely terminates within a few minutes (using python). We estimate, that the number of topological sorts typically exceeds \num{1e6}. This, surprisingly high number of topological sorts could be the reason why the positional encodings for directed graphs are superior to the sinusoidal encodings (see \autoref{sec:sorting_networks}) and shows the significance of this relationship.

\section{Application: Function Name Prediction}\label{sec:appendix_source_code}

We next describe how we construct the directed graph in the function name prediction task. Recall, that we construct a data-flow-centric directed graph that is also able to handle the sharp bits like if-else, loops, and exceptions. In our graph, we avoid sequential connections that originate from the sequence of statements in the source. We aim for a similar reduction of the input space size to the sorting network task \autoref{sec:sorting_networks}. To explain how we construct this graph, we will first give a high-level description of the design narratives. Then, we include the Abstract Syntax Tree (AST) in the discussion.

\begin{figure}[H]
    \hspace{-15pt}
    \begin{minipage}{0.325\textwidth}
        \begin{minted}{python}
        def func(a, b, c):
          d = min(a, b)
          if a > 0:
            e = a ** 2
            f = b ** 2
            a = sqrt(e + f)
          else:
            a = -a
          if b < 0:
            b = b + d
          return a + b
        \end{minted}
    \end{minipage}
    \hspace{25pt}
    \begin{minipage}{0.6\textwidth}
        \includegraphics[width=\linewidth]{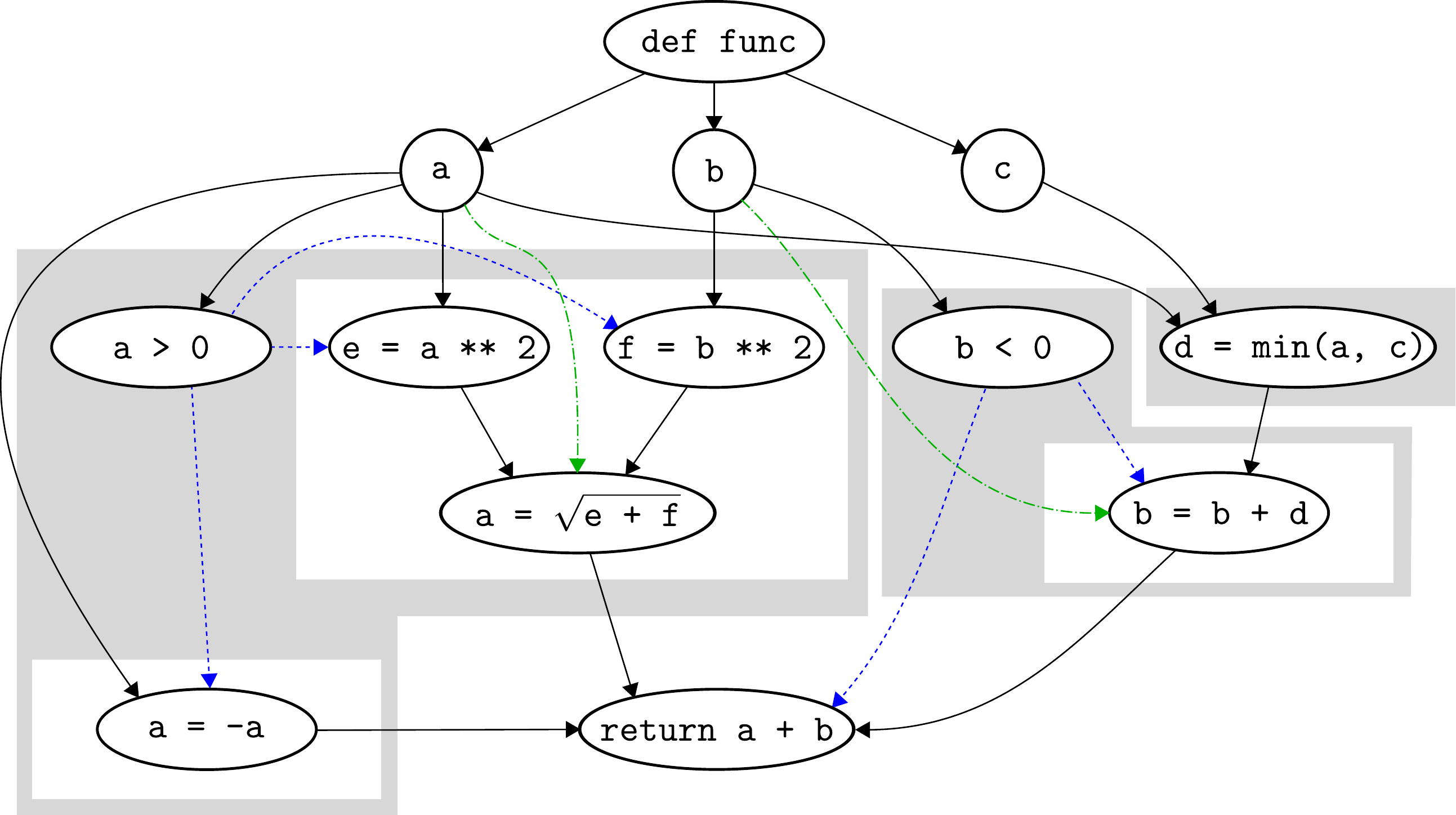}
    \end{minipage}
    \hfill{}
    \caption{Exemplary data-flow-centric graph construction\label{fig:appendix_data_centric_construction}}
\end{figure}

\textbf{Design principles.} In \autoref{fig:appendix_data_centric_construction}, we give an example function and give a data-flow-centric dependency graph for the actual computations. The function can be clustered into three blocks (excluding function definition and return statement): (1) variable \(d\) is calculated, (2) if-else-statement for refining the value of variable \(a\), and (3) variable \(b\) is changed if negative. These three blocks are each represented by a grey box. Further, we highlight (sub-) blocks white that correspond to the bodies of the if-else statement.

We connect nodes based on the required inputs for the computation (aka data flow). Moreover, we add edges to describe the control flow of the if-statements (dashed blue lines). Last, we add edges if values are being overwritten (dash-dotted green line). Conceptually, we generate a Directed Acyclic Graph (DAG) for each block and then recursively traverse the hierarchy and connect the contained instructions. Hence, the resulting graph is not necessarily acyclic.

\textbf{Order of computation.} In this example, each computation can only take place after all its immediate ancestors have been calculated (and if these values are kept in memory). Vice versa, all computations could take in arbitrary order as long as the parent nodes have been executed. For general functions, the picture is a bit more complicated (but similar) due to, for example, loops, exceptions, or memory constraints.

\begin{figure}[H]
    \hfill
    \subfloat[OGB Code 2 graph \label{fig:appendix_ogb_vs_ours:a}]{\includegraphics[width=0.25\linewidth]{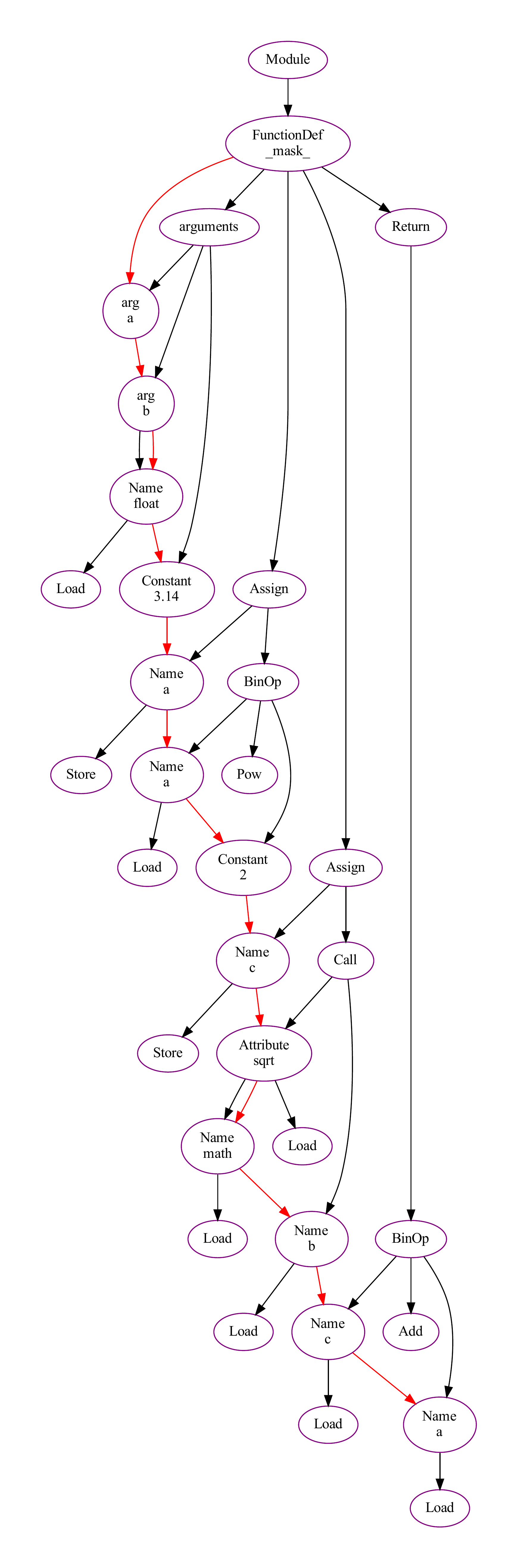}}
    \hfill
    \subfloat[Our data-centric graph\label{fig:appendix_ogb_vs_ours:b}]{\includegraphics[width=0.74\linewidth]{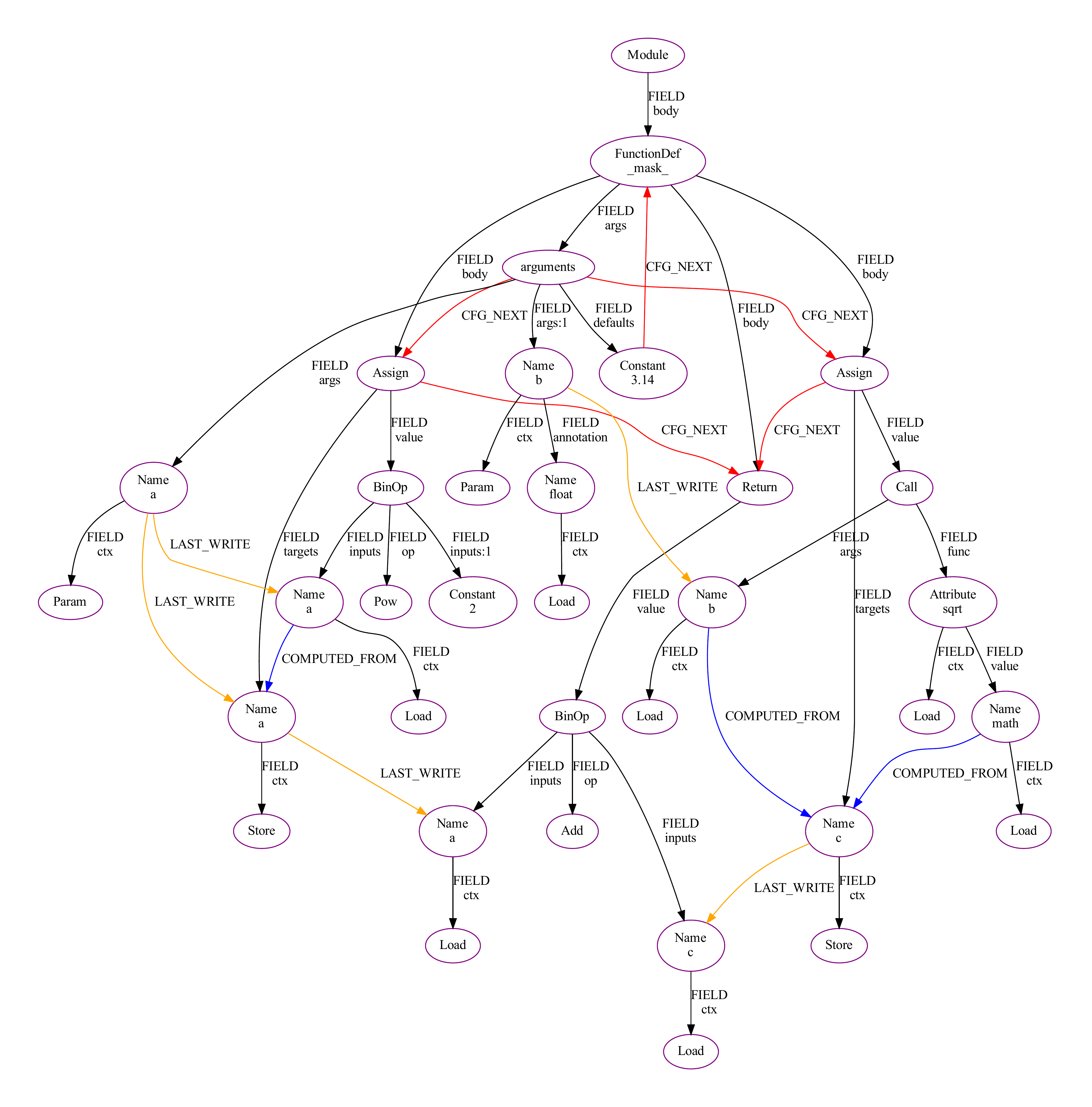}}
    \hfill
    \caption{Comparison of OGB Code2's graph construction to ours. \label{fig:appendix_ogb_vs_ours}}
\end{figure}

\textbf{Hierarchy.} To generate the connections we rely on the fact that (Python) code can be decomposed into a hierarchy of blocks. Except for ancestor and successor nodes, these blocks build a mutually exclusive separation of instructions. This decomposition is what an AST provides. While also prior work uses ASTs for their graph construction, they retain the sequential structure of the source code in the graph construction. For example in \autoref{fig:appendix_ogb_vs_ours:a}, we show the graph constructed by the OGB Code2 dataset~\citep{hu_open_2020} for the code in \autoref{fig:appendix_ogb_vs_ours_code}. From this, it should be clear without additional explanation why we argue that the AST solely enriches the sequence of code statements with a hierarchy. In stark contrast, our graph construction (\autoref{fig:appendix_ogb_vs_ours:b}) is by far not as sequential.

\textbf{Sequentialization.} Generating semantically equivalent sequences of program statements from such a directed graph is more challenging than determining feasible orders of computation or in the sorting network task \autoref{sec:sorting_networks}. For example, in DAG of \autoref{fig:appendix_data_centric_construction}, not every possible topological sort corresponds to a possible sequentialization of the program. To determine sequentializations one needs to consider the hierarchical block structure. For example, it is possible to reorder the blocks highlighted in grey, depending on their dependencies. However, our data and control flow does not capture all dependencies required to generate the program. For example, as already hinted above, one caveat resides in the availability of intermediate values. Although the first block (to determine \verb|d|) and second block (if else construct) are independent in the shown graph, \emph{overwriting} the value of \verb|a| has not been modeled. In other words, it would make a difference to swap these blocks since \verb|a| changes its value in the second block. Thus, for constructing a possible sequence of program instructions, we would also need to address changing variable values. For example, we could assign a new and unique name after changing a variable's value (as in functional programming or like a compiler does). Alternatively, adding further edges could be sufficient. Nevertheless, these dependencies are not important for the semantic meaning of a program.

\begin{wrapfigure}[9]{r}{0.45\textwidth}
    \vspace{-15pt}
    \hspace{-25pt}
    \begin{minted}{python}
    def transform_add(
        a, b: float = 3.14):
      a = a**2
      c = math.sqrt(b)
      return c + a
    \end{minted}
    \caption{Code used for \autoref{fig:appendix_ogb_vs_ours}.\label{fig:appendix_ogb_vs_ours_code}}
\end{wrapfigure}
\textbf{OGB's graph construction} first converts the source to AST and adds additional edges to simulate the sequential order of instructions. In \autoref{fig:appendix_ogb_vs_ours:a}, the black edges are the edges from the AST and the red edges for the sequential order.

\textbf{Our graph construction.} We also construct the AST from source (\verb|FIELD| edges) and build on top of the graph construction / static code analysis of \citet{bieber_library_2022}. In the example in \autoref{fig:appendix_ogb_vs_ours:b}, we have the same amount of nodes as \autoref{fig:appendix_ogb_vs_ours:a} except for two ``Param'' nodes belonging to the argument nodes. Similarly to the example in \autoref{fig:appendix_data_centric_construction}, we augment the AST with additional edges mimicking the data flow and program flow. Here, we have Control Flow Graph edges \verb|CFG_NEXT| that model the possible order of statement execution. Moreover, the variable nodes (close to leaf nodes) are connected by \verb|LAST_WRITE| and \verb|CALCULATED_FROM| edges. These edges model where the same variable has been written the last time and from which variables is was constructed. Additionally, we use a \verb|CALLS| edge that model function calls / recursion (not present in this example). Thereafter, since the task is function name prediction, we need to prevent data leakage. For this, we mask the occurrences of the function name in its definition as well as in recursive calls.

\textbf{Comparison to \citet{bieber_library_2022}.} We largely follow them and build upon their code base. The most significant difference is the avoidance of unnecessary sequentialism. Specifically, (1) their \verb|CFG_NEXT| edges connect instructions sequentially while ours form a dependency graph, and (2) we omit their \verb|LAST_READ| edges. Moreover, we address commutative properties of the basic python operations (\verb|And|, \verb|Or|, \verb|Add|, \verb|Mult|, \verb|BitOr|, \verb|BitXor|, and \verb|BitAnd|). This can also be observed in \autoref{fig:appendix_ogb_vs_ours:b}, where we name the edges for the inputs to these operations \verb|input| and concatenate \verb|:<order>| if the operation is non-commutative and \verb|<order>| > 1. Last, we do not reference the tokenized source code and choose a less verbose graph similar to OGB. 

\textbf{Sequentialization.} Reconstructing the code from AST is a straightforward procedure. Following our discussion above, we need to acknowledge the hierarchical structure to retrieve valid programs. Fortunately, this hierarchical structure is provided by the AST. However, similar to the example above, we do not model all dependencies for an immediate sequentialization. However, as stated before, these dependencies are not important for semantics. Thus, we most certainly map more semantically equivalent programs to the same directed graph, as if we would compare to a graph construction that models all dependencies.

\end{document}